\pdfoutput=1

\documentclass[11pt]{article}

\usepackage[preprint]{acl}

\usepackage{times}
\usepackage{latexsym}

\usepackage[T1]{fontenc}

\usepackage[utf8]{inputenc}

\usepackage{microtype}

\usepackage{inconsolata}

\usepackage{graphicx}
\usepackage{booktabs} 
\usepackage{multirow}
\usepackage{array} 
\usepackage{subcaption}
\usepackage{xcolor,colortbl}

\usepackage{amssymb}
\usepackage{bbding}

\newcommand{\benchname}{\emph{MFC-Bench}}

\newcommand{\MFCemoji}{\includegraphics[height=1.8\fontcharht\font`\B]{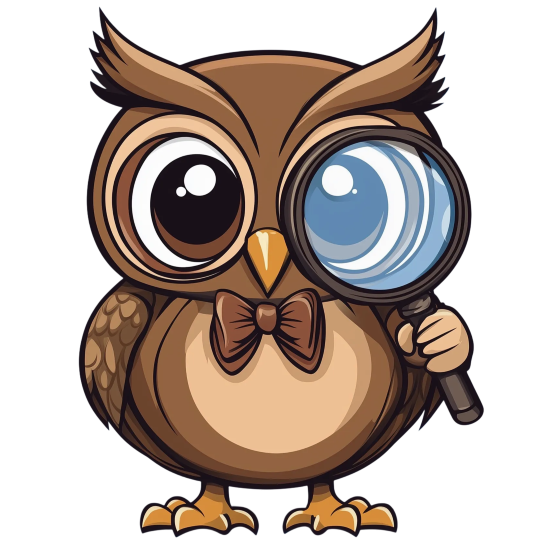}}

\newcommand{\humanemoji}{\includegraphics[height=1.5\fontcharht\font`\B]{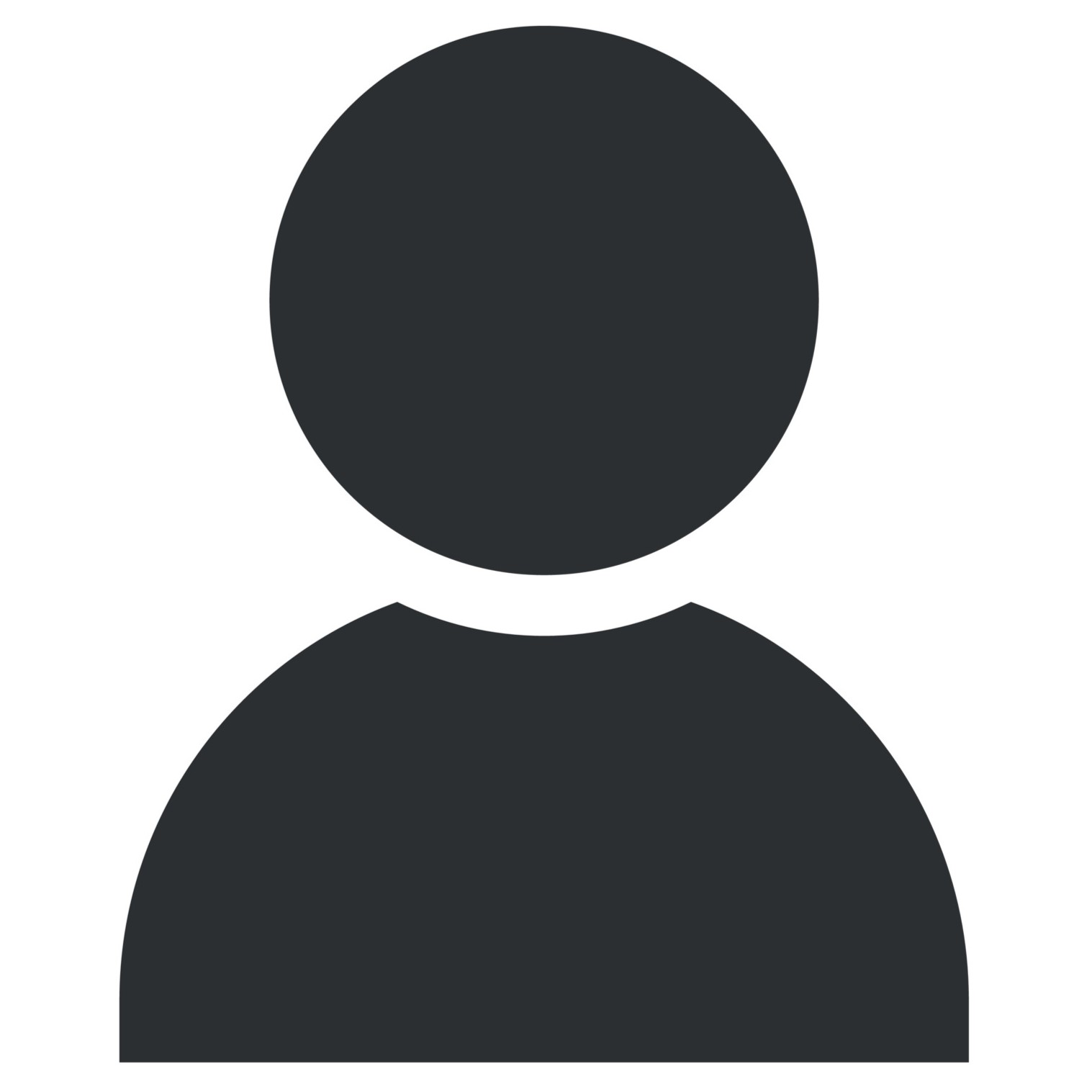}}

\newcommand{\Qwenemoji}{\includegraphics[height=1.2\fontcharht\font`\B]{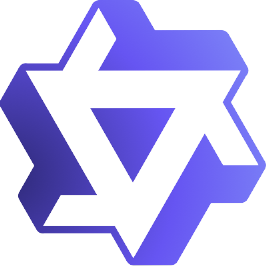}}

\newcommand{\Googleemoji}{\includegraphics[height=1.3\fontcharht\font`\B]{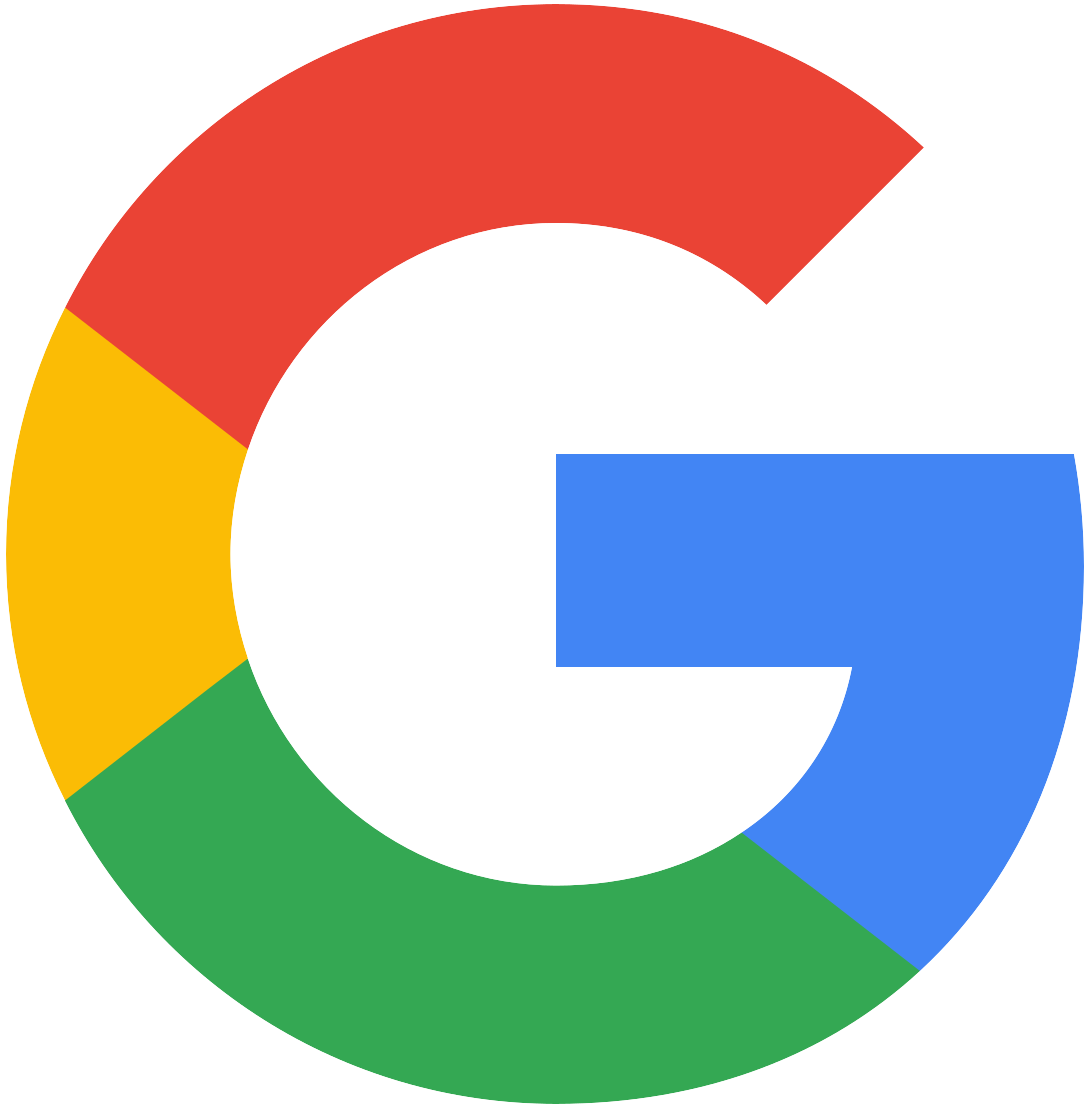}}

\newcommand{\Openaiemoji}
{\includegraphics[height=1.2\fontcharht\font`\B]{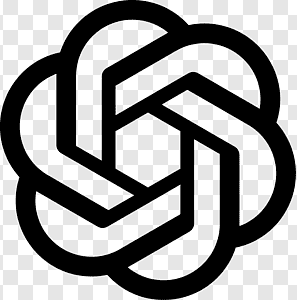}}
\newcommand{\Claudeemoji}{\includegraphics[height=1.2\fontcharht\font`\B]{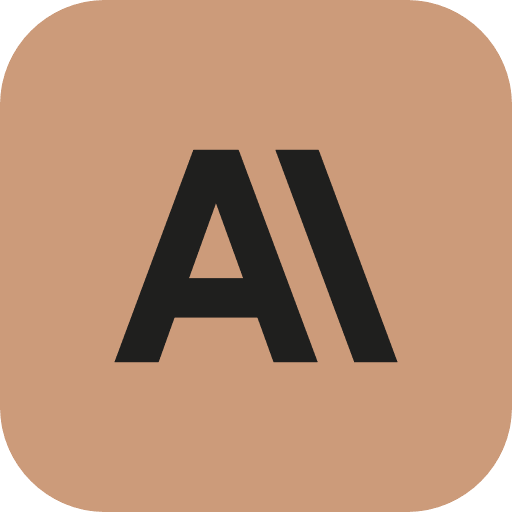}}

\newcommand{\BAAIemoji}{\includegraphics[height=1.5\fontcharht\font`\B]{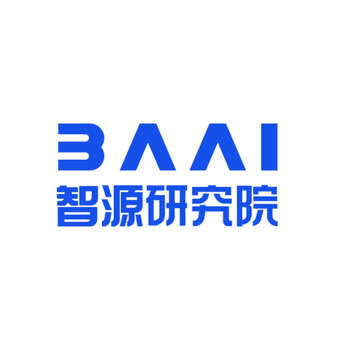}}
\newcommand{\InternVLemoji}{\includegraphics[height=1.5\fontcharht\font`\B]{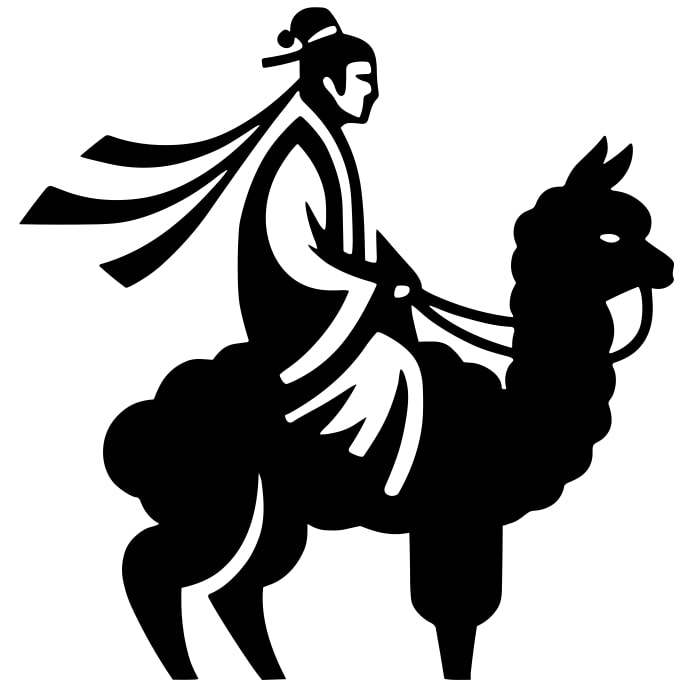}}
\newcommand{\CogVLMemoji}{\includegraphics[height=1.5\fontcharht\font`\B]
{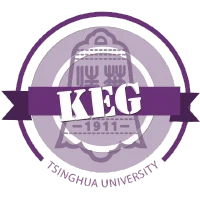}}

\newcommand{\llavanextmoji}{\includegraphics[height=1.5\fontcharht\font`\B]{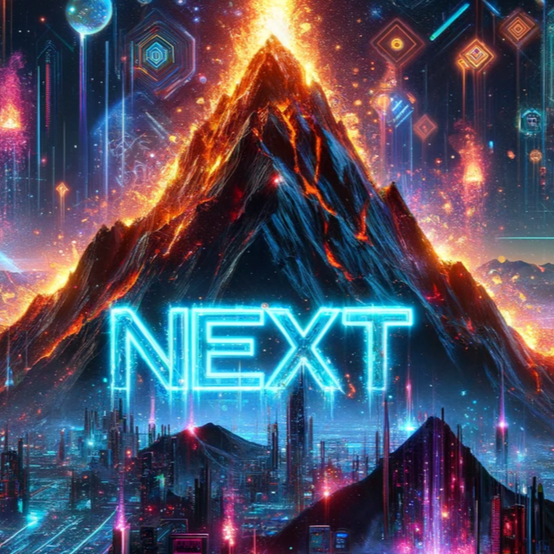}}

\newcommand{\Instructblipemoji}{\includegraphics[height=1.5\fontcharht\font`\B]{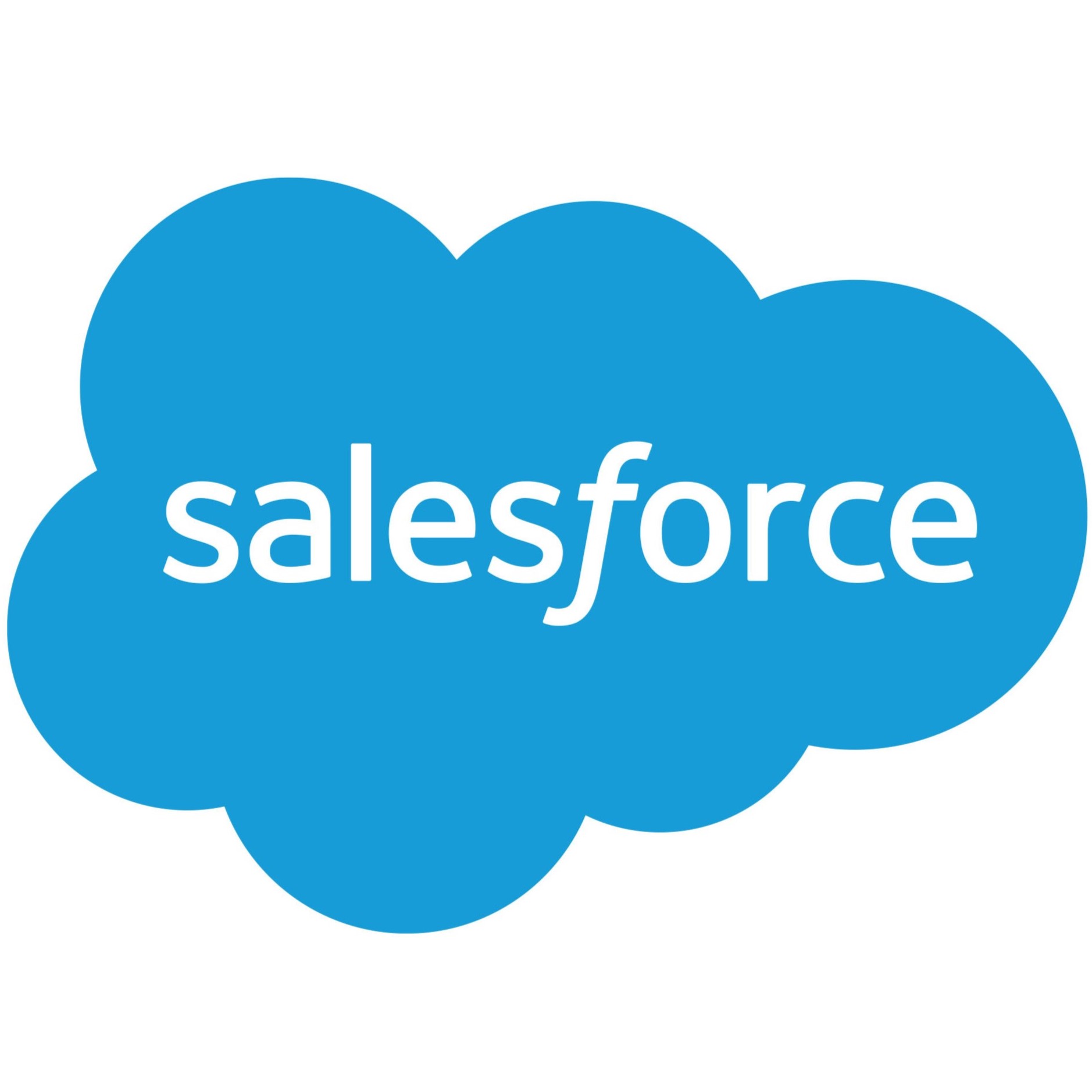}}
\newcommand{\mPLUGOwlemoji}{\includegraphics[height=1.5\fontcharht\font`\B]{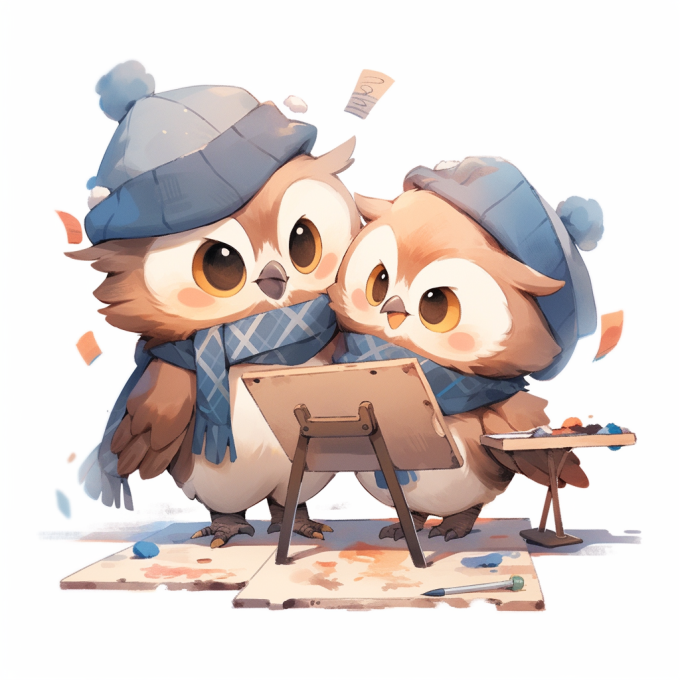}}

\newcommand{\YiVLemoji}{\includegraphics[height=1.5\fontcharht\font`\B]{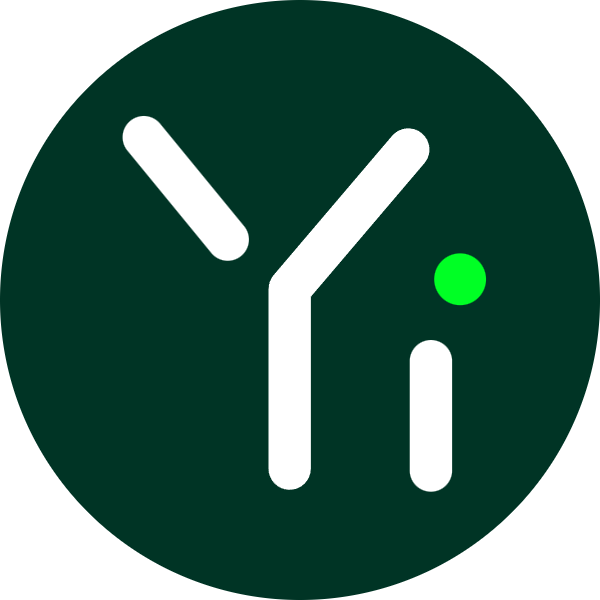}}

\newcommand{\molmoemoji}{\includegraphics[height=1.5\fontcharht\font`\B]{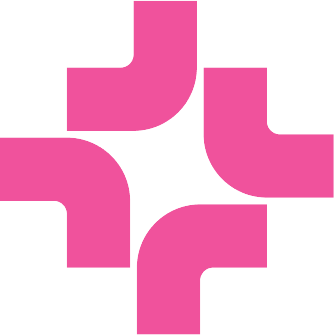}}
\newcommand{\pixtralemoji}{\includegraphics[height=1.5\fontcharht\font`\B]{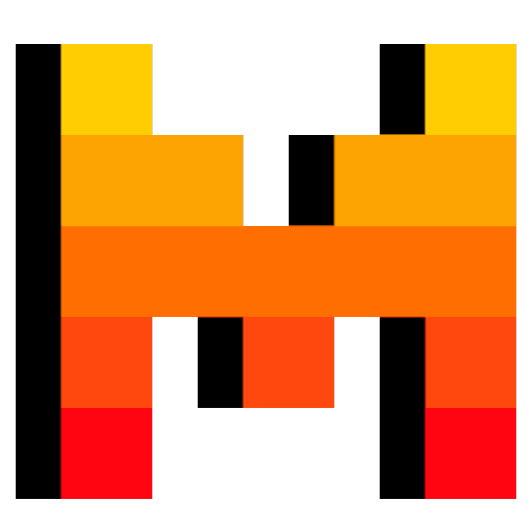}}

\newcommand{\MiniCPMVemoji}{\includegraphics[height=1.5\fontcharht\font`\B]{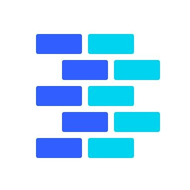}}

\title{\MFCemoji{}\benchname{}: Benchmarking Multimodal Fact-Checking with Large Vision-Language Models}

\author{%
  Shengkang Wang$^{1,}$\thanks{Equal contribution.}, Hongzhan Lin$^{2,*}$, Ziyang Luo$^{2,*}$,\\ \textbf{Zhen Ye}$^{3}$, \textbf{Guang Chen}$^{1,\dagger}$, \textbf{Jing Ma}$^{2,}$\thanks{Corresponding authors.} \\
  $^{1}$Beijing University of Posts and Telecommunications\\
  $^{2}$Hong Kong Baptist University\\
  $^{3}$Hong Kong University of Science and Technology\\
  \texttt{\{wsk, chenguang\}@bupt.edu.cn}, \texttt{\{cshzlin, cszyluo, majing\}@comp.hkbu.edu.hk} \\
}

\begin{document}
\maketitle
\begin{abstract}
  Large vision-language models (LVLMs) have significantly improved multimodal reasoning tasks, such as visual question answering and image captioning. These models embed multimodal facts within their parameters, rather than relying on external knowledge bases to store factual information explicitly. However, the content discerned by LVLMs may deviate from factuality due to inherent bias or incorrect inference. To address this issue, we introduce \benchname{}, a rigorous and comprehensive benchmark designed to evaluate the factual accuracy of LVLMs across three stages of verdict prediction for multimodal fact-checking (MFC): Manipulation, Out-of-Context, and Veracity Classification. Through our evaluation on \benchname{}, we benchmarked a dozen diverse and representative LVLMs, uncovering that current models still fall short in MFC and demonstrate insensitivity to various forms of manipulated content. We hope that \benchname{} could raise attention to the trustworthy AI potentially assisted by LVLMs in the future. The \benchname{} and accompanying resources are publicly accessible at \textcolor{magenta}{\url{https://github.com/wskbest/MFC-Bench}}, contributing to ongoing research in the MFC field.
\end{abstract}

\section{Introduction}

Recent advancements in natural language processing (NLP), particularly with large language models (LLMs) \citep{chang2023survey}, have introduced tools like ChatGPT and GPT-4 \citep{OpenAI2023GPT4TR} that excel in understanding human instructions using strategies such as instruction tuning and reinforcement learning from human feedback~\citep{ouyang2022training}. These models demonstrate strong zero-shot or few-shot capabilities, performing tasks without additional fine-tuning~\citep{kojima2022large, lin2023zero}. Simultaneously, large vision-language models (LVLMs) \citep{Dai2023InstructBLIPTG, gong2023multimodal} have extended this proficiency to multimodal understanding tasks \citep{fu2023mme}. These advancements mark a significant step forward in artificial intelligence, enabling more cohesive applications across modalities.

\begin{figure*}[t]
  \centering
    \includegraphics[width=\linewidth]{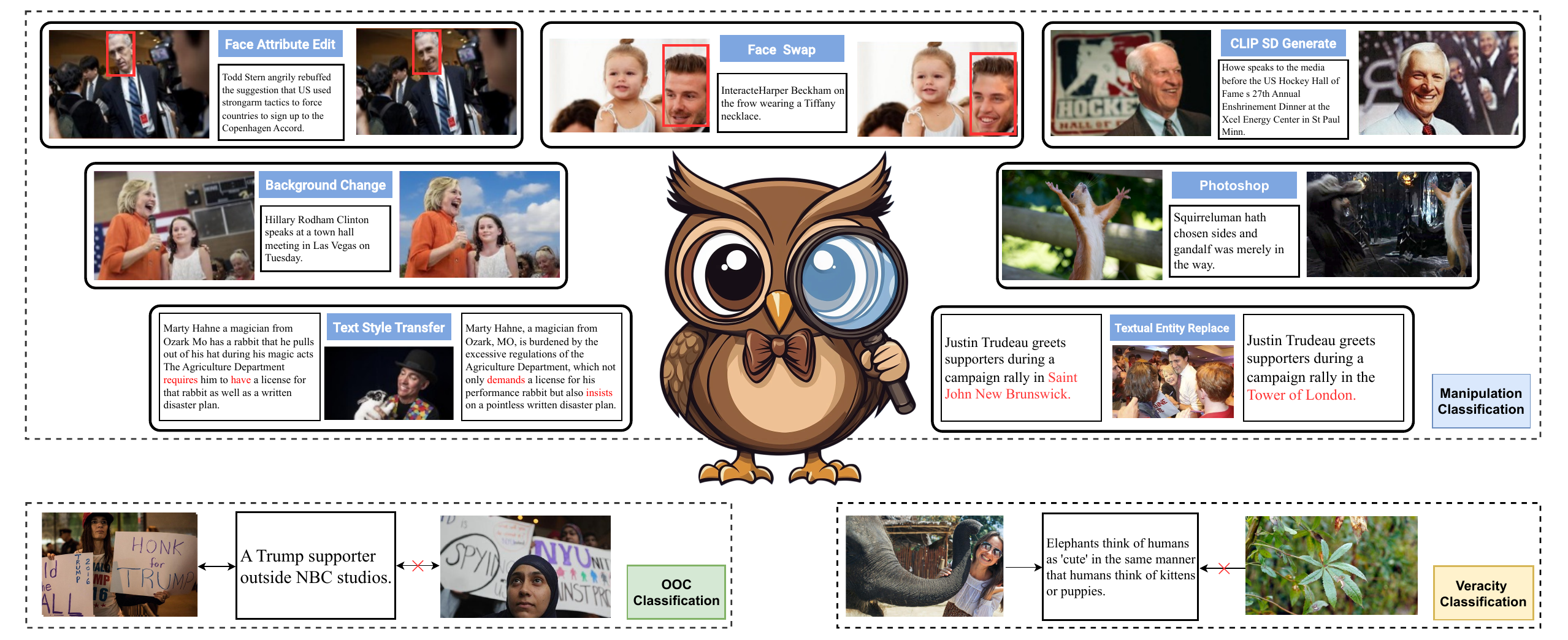}
    \vspace{-0.5cm}
  \caption{\benchname{} is a comprehensive benchmark designed to evaluate the LVLMs across three stages of verdict prediction for MFC: Manipulation Classification, Out-of-Context Classification, and Veracity Classification.}
  \label{fig:architecture}
  \vspace{-0.4cm}
\end{figure*}

Recent studies have thoroughly investigated the extent to which LLMs hold factual information and their capacity to reason with such knowledge~\citep{hu2024large}, which hypothesized that LLMs, trained on vast data, could adequately substitute for evidence retrieval and conduct fact-checking autonomously, relying solely on their parametric knowledge. Beyond text-only fact-checking~\citep{guo2022survey, thorne2018fever}, multimodal content is often perceived as more credible and spreads more quickly than similar textual claims~\citep{li2020picture, newman2012nonprobative}. However, the capabilities and limitations of LVLMs in managing multimodal reasoning tasks~\citep{akhtar2023multimodal} related to factuality, particularly in identifying online unverified information within multimodal inputs, remain underexplored. These multimodal fact-checking tasks \citep{nakamura-etal-2020-fakeddit, Shao_2023_CVPR,10.1145/3539618.3591879} are crucial for understanding social dynamics and require sophisticated social judgment and decision-making abilities. Thus, a fundamental question remains: \textit{Can LVLMs discern factuality in a multimodal context?} Given that LVLMs are trained on extensive and varied image-text corpora and demonstrate remarkable generalization capabilities~\citep{liu2023visual}, it is vital to evaluate both their strengths and potential challenges in handling factual knowledge and reasoning. This inquiry is particularly relevant to ensuring trustworthy insights, focusing on how LVLMs analyze and integrate complex visual and textual elements accurately and responsibly.

Previous literature~\cite{akhtar2023multimodal} has surveyed that there are three important stages for verdict prediction in multimodal fact-checking process: 1) Manipulation Classification; 2) Out-of-Context (OOC) Classification; 3) Veracity Classification. In this work, we aim to comprehensively explore the helpfulness of \textbf{LVLMs} in benchmarking multimodal fact-checking within these three tasks. To this end, we introduce \benchname{} as shown in Figure \ref{fig:architecture}, a comprehensive Multimodal Fact-Checking testbed designed to evaluate LVLM in terms of identifying factual inconsistencies and counterfactual scenarios.
\benchname{} encompasses a wide range of visual and textual queries, organized into the three verdict prediction tasks: Manipulation Classification, OOC Classification, and Veracity Classification. These three sub-tasks of multimodal fact-checking draw from a mix of diverse datasets~\citep{newman2012nonprobative, Shao_2023_CVPR,10.1145/3539618.3591879} and our newly created datasets specifically designed for analyzing awareness of multimodal facts: 1) The Manipulation Classification task targets various alterations like face swapping, face attribute editing, background changing, image generation, photoshop, entity replacement, and style transfer;  
2) The OOC Classification task focuses on identifying the false connection between the image and text that may be both true; 3) The Veracity Classification task is the multimodal counterpart to classifying the veracity of \textit{textual claims} given the visual evidence, by leveraging the inherent knowledge embedded in LVLMs. As a result, such a three-task design philosophy could facilitate evaluating the effectiveness of LVLMs in supporting verdict prediction during the multimodal fact-checking process.
We presented MFC tasks to LVLMs with carefully crafted prompts, gathered the model's feedback, and conducted a comprehensive analysis of the outcomes, which ensures a thorough understanding of LVLMs' capabilities and limitations on \benchname{}.

Through \benchname{}, we comprehensively assess the ability of various LVLMs~\citep{bai2023qwen, Dai2023InstructBLIPTG, liu2023visual, OpenAI2023GPT4TR} to accurately identify manipulated and misleading content within multimodal inputs.
Our benchmark offers a rigorous examination of current LVLMs, highlighting the considerable gaps in their performance. Tasks aimed at detecting false connections, such as OOC Classification, reveal pronounced disparities in LVLM efficacy. For more intricate tasks like Manipulation Classification, which necessitates deep background knowledge and sophisticated reasoning, LVLMs typically demonstrate only mediocre performance. Besides, we further explore the justification production of LVLMs for multimodal fact-checking with human subject evaluation.
Overall,
\benchname{} is designed to provide researchers with a multi-dimensional understanding of their LVLMs’ capabilities in multimodal fact-checking. Our goal is to advance auditing insights within LVLMs, playing a crucial role in curbing the spread of online disinformation and promoting the stability and cohesion of diverse communities.

\label{sec:introduction-contribution}
Our contributions are three-fold: 1) We introduce \benchname{}, a comprehensive testbed with 35K multimodal samples across three stage sub-tasks of verdict prediction in the multimodal fact-checking process to assess LVLMs' trustworthiness; 2) Extensive evaluation of a dozen advanced LVLMs reveals significant challenges, with GPT-4o only achieving F1 scores of 69.4\% on the \benchname{}; 3) We provide a detailed analysis of performance variations among different LVLMs on prompting strategies and justification production.

\begin{table*}[t]
\centering
\footnotesize
\begin{tabular}{>{\raggedright}p{2cm}p{3.5cm}p{4.5cm}ccc}
\toprule
\multirow{2}{*}{\textbf{Types}} & \multirow{2}{*}{\textbf{Description}} & \multirow{2}{*}{\textbf{Sources}} & \multicolumn{3}{c}{\textbf{Distribution}} \\
\cmidrule(lr){4-6}
 &  &  & \textbf{Fact.} & \textbf{Non-Fact.} & \textbf{All} \\
\midrule
\multirow{7}{*}{\textbf{Manipulation}} 
 & Face Swap  & DGM4~\citep{Shao_2023_CVPR} & 4,000  & 2,000 & 6,000 \\
 & Face Attribute Edit & DGM4~\citep{Shao_2023_CVPR} & 4,000  & 2,000 & 6,000 \\
 & Background Change  & - & 1,000  & 2,000 & 3,000 \\
 & CLIP-based SD Generate & - & 5,000  & 5,000 & 10,000 \\
 & Photoshop & Fakeddit~\citep{nakamura-etal-2020-fakeddit} & 1,000 & 1,000 & 2,000 \\
 & Textual Entity Replace  & -  & 1,162  & 838 & 2,000 \\
 & Text Style Transfer  & - & 1,000  & 1,000 & 2,000 \\
\midrule
 \textbf{OOC} & Detect out of context   & NewsCLIPpings~\citep{luo-etal-2021-newsclippings} & 1,000  & 1,000 & 2,000 \\
 \midrule
\textbf{Veracity}  & Verify the claim w/ image & Mocheg~\citep{10.1145/3539618.3591879} & 469  & 1,531 & 2,000 \\
\bottomrule
\end{tabular}
\caption{Dataset sources, description, and distribution.}
\label{tab:1}
\vspace{-0.4cm} 
\end{table*}

\section{Dataset Constitution}
To systematically assess the visual and textual factual knowledge related to inconsistencies and counterfactual reasoning abilities of LVLMs, we have formulated our benchmark into three decomposed sub-tasks of verdict prediction for the multimodal fact-checking process: Manipulation Classification, Out-of-Context Classification, and Veracity Classification, by considering prevalent multimodal misinformation types~\cite{akhtar2023multimodal}. 
For these multimodal misinformation types of data for verification, we carefully curate appropriate visual and textual queries from a variety of sources to ensure a comprehensive evaluation of LVLMs in multimodal fact-checking, as summarized in Table \ref{tab:1}.

\subsection{MFC Data Types}
\subsubsection{Manipulation Classification}
Manipulation Classification is a task meticulously designed to ascertain whether multimodal data encompasses fabricated elements~\citep{qi2019exploiting} by using LVLMs. To investigate LVLMs' proficiency in identifying multimodal content altered through various manipulative techniques, in \benchname{}, we organized seven types of manipulation methods\footnote{Here, we consider the most challenging setting~\cite{akhtar2023multimodal} that the correct content in one modality, accompanied by the manipulated content in the other modality, which increases credibility.}: The first five focus on visual alterations, while the last two target textual modifications. 

\textbf{Method 1: \underline{F}ace \underline{S}wap (FS).} Face Swap involves the process of cutting a face from one image and replacing it with a different face in another image. As shown in Figure \ref{fig:architecture}, through the use of face swap, Beckham's face has been replaced with a different face. We include the Face Swap data to \textit{assess whether LVLMs can recognize public figures and retrieve information related to individuals, finding the counterfactuals that emerge from these swapped faces} in the multimodal context. 

\textbf{Method 2: Face \underline{A}ttribute \underline{E}dit (AE).} Face Attribute Edit achieves deception by altering the facial expressions of humans like newsmakers. For example, in Figure \ref{fig:architecture}, Todd Stern originally had an angry expression, which was changed to a happy expression through Face Attribute Edit. This inclusion allows us to \textit{evaluate the multimodal fact-checking capabilities of LVLMs in recognizing the scene, identifying personal information and detecting the correctness of face's status} in visual content assisted with an accompanying text.

\textbf{Method 3: \underline{B}ackground \underline{C}hange (BC).} 
  Background Change alters images, transforming public individuals into scenes where he/she never showed up in reality. As depicted in Figure \ref{fig:architecture}, Hillary Rodham Clinton was originally indoors, but BC makes it seem like she is now outside. The objective is to \textit{examine the capability of LVLMs for accurate identification of individuals and scenes in images, evaluating their correspondence and authenticity in relation to the descriptions provided in texts.} 


\textbf{Method 4: \underline{C}LIP-based Stable Diffusion  \underline{G}enerate (CG).} 
  CLIP-based Stable Diffusion \cite{Ramesh2022HierarchicalTI} features an image-to-image generation pipeline that enables the manipulated image to retain the linguistic information from the original image, producing stable-diffusion versions for image replacement. Originally, Figure \ref{fig:architecture} showed Howe speaking, but with CG, the image was altered to display a generated individual giving the speech, retaining much of the original visual content. This design enables us to \textit{assess the fact-checking capabilities of LVLMs regarding their awareness of whether multimodal content is fabricated, even when the manipulated image retains elements of the original alongside the raw text.}
  

\textbf{Method 5: \underline{P}hoto\underline{s}hop (PS).} Photoshop has long been a leading manipulation for manual image editing, enabling users to alter human figures and merge different images to create potentially misleading visuals. As demonstrated in Figure \ref{fig:architecture}, using Photoshop, an ordinary squirrel can be seen battling Gandalf in a single picture. Including this data type allows us to \textit{assess whether LVLMs can discern the traces of human manipulation in image accompanying the original text. }
  

\textbf{Method 6: Textual \underline{E}ntity \underline{R}eplace (ER).} Textual Entity Replace involves substituting entities other than the target persons in the data, with randomly chosen locations and time. As exemplified in Figure \ref{fig:architecture}, Justin Trudeau was originally shown greeting in Saint John, New Brunswick, Canada, but with Textual Entity Replace, it was changed to him greeting at the Tower of London, British.
  This method seeks to \textit{assess the capability of LVLMs to effectively associate individuals with the entities depicted in both images and texts, highlighting any inconsistencies.}

\textbf{Method 7: Text \underline{S}tyle \underline{T}ransfer (ST).} Text Style Transfer is the process of modifying the tone and style of a text to alter the perception of the same person or event, potentially leading to a different factual impression. As Figure \ref{fig:architecture}, by Text Style Transfer, the tone shifts from a neutral, factual statement about Marty Hahne needing a license and disaster plan for his rabbit, to a more critical and dramatic tone, portraying the requirements as burdensome and excessive. The process \textit{examines LVLMs' ability to rigorously comprehend the events and associated sentiments depicted in images and claims, and to correctly correlate them.}

\subsubsection{Out-of-Context Classification}
Out-of-Context (OOC) Classification in \benchname{} aims to decipher the coherence and correspondence of context across various modalities~\citep{luo-etal-2021-newsclippings} with LVLMs. Unlike the aforementioned manipulation techniques that require modifying images and texts, OOC Classification combines real but misused images and texts. If the image and the text are contextually aligned, the relationship is regarded as true, naturally representing fact. Conversely, if the image and the text are not contextually aligned, the relationship is regarded as false, indicating non-fact.

As shown in Figure \ref{fig:architecture}, the left image accurately reflects the text, with a girl holding a sign to support Trump, whereas the right image, showing a Muslim protest, is an incorrect connection.
We collected multimodal samples from the NewsCLIPpings dataset~\cite{luo-etal-2021-newsclippings}, using embedding methods such as CLIP and SBERT-WK~\cite{Wang2020SBERTWKAS} to extract the most similar misused images, for \textit{the evaluation of LVLMs' ability in discerning subtle semantic inconsistencies between images and texts} in OOC Classification.

\subsubsection{Veracity Classification}
Veracity Classification in \benchname{} serves to classify the factuality of textual claims based on visual evidence~\citep{10.1145/3539618.3591879} by employing LVLMs. 
Based on the image evidence, the LVLMs need to predict the truthfulness of the textual claim. We curated a subset of the Mocheg dataset~\cite{10.1145/3539618.3591879} for this task. If the image evidence supports the truthfulness of the textual claim, the relationship between the image and the claim is supported, indicating fact. Otherwise, the claim is treated as refuted by the image, exhibiting non-fact.

As depicted  in Figure \ref{fig:architecture}, the friendly interaction between the elephant and the human on the left image can be inferred to support the statement ``Elephants think of humans as `cute' in the same manner that humans think of kittens or puppies.'' However, the right image is plants, making it impossible to verify the truth of the text based on the image evidence.
This is a cross-modal semantic transformation task designed to 
\textit{test whether LVLMs can accurately interpret and analyze visual information to support or refute textual claims}.

\subsection{Label Setting}

To unify the three tasks and facilitate a more effective analysis of benchmark results, we formulate the tasks into binary classification, we define the label \( L =\{\texttt{Fact.}, \texttt{Non-Fact.}\} \). The Manipulation Classification task involves determining whether multimodal news is fabricated, with labels indicating ``\texttt{Manipulated}'' (Non-Fact.) or ``\texttt{Not Manipulated}'' (Fact.). The OOC Classification task assesses whether the image and claim are inconsistent, with labels indicating ``\texttt{Matched}'' (Fact.) or ``\texttt{Not Matched}'' (Non-Fact.). The Veracity Classification task evaluates whether the claim is true based on image evidence, with labels indicating ``\texttt{Supported}'' (Fact.) or ``\texttt{Refuted}'' (Non-Fact.).

\subsection{Quality Assurance}
Multiple levels of measures were implemented to guarantee data reliability.
First, we utilized established and reputable technologies such as Stable Diffusion and GPT-4 for data processing, ensuring that the operations were reasonable and aligned with our expectations. Second, we incorporated other well-regarded datasets that are time-tested and frequently cited. The tasks represented by these datasets coincide with the objectives of our benchmark. Third, after constructing the dataset, we conducted human subject studies to verify its integrity by randomly selecting 100 entries from each category, ensuring the effectiveness of our adopted manipulation methods. Finally, our benchmarking included two types of human-involved experiments. The first type involves comparing the LVLM's performance to human performance; the second type entails human subject evaluation of the LVLM's performance based on its justification production.

\section{Methodology} 
\subsection{Models}
To provide an exhaustive perspective on the current state of emerging LVLMs within the context of multimodal fact-checking, we conducted comprehensive evaluations on representative accessible LVLMs. Our selection encompasses a range of models from diverse organizations, differing in size, which allows for a thorough understanding of the capabilities and limitations of LVLMs in handling multimodal content concerned with factuality.

For the open-source and accessible LVLMs, we adopt the representative models like 
Emu2~\citep{Sun2023GenerativeMM},
InternVL~\citep{Chen2023InternVLSU},
CogVLM~\citep{Wang2023CogVLMVE},
LLaVA-NeXT~\citep{ liu2024llavanext},
InstructBLIP~\citep{Dai2023InstructBLIPTG},
Pixtral\footnote{https://mistral.ai/news/pixtral-12b/},
MiniCPM-V-2.6~\citep{yao2024minicpm},
LLaVA-OneVsion~\citep{li2024llavaonevisioneasyvisualtask},
Molmo~\citep{deitke2024molmopixmoopenweights},
Qwen-VL~\cite{bai2023qwen},
Qwen2-VL~\citep{Qwen2VL},
Yi-VL~\citep{Young2024YiOF} and 
xGen-MM~\citep{xue2024xgenmmblip3familyopen}.
As five of the most powerful closed-source LVLMs, GPT-4o, GPT-4V, Claude3.5-Sonnet, Claude3-Haiku and Gemini-1.5-Pro are included in our testing scope. 


\subsection{Prompt Strategy}
We define a multimodal content \( M = \{I, C\} \) as a tuple consisting of an image \( I \) and an accompanying textual claim \( C \) to be fact-checked. 

Given that our benchmark comprises three important decomposed sub-tasks for verdict prediction in the MFC process~\cite{akhtar2023multimodal}, we have developed three task instructions \(T_i \) specifically designed to elicit the multimodal fact-checking capabilities of the LVLMs as follows:

\textbf{Manipulation Classification (Task $T_1$)}: ``\textit{Manipulation encompasses various alterations such as face swapping, face attribute editing, background changing, image generation, entity replacement, and style transfer. Your task is to determine if the image and caption have been manipulated.}''

\textbf{Out-of-Context Classification (Task $T_2$)}: ``\textit{Out-of-Context Classification is a task in which the goal is to identify whether a given image and accompanying text are contextually mismatched or falsely connected. Your task is to identify whether a given image and its accompanying text are contextually mismatched or falsely connected.}''
    
\textbf{Veracity Classification (Task $T_3$)}: ``\textit{The Veracity task in a multimodal context involves assessing the truthfulness or accuracy of textual claims by using visual evidence. Your task is to determine the truthfulness of textual claims based on the accompanying visual evidence.}''

Besides, we carefully design three questions for the three MFC sub-tasks and incorporate the image \( I \) and claim \( C \) into them, to enable the model to answer questions for verdict prediction as follows:

\textbf{Manipulation Classification (Question $Q_1$)}: ``Given a claim \{\( C \)\} and its image \{\( I \)\}, is this multimodal content manipulated?''

\textbf{Out-of-Context Classification (Question $Q_2$)}: ``Does this claim \{\( C \)\} match its image \{\( I \)\}?''

\textbf{Veracity Classification (Question $Q_3$)}: ``Based on the image \{\( I \)\}, is this claim \{\( C \)\} true?''

At the end of each prompt template, we instruct the required output format \( F\): ``Answer yes or no.''.
As demonstrated in Figure~\ref{fig:prompt-types}, to explore the effect of different prompt strategies like Chain-of-Thought (CoT)~\cite{wei2022chain} or In-Context Learning (ICL) prompting, we utilized the four following prompt methods for the \benchname{}: \textit{Zero-shot}, \textit{Zero-shot with CoT}~\cite{kojima2022large}, \textit{Few-shot}, and \textit{Few-shot with CoT}~\cite{wei2022chain}. Specifically, we design the prompt as follows:

\begin{figure*}
  \centering
    \includegraphics[width=\linewidth]{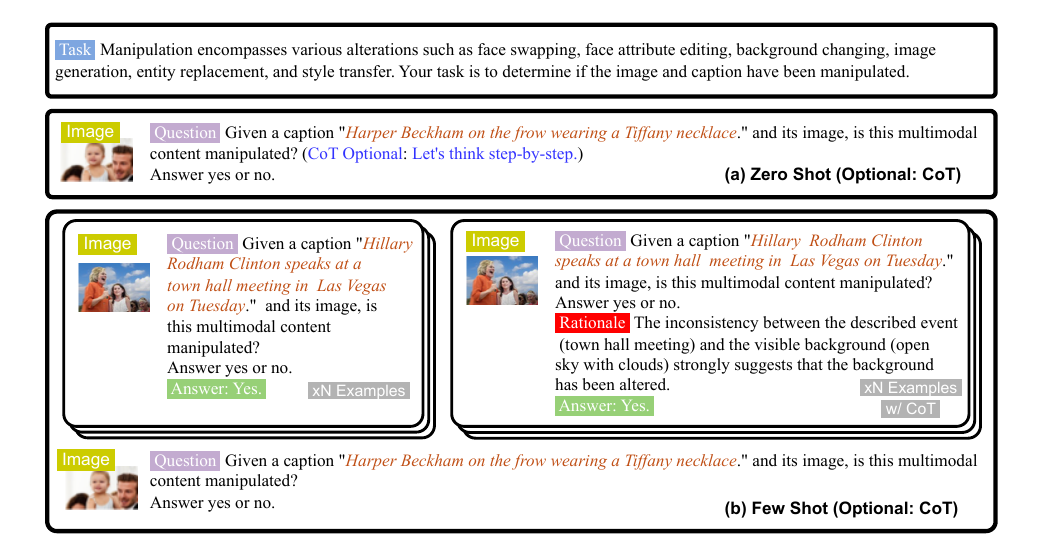}
    \vspace{-0.5cm}
  \caption{Comparison of prompts in zero-shot and few-shot scenarios with and without CoT.}
  \label{fig:prompt-types}
  \vspace{-0.4cm}
\end{figure*}

\textbf{Zero-shot Prompt.} We initially employed the zero-shot setting to activate the fact-checking capabilities of LVLMs. Given a task instruction \( T_i \), a question unit \( Q_i \),  and the return format \( F \), the LVLMs \(f(\cdot)\) are expected to determine whether the output \( Y = f(T_i,Q_i,F) \) is ``Yes'' or ``No'', as depicted in Figure~\ref{fig:prompt-types}(a).
To extend the Zero-shot with CoT setting in LLMs described in \cite{kojima2022large}, we simply incorporated the CoT prompt \( C_p \) ``Let's think step by step'' into the original prompt, to encourage the LVLMs to implicitly conduct complex reasoning by retrieving internal evidence, for determining the label \(L\). Consequently, LVLMs will process \( f( T_i,Q_i,C_p,F)\) and finally return the answer to multimodal fact verification.

\textbf{Few-shot Prompt.} Previous literature has indicated that pre-trained LLMs can significantly benefit from the inclusion of a few ICL demonstrations~\citep{brown2020language}. To assess whether the LVLMs could gain similar advantages from the in-context demonstrations in multimodal fact-checking, we employed the few-shot setting. For the Few-shot examples, we define each example \(E = \{Q_i, L\}\) consisting of a question \(Q_i\) and its corresponding factuality label \(L\) for fact verification. The inputs of LVLMs are given as \( \{T_i, E^N, Q_i, F\} \), where \(E^N\) represents multiple examples and \(N\) denotes the number of examples, as demonstrated in Figure~\ref{fig:prompt-types}(b). In terms of the Few-shot with CoT prompt, we manually curated a rationale \( R \) for each example to guide the LVLMs, where the example is represented as \(E_c = \{Q_i, R, L\}\) and the input is \( \{ T_i, E_{c}^{N}, Q_i, F\} \).

Furthermore, to gain deeper insights into the model interpretability of LVLMs, we expand our research on the evaluation of the justification production of LVLMs. The output format \( F\): ``Answer yes or no.'' was removed to allow the model to produce more intermediate reasoning steps. The model's interpretability was evaluated by GPT-4 and humans across four dimensions: Misleadingness (M), Informativeness (I), Soundness (S), and Readability (R). A 5-point Likert scale was used, where 1 indicates the lowest quality and 5 the highest for Informativeness, Soundness, and Readability, but the scale is reversed for Misleadingness.

\section{Experiments and Results}

\begin{table*}
    \centering
   \small
    \begin{tabular}{lcccccccccc}
        \toprule
        \multirow{2}{*}{\textbf{Models}} & \multirow{2}{*}{\textbf{Size}} & \multicolumn{2}{c}{\textbf{Manipulation}} & \multicolumn{2}{c}{\textbf{OOC}}& \multicolumn{2}{c}{\textbf{Veracity}} & \multicolumn{2}{c}{\textbf{Overall}}  \\
        \cmidrule(lr){3-4} \cmidrule(lr){5-6}  \cmidrule(lr){7-8} \cmidrule(lr){9-10} 
        {} & {} & \textbf{Accuracy} & \textbf{F1} & \textbf{Accuracy} & \textbf{F1} &  \textbf{Accuracy} & \textbf{F1}  &  \textbf{Accuracy} & \textbf{F1}\\
        \midrule
        
        \rowcolor{pink!50}
        \multicolumn{10}{c}{\textit{Proprietary Models}}\\
        \Openaiemoji{}~\textbf{GPT-4o} & - & \textbf{65.7} & \underline{60.4} & \textbf{84.8 }& \textbf{84.8} & \underline{80.1} & \textbf{63.0}&\textbf{67.7} & \textbf{69.4}\\
        \Openaiemoji{}~\textbf{GPT-4V} & - & 58.4 & 50.2 & 75.8 & 75.2 & 77.4& \underline{60.0} & 60.6 & 61.8\\
         \Claudeemoji{}~\textbf{Claude3.5-Sonnet} & - &59.9 & 41.7 & 49.9 & 37.6&72.7&47.4 &60.1 & 42.2\\
        \Claudeemoji{}~\textbf{Claude3-Haiku} & - & 51.4 & 37.8 & 59.8 & 59.5 & \textbf{80.3} & 57.4 & 53.7 & 51.6\\
        \Googleemoji{}~\textbf{Gemini-1.5-Pro}& - & \underline{64.2} & \textbf{61.6} & \underline{80.2} & \underline{80.1} & 79.6 & 56.6 & \underline{66.1} & \underline{66.1}\\
        \midrule
        \rowcolor{green!30}
        \multicolumn{10}{c}{\textit{Open-Source Models}}\\
        \BAAIemoji{}~\textbf{Emu2} & 37B & 38.7 & 33.0  & 51.9 & 51.1 & 70.0 & 52.6 & 41.4 & 45.6\\
        \InternVLemoji{}~\textbf{InternVL} & 25.5B & 60.1 & 44.6 & {73.4} & 73.0 & 80.0 & 57.4& 62.1 & 58.3\\
        \CogVLMemoji{}~\textbf{CogVLM} & 17B & 56.3 & 52.3& 61.4&56.2 &76.4 &63.4 & 57.8 & 57.3\\
        \llavanextmoji{}~\textbf{LLaVA-NeXT} & 13B & \textbf{62.5} & \underline{56.5} & 61.8 & 57.2 & 78.4 & 51.3& \underline{63.4} & 55.0\\
        \Instructblipemoji{}~\textbf{InstructBLIP} & 13B & 41.7 & 30.5& 59.5& 52.3 & 49.6 & 49.3& 43.3 & 44.0\\
        \pixtralemoji{}~\textbf{Pixtral} & 12B & 58.5 & 43.9 & 64.8 &63.5 & 80.9 & 65.0 & 60.2 & 57.5 \\
        \MiniCPMVemoji{}~\textbf{MiniCPM-V-2.6} & 8B& 58.9 & 39.7 & 71.2 & 71.0 & 80.4 & 65.1& 60.9 & 58.6 \\
        \llavanextmoji{}~\textbf{LLaVA-OneVision} &7B&\underline{61.5}&55.5&\underline{75.7}&\underline{75.4}&80.9&60.3 & \textbf{63.5} & \underline{63.7}\\
        \molmoemoji{}~\textbf{Molmo} & 7B & 59.3&\textbf{59.3}&58.9&52.3&79.9&57.6 & 60.5 & 56.4\\
        \Qwenemoji{}~\textbf{Qwen-VL} & 7B  & 45.7 & 45.4 & 69.7& 69.4 & \underline{82.7} & \underline{69.3} & 49.4 & 61.4\\
        \Qwenemoji{}~\textbf{Qwen2-VL} & 7B &59.9&46.6&\textbf{80.1}&\textbf{80.1}&\textbf{85.7}&\textbf{75.5}& 62.7 & \textbf{67.4}\\
        \YiVLemoji{}~\textbf{Yi-VL} & 6B & 56.4& 43.8 & 70.4 &70.4 & 78.4 & 60.0& 58.6 & 58.1\\
        \Instructblipemoji{}~\textbf{xGen-MM} & 5B&42.7&33.8&50.0&44.8&64.7&48.7& 44.5 & 42.4 \\
        \midrule
        \rowcolor{blue!30}
        \multicolumn{10}{c}{\textit{Human}}\\
        \humanemoji{}~\textbf{Human} & - & {75.7} & {75.6} & 74.0 & 73.5 & {96.0} & {91.7} & {76.8} & {80.3}\\
        \bottomrule
    \end{tabular}
    \vspace{-0.3cm}
     \caption{Results of different LVLMs on the \benchname{}, in the zero-shot setting. The accuracy and macro-averaged F1 score (\%) are reported as the metrics. The best and second test results are in bold and underlined, respectively.}
    \label{tab:result-zero-shot}
    \vspace{-0.4cm}
\end{table*}

\subsection{Experimental Setup}
\label{sec:random-seed}
We conduct extensive experiments on the \benchname{} to evaluate a total of 18 representative LVLMs: 
1)~GPT-4o; 
2)~GPT-4V;
3)~Claude3.5-Sonnet; 
4)~Claude3-Haiku;
5)~Gemini-1.5-Pro;
6)~Emu2;
7)~InternVL;
8)~CogVLM;
9)~LLaVA-NeXT;
10)~InstructBLIP;
11)~Pixtral;
12)~MiniCPM-V-2.6;
13)~LLava-OneVsion;
14)~Molmo;
15)~Qwen-VL;
16)~Qwen2-VL;
17)~Yi-VL;
18)~xGen-MM.
To ensure our results are reproducible, we set the temperature as 0 without any sampling mechanism. We also have incorporated human performance as the benchmark baseline for comparison.
We use the accuracy and macro-averaged F1 score (dominant) as the evaluation metrics. More implementation details and baseline descriptions are provided in Appendix \S\ref{sec:baselines}.

\subsection{Main Results}
In Table~\ref{tab:result-zero-shot}, we present the average outcomes of the listed 18 accessible and representative LVLMs in a zero-shot setting on the \benchname{}. From the results, we derive the following observations: 

1) For the overall performance of the LVLMs on the Manipulation Classification, the proprietary model Gemini-1.5-Pro achieves the best performance with the 61.6\% F1 score. In open-source models, Molmo performs the best, with the 59.3\% F1 score. Counterintuitively, the more powerful closed-source models, namely GPT-4V, Claude3.5-Sonnet and Claude3-Haiku, fail to produce promising results in this sub-task. 
2) None of the models exceeded the 62\% F1 score, exposing weaknesses in vision-language models for this multimodal fact-checking stage. In contrast, human performance reached over 75\%, indicating significant room for improvement in LVLMs. This discrepancy highlights that computational power alone does not ensure superior performance in Manipulation Classification.
3) In OOC Classification, GPT-4o stands out as the preeminent model with the highest 84.8\% F1 score. In terms of Veracity Classification, Qwen2-VL is distinguished by its considerable F1 score of 75.5\%. 4) Overall, we can find most of the LVLMs could achieve better performance on OOC Classification but worse on Manipulation Classification, and performance on Veracity Classification lies in the intermediate range. This pattern underscores the rational distribution of task difficulty within our proposed benchmark, \benchname{}, which comprehensively spans a spectrum from challenging to straightforward multimodal fact-checking tasks. 5) In comparison to humans, LVLMs show considerable potential for further development in addressing more complex fact-checking challenges like Manipulation Classification. Despite this, their performance is solid in simpler tasks like OOC Classification.

\begin{figure*}[htbp]
  \includegraphics[width=0.32\linewidth]{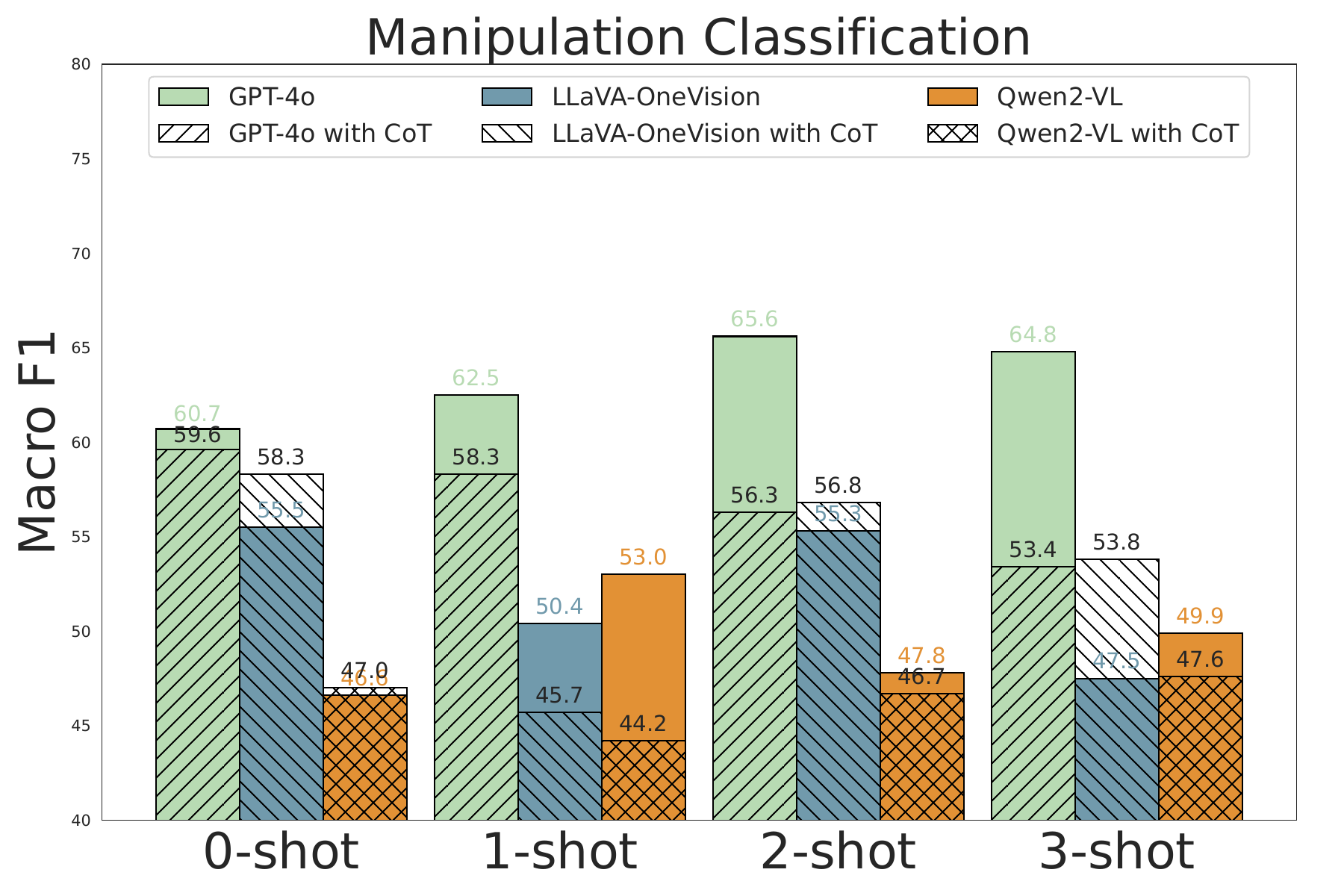} \hfill
  \includegraphics[width=0.32\linewidth]{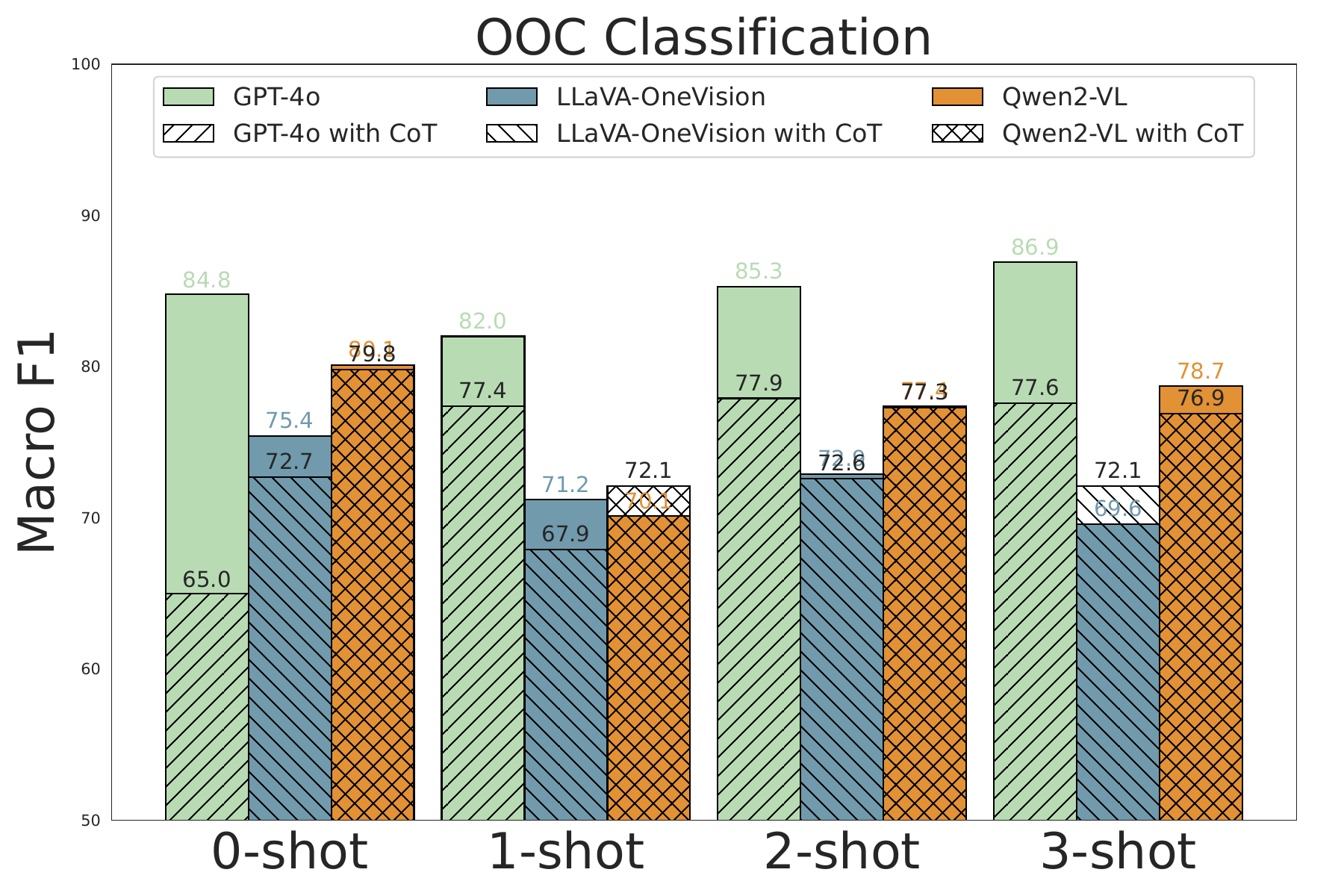} \hfill
  \includegraphics[width=0.32\linewidth]{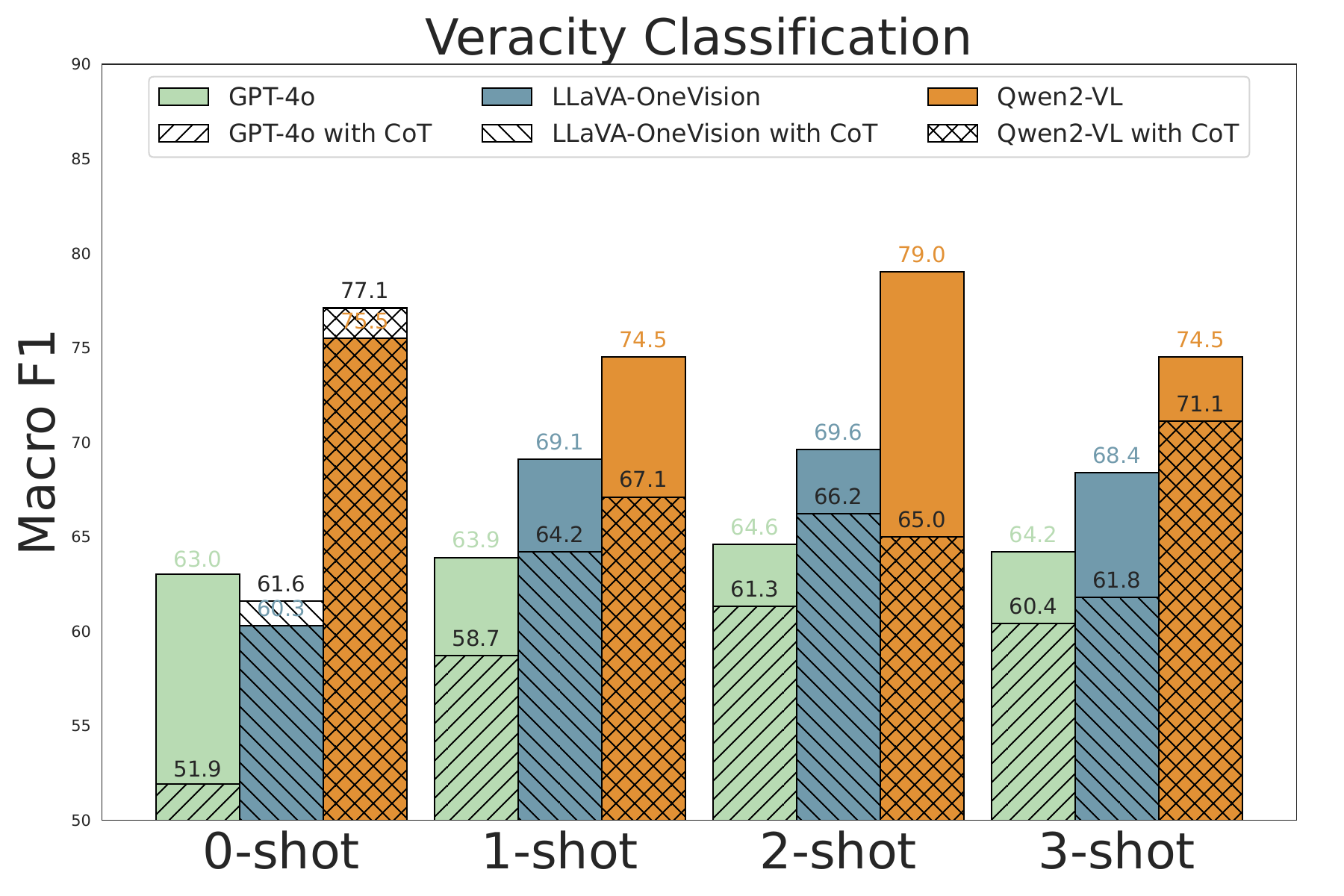} 
  \vspace{-0.3cm}
  \caption {Comparison between few-shot conditions w/ and w/o CoT for GPT-4o, LLaVA-OneVision and Qwen2-VL.}
  \label{fig:few-shot-cot}
  \vspace{-0.4cm}
\end{figure*}

\subsection{Model Interpretability}
We conducted a post-hoc interpretability analysis about Justification Production across six selected models: LLaVA-NeXT (7B\&13B), InstructBLIP (7B\&13B), Qwen-VL, and Yi-VL. This investigation explored the differences in justification production within the same model family yet varying parameter sizes, as well as the differences between distinct models.
In Table \ref{tab:model_interpretability-misr}, evaluations by GPT-4 and human evaluators show that the LLaVA-NeXT models perform exceptionally well, achieving high scores in Informativeness, Soundness, and Readability. In contrast, the InstructBLIP models struggle with interpretability. We speculate the reason is that the models are often limited to binary `yes' or `no' biased responses, and additional prompts fail to improve their explanatory capabilities.
Additionally, an increase in the size of the LVLMs, from 7 billion to 13 billion parameters, correlates with enhanced interpretability, as observed in the improved metrics for both LLaVA-NeXT and InstructBLIP families. We provide more details of human evaluation and bias in Appendix \S\ref{he}-\S\ref{sec:yes-no-bias}.

\begin{table}[t]
    \centering
    \small
    \scalebox{1}{
    \begin{tabular}{lccccc}
    \toprule
    \textbf{Models}  &  \textbf{M} & \textbf{I}  & \textbf{S} & \textbf{R} \\
    \midrule
    \rowcolor{pink!50}
    \multicolumn{5}{c}{\textit{Evaluated by GPT-4}}\\
    \llavanextmoji{}~\textbf{LLaVA-NeXT(7B)}  & 3.82 & 2.96 & 3.30 & 4.39 \\
    \llavanextmoji{}~\textbf{LLaVA-NeXT(13B)}  & 3.61 & 3.07 & 3.48 & 4.49\\
    \Instructblipemoji{}~\textbf{InstructBLIP(7B)}  & 3.41 & 1.06 & 1.63 & 2.35  \\
    \Instructblipemoji{}~\textbf{InstructBLIP(13B)}  & 3.32 & 1.16 & 1.71 & 2.46 \\
    \Qwenemoji{}~\textbf{Qwen-VL}  &3.76 & 1.77 & 2.63 & 3.68  \\
    \YiVLemoji{}~\textbf{Yi-VL}  &3.04 & 2.04 & 3.31 & 4.20  \\
    \rowcolor{blue!30}
    \multicolumn{5}{c}{\textit{Evaluated by Human}}\\
    \llavanextmoji{}~\textbf{LLaVA-NeXT(7B)} & 3.56 & 3.02 & 3.71 & 4.46    \\
    \llavanextmoji{}~\textbf{LLaVA-NeXT(13B)}  &3.68 & 3.50 & 3.77 & 4.63 \\
    \Instructblipemoji{}~\textbf{InstructBLIP(7B)} & 3.36 & 2.22 & 2.45 & 3.22  \\
    \Instructblipemoji{}~\textbf{InstructBLIP(13B)} & 3.32 & 2.21 & 2.54 & 3.51\\
    \Qwenemoji{}~\textbf{Qwen-VL} &3.61 & 2.63 & 3.11 & 3.64  \\
    \YiVLemoji{}~\textbf{Yi-VL} & 3.30 & 2.34 & 3.56 & 4.50 \\
    \bottomrule
    \end{tabular}
    }
    \vspace{-0.3cm}
    \caption{Justification Evaluated by GPT-4 and Human.}
    \label{tab:model_interpretability-misr}
    \vspace{-0.3cm}
\end{table}

\begin{table}[t!] 
    \centering
    \Large
    \scalebox{0.55}{
    \begin{tabular}{lcccccc}
        \toprule
        \multirow{2}{*}{\textbf{Models}}  & \multicolumn{2}{c}{\textbf{Manipulation}} & \multicolumn{2}{c}{\textbf{OOC}}& \multicolumn{2}{c}{\textbf{Veracity}} \\
        \cmidrule(lr){2-3} \cmidrule(lr){4-5}  \cmidrule(lr){6-7} 
        {}  & \textbf{Acc.} & \textbf{F1} & \textbf{Acc.} & \textbf{F1} &  \textbf{Acc.} & \textbf{F1} \\
        \midrule
        \rowcolor{pink!50}
        \multicolumn{7}{c}{\textit{Proprietary Models}}\\
        \Openaiemoji{}~\textbf{GPT-4o}  & 65.8 & 59.6 & 67.6 & 65.0 & 77.6 & 51.9\\
        \rowcolor{green!30}
        \multicolumn{7}{c}{\textit{Open-Source Models}}\\
        \llavanextmoji{}~\textbf{LLaVA-NeXT}   & 58.1 & 55.1 & 52.4 & 39.1 &  77.2 & 46.2  \\
        \Instructblipemoji{}~\textbf{InstructBLIP}  & 41.9 & 31.0 & 57.0 & 47.6 & 37.2 & 36.9 \\
        \llavanextmoji{}~\textbf{LLaVA-OneVision} & 61.2 & 58.3 & 73.3 & 72.7 & 81.3& 61.6 \\
        \Qwenemoji{}~\textbf{Qwen-VL}   & 45.7 & 45.2 & 71.9 & 71.8 & 81.8 & 65.3  \\
        \Qwenemoji{}~\textbf{Qwen2-VL}   & 59.3 & 47.0 & 79.8 & 79.8 & 86.6 & 77.1  \\
        \YiVLemoji{}~\textbf{Yi-VL}  & 59.9 & 42.5 & 69.4 & 69.3& 78.0& 56.1 \\
        \bottomrule
    \end{tabular}
    }
    \vspace{-0.3cm}
    \caption{Results of selected emerging LVLMs on the \benchname{} with the zero-shot CoT setting.}
    \label{tab:result-zero-shot-cot}
    \vspace{-0.4cm}
\end{table}

\subsection{Effect of CoT}
The comparison between Table~\ref{tab:result-zero-shot} and Table~\ref{tab:result-zero-shot-cot} shows that the impact of CoT in the zero-shot setting varies across different selected representative LMMs on \benchname{}. For Manipulation Classification, the impact of CoT on model performance differs, as seen in GPT-4o, where the F1 score decrease from 60.4\% to 59.6\%, and in LLaVA-OneVision, where it rose from 55.5\% to 58.3\% 
In the case of OOC Classification, CoT proves beneficial for some LVLMs, such as Qwen-VL, while it negatively affects others, like Qwen2-VL. For Veracity Classification, CoT generally does not significantly impact performance and may even reduce it for certain models. In few-shot settings, as shown in Figure~\ref{fig:few-shot-cot}, CoT does not enhance the performance of LLaVA-OneVision and Qwen2-VL. For LLaVA-OneVision, CoT has a minimal to slightly positive impact on performance in Manipulation Classification and a somewhat negative impact in Veracity Classification. Conversely, the effect of CoT on the GPT-4o is continuously negative. The possible reasons for these observations include the underdeveloped ability of the LVLM to handle multiple image inputs and the excessive length of the rationale, which diminishes the model's ability to understand the task effectively.

\subsection{Effect of ICL}
To thoroughly investigate the impact of In-Context Learning (ICL) on model performance, we selected GPT-4o, Qwen2-VL and LLaVA-OneVision that support multiple image inputs to conduct few-shot experiments. 
We calculated the macro-averaged F1 scores as the evaluation metric. 1) The results, as illustrated in Figure~\ref{fig:few-shot-cot}, indicate that the implementation of few-shot learning does not enhance the fact-checking capabilities of these models. 2) For the performance of Qwen2-VL in Figure~\ref{fig:few-shot-cot}, the few-shot prompt (i.e., ICL) did not result in a performance improvement. Instead, we found that it induced model inertia, leading it to predominantly respond with ``no'' in most instances.   We provide more qualitative analysis in Appendix \S\ref{analysis}.

\section{Conclusion and Future Work}
In this study, we aim to investigate the trustworthy insight of LVLMs by examining the multimodal fact-checking ability of LVLMs across a spectrum of data categories. For this purpose, we have developed the \benchname{}, a comprehensive testbed consisting of 35K multimodal samples, spanning three tasks of varied complexity. Our evaluation of various LVLMs using different prompting methods, including those with CoT or ICL prompts, on the \benchname{} reveals that these models still exhibit limitations in accurately addressing multimodal fact-checking tasks. In our future work, we plan to systematically study justification production for multimodal fact-checking with LVLMs.

\section*{Limitations}
\label{sec:limitation}
As this is the first benchmark work to evaluate the multimodal fact-checking capacity of LVLMs, there are no doubt multiple efforts needed to improve the work in the future:
\begin{itemize}
    \item The dynamic and context-specific nature of multimodal fact-checking presents a challenge in interpretation and analysis. The current benchmark may not fully capture this complexity, potentially limiting the generalizability of our findings. Human interpretation of multimodal disinformation is inherently intricate and contextual. Real-world data from diverse domains will help advance this benchmark into various use case applications. Adding temporal dynamics will provide value when fact-checking historical facts. Additionally, future studies could be enhanced by a more comprehensive examination of bias and fairness in model evaluations to prevent the reinforcement or exacerbation of stereotypical hallucinations.
    \item While this pioneering work delivers comprehensive results related to multimodal fact-checking, further improving the interpretability of these findings could provide deeper, more actionable insights for practical applications and further development of models. Delving into the underlying reasons for the fact-checking outcomes observed in LVLMs and discussing these in detail would not only shed light on model behaviors but also suggest avenues for optimization. Expanding on how these results can be translated into model enhancements and identifying specific aspects that could benefit from refinement would make the findings more applicable. Additionally, exploring how these interpretations align with real-world multimodal data usage could guide future research directions, fostering advancements in both theoretical and applied domains of multimodal fact-checking.
    \item During the benchmarking process, we not only explore the three stages of verdict prediction for MFC: Manipulation Classification, OOC Classification, and Veracity Classification, but also investigate the last stage: Justification Production that requires the selected models to provide the post-hoc explanations. However, there might be a deeper of model interpretability that is not touched in this work, which is to explain how an LVLM works internally. In future work, we should investigate the model's internal reasoning mechanisms and how it arrives at its conclusions from the perspective of the model architecture. Furthermore, the current LVLM demonstrates grounding capabilities that can be leveraged to better understand the model's interpretation of images and its fact-checking judgments.
    \item Expanding the scope to include a broader array of models could enhance the robustness and applicability of the results. Incorporating diverse multilingual datasets, the audio modality, and emerging LVLMs into our benchmark work could provide a more nuanced understanding of LVLMs' capabilities across various languages. Although there is a long way to go, where there is a will, there is a way.
\end{itemize}

\section*{Ethics Statement}
\label{sec:ethics}
The aim of this research is to focus on the multimodal fact-checking issue related to LVLMs, to curb the dissemination of multimodal disinformation, and to protect individuals from exposure to fake news. However, we acknowledge the risk that malicious actors might attempt to reverse-engineer misinformation that could evade detection by AI systems trained on LVLMs. We vehemently discourage and denounce such practices, and emphasize that human moderation is essential to prevent such occurrences. Our utilization of data adheres to the terms of the datasets~\cite{Shao_2023_CVPR, luo-etal-2021-newsclippings, 10.1145/3539618.3591879}. All the data in this work only includes text and image modalities and does not contain any user information on social media.

To protect our human evaluators, we establish three guidelines: 1) ensuring their acknowledgment of viewing potentially uncomfortable content, 2) limiting weekly evaluations and encouraging a lighter daily workload, and 3) advising them to stop if they feel overwhelmed. Finally, we regularly check in with evaluators to ensure their well-being.


\bibliography{custom}

\begin{thebibliography}{108}
\providecommand{\natexlab}[1]{#1}

\bibitem[{Agarwal et~al.(2019)Agarwal, Farid, Gu, He, Nagano, and Li}]{Agarwal2019ProtectingWL}
Shruti Agarwal, Hany Farid, Yuming Gu, Mingming He, Koki Nagano, and Hao Li. 2019.
\newblock \href {https://api.semanticscholar.org/CorpusID:195732375} {Protecting world leaders against deep fakes}.
\newblock In \emph{CVPR Workshops}.

\bibitem[{Agrawal et~al.(2019)Agrawal, Desai, Wang, Chen, Jain, Johnson, Batra, Parikh, Lee, and Anderson}]{agrawal2019nocaps}
Harsh Agrawal, Karan Desai, Yufei Wang, Xinlei Chen, Rishabh Jain, Mark Johnson, Dhruv Batra, Devi Parikh, Stefan Lee, and Peter Anderson. 2019.
\newblock Nocaps: Novel object captioning at scale.
\newblock In \emph{Proceedings of the IEEE/CVF international conference on computer vision}, pages 8948--8957.

\bibitem[{Akhtar et~al.(2023)Akhtar, Schlichtkrull, Guo, Cocarascu, Simperl, and Vlachos}]{akhtar2023multimodal}
Mubashara Akhtar, Michael Schlichtkrull, Zhijiang Guo, Oana Cocarascu, Elena Simperl, and Andreas Vlachos. 2023.
\newblock Multimodal automated fact-checking: A survey.
\newblock In \emph{Findings of the Association for Computational Linguistics: EMNLP 2023}, pages 5430--5448.

\bibitem[{Alayrac et~al.(2022)Alayrac, Donahue, Luc, Miech, Barr, Hasson, Lenc, Mensch, Millican, Reynolds et~al.}]{alayrac2022flamingo}
Jean-Baptiste Alayrac, Jeff Donahue, Pauline Luc, Antoine Miech, Iain Barr, Yana Hasson, Karel Lenc, Arthur Mensch, Katherine Millican, Malcolm Reynolds, et~al. 2022.
\newblock Flamingo: a visual language model for few-shot learning.
\newblock In \emph{Advances in Neural Information Processing Systems}.

\bibitem[{Aneja et~al.(2021)Aneja, Bregler, and Nie{\ss}ner}]{aneja2021cosmos}
Shivangi Aneja, Chris Bregler, and Matthias Nie{\ss}ner. 2021.
\newblock {COSMOS}: Catching {O}ut-of-{C}ontext {M}isinformation with {S}elf-{S}upervised {L}earning.
\newblock In \emph{ArXiv preprint arXiv:2101.06278}.

\bibitem[{Aneja et~al.(2023)Aneja, Bregler, and Nie\ss{}ner}]{10.1609/aaai.v37i12.26648}
Shivangi Aneja, Chris Bregler, and Matthias Nie\ss{}ner. 2023.
\newblock \href {https://doi.org/10.1609/aaai.v37i12.26648} {Cosmos: catching out-of-context image misuse with self-supervised learning}.
\newblock In \emph{Proceedings of the Thirty-Seventh AAAI Conference on Artificial Intelligence and Thirty-Fifth Conference on Innovative Applications of Artificial Intelligence and Thirteenth Symposium on Educational Advances in Artificial Intelligence}, AAAI'23/IAAI'23/EAAI'23. AAAI Press.

\bibitem[{Bai et~al.(2023)Bai, Bai, Yang, Wang, Tan, Wang, Lin, Zhou, and Zhou}]{bai2023qwen}
Jinze Bai, Shuai Bai, Shusheng Yang, Shijie Wang, Sinan Tan, Peng Wang, Junyang Lin, Chang Zhou, and Jingren Zhou. 2023.
\newblock Qwen-vl: A frontier large vision-language model with versatile abilities.
\newblock \emph{arXiv preprint arXiv:2308.12966}.

\bibitem[{Bavishi et~al.(2023)Bavishi, Elsen, Hawthorne, Nye, Odena, Somani, and Ta\c{s}\i{}rlar}]{fuyu-8b}
Rohan Bavishi, Erich Elsen, Curtis Hawthorne, Maxwell Nye, Augustus Odena, Arushi Somani, and Sa\u{g}nak Ta\c{s}\i{}rlar. 2023.
\newblock \href {https://www.adept.ai/blog/fuyu-8b} {Introducing our multimodal models}.

\bibitem[{Brown et~al.(2020)Brown, Mann, Ryder, Subbiah, Kaplan, Dhariwal, Neelakantan, Shyam, Sastry, Askell et~al.}]{brown2020language}
Tom~B Brown, Benjamin Mann, Nick Ryder, Melanie Subbiah, Jared Kaplan, Prafulla Dhariwal, Arvind Neelakantan, Pranav Shyam, Girish Sastry, Amanda Askell, et~al. 2020.
\newblock Language models are few-shot learners.
\newblock In \emph{Proceedings of the 34th International Conference on Neural Information Processing Systems}, pages 1877--1901.

\bibitem[{Cao et~al.(2021)Cao, Lin, Han, Sun, Yan, Liao, Xue, and Xu}]{cao2021knowledgeable}
Boxi Cao, Hongyu Lin, Xianpei Han, Le~Sun, Lingyong Yan, Meng Liao, Tong Xue, and Jin Xu. 2021.
\newblock Knowledgeable or educated guess? revisiting language models as knowledge bases.
\newblock In \emph{Proceedings of the 59th Annual Meeting of the Association for Computational Linguistics and the 11th International Joint Conference on Natural Language Processing (Volume 1: Long Papers)}, pages 1860--1874.

\bibitem[{Chang et~al.(2023)Chang, Wang, Wang, Wu, Yang, Zhu, Chen, Yi, Wang, Wang et~al.}]{chang2023survey}
Yupeng Chang, Xu~Wang, Jindong Wang, Yuan Wu, Linyi Yang, Kaijie Zhu, Hao Chen, Xiaoyuan Yi, Cunxiang Wang, Yidong Wang, et~al. 2023.
\newblock A survey on evaluation of large language models.
\newblock \emph{ACM Transactions on Intelligent Systems and Technology}.

\bibitem[{Chang et~al.(2024)Chang, Wang, Wang, Wu, Yang, Zhu, Chen, Yi, Wang, Wang et~al.}]{chang2024survey}
Yupeng Chang, Xu~Wang, Jindong Wang, Yuan Wu, Linyi Yang, Kaijie Zhu, Hao Chen, Xiaoyuan Yi, Cunxiang Wang, Yidong Wang, et~al. 2024.
\newblock A survey on evaluation of large language models.
\newblock \emph{ACM Transactions on Intelligent Systems and Technology}, 15(3):1--45.

\bibitem[{Chen et~al.(2023{\natexlab{a}})Chen, Zhang, Han, Chen, Shi, Xu, and Xu}]{chen2023vlp}
Fei-Long Chen, Du-Zhen Zhang, Ming-Lun Han, Xiu-Yi Chen, Jing Shi, Shuang Xu, and Bo~Xu. 2023{\natexlab{a}}.
\newblock Vlp: A survey on vision-language pre-training.
\newblock \emph{Machine Intelligence Research}, 20(1):38--56.

\bibitem[{Chen et~al.(2023{\natexlab{b}})Chen, Zhu, Shen, Li, Liu, Zhang, Krishnamoorthi, Chandra, Xiong, and Elhoseiny}]{Chen2023MiniGPTv2LL}
Jun Chen, Deyao Zhu, Xiaoqian Shen, Xiang Li, Zechun Liu, Pengchuan Zhang, Raghuraman Krishnamoorthi, Vikas Chandra, Yunyang Xiong, and Mohamed Elhoseiny. 2023{\natexlab{b}}.
\newblock \href {https://api.semanticscholar.org/CorpusID:264146906} {Minigpt-v2: large language model as a unified interface for vision-language multi-task learning}.
\newblock \emph{ArXiv}, abs/2310.09478.

\bibitem[{Chen et~al.(2023{\natexlab{c}})Chen, Wu, Wang, Su, Chen, Xing, Muyan, Zhang, Zhu, Lu, Li, Luo, Lu, Qiao, and Dai}]{Chen2023InternVLSU}
Zhe Chen, Jiannan Wu, Wenhai Wang, Weijie Su, Guo Chen, Sen Xing, Zhong Muyan, Qinglong Zhang, Xizhou Zhu, Lewei Lu, Bin Li, Ping Luo, Tong Lu, Yu~Qiao, and Jifeng Dai. 2023{\natexlab{c}}.
\newblock \href {https://api.semanticscholar.org/CorpusID:266521410} {Internvl: Scaling up vision foundation models and aligning for generic visual-linguistic tasks}.
\newblock \emph{ArXiv}, abs/2312.14238.

\bibitem[{Chiang et~al.(2023)Chiang, Li, Lin, Sheng, Wu, Zhang, Zheng, Zhuang, Zhuang, Gonzalez et~al.}]{chiang2023vicuna}
Wei-Lin Chiang, Zhuohan Li, Zi~Lin, Ying Sheng, Zhanghao Wu, Hao Zhang, Lianmin Zheng, Siyuan Zhuang, Yonghao Zhuang, Joseph~E Gonzalez, et~al. 2023.
\newblock Vicuna: An open-source chatbot impressing gpt-4 with 90\%* chatgpt quality.
\newblock \emph{See https://vicuna. lmsys. org (accessed 14 April 2023)}.

\bibitem[{Chowdhery et~al.(2022)Chowdhery, Narang, Devlin, Bosma, Mishra, Roberts, Barham, Chung, Sutton, Gehrmann et~al.}]{chowdhery2022palm}
Aakanksha Chowdhery, Sharan Narang, Jacob Devlin, Maarten Bosma, Gaurav Mishra, Adam Roberts, Paul Barham, Hyung~Won Chung, Charles Sutton, Sebastian Gehrmann, et~al. 2022.
\newblock Palm: Scaling language modeling with pathways.
\newblock \emph{arXiv preprint arXiv:2204.02311}.

\bibitem[{Da et~al.(2021)Da, Forbes, Zellers, Zheng, Hwang, Bosselut, and Choi}]{da-etal-2021-edited}
Jeff Da, Maxwell Forbes, Rowan Zellers, Anthony Zheng, Jena~D. Hwang, Antoine Bosselut, and Yejin Choi. 2021.
\newblock \href {https://doi.org/10.18653/v1/2021.acl-long.158} {Edited media understanding frames: Reasoning about the intent and implications of visual misinformation}.
\newblock In \emph{Proceedings of the 59th Annual Meeting of the Association for Computational Linguistics and the 11th International Joint Conference on Natural Language Processing (Volume 1: Long Papers)}, pages 2026--2039, Online. Association for Computational Linguistics.

\bibitem[{Dai et~al.(2023)Dai, Li, Li, Tiong, Zhao, Wang, Li, Fung, and Hoi}]{Dai2023InstructBLIPTG}
Wenliang Dai, Junnan Li, Dongxu Li, Anthony Meng~Huat Tiong, Junqi Zhao, Weisheng Wang, Boyang~Albert Li, Pascale Fung, and Steven C.~H. Hoi. 2023.
\newblock \href {https://api.semanticscholar.org/CorpusID:258615266} {Instructblip: Towards general-purpose vision-language models with instruction tuning}.
\newblock \emph{ArXiv}, abs/2305.06500.

\bibitem[{Deitke et~al.(2024)Deitke, Clark, Lee, Tripathi, Yang, Park, Salehi, Muennighoff, Lo, Soldaini, Lu, Anderson, Bransom, Ehsani, Ngo, Chen, Patel, Yatskar, Callison-Burch, Head, Hendrix, Bastani, VanderBilt, Lambert, Chou, Chheda, Sparks, Skjonsberg, Schmitz, Sarnat, Bischoff, Walsh, Newell, Wolters, Gupta, Zeng, Borchardt, Groeneveld, Dumas, Nam, Lebrecht, Wittlif, Schoenick, Michel, Krishna, Weihs, Smith, Hajishirzi, Girshick, Farhadi, and Kembhavi}]{deitke2024molmopixmoopenweights}
Matt Deitke, Christopher Clark, Sangho Lee, Rohun Tripathi, Yue Yang, Jae~Sung Park, Mohammadreza Salehi, Niklas Muennighoff, Kyle Lo, Luca Soldaini, Jiasen Lu, Taira Anderson, Erin Bransom, Kiana Ehsani, Huong Ngo, YenSung Chen, Ajay Patel, Mark Yatskar, Chris Callison-Burch, Andrew Head, Rose Hendrix, Favyen Bastani, Eli VanderBilt, Nathan Lambert, Yvonne Chou, Arnavi Chheda, Jenna Sparks, Sam Skjonsberg, Michael Schmitz, Aaron Sarnat, Byron Bischoff, Pete Walsh, Chris Newell, Piper Wolters, Tanmay Gupta, Kuo-Hao Zeng, Jon Borchardt, Dirk Groeneveld, Jen Dumas, Crystal Nam, Sophie Lebrecht, Caitlin Wittlif, Carissa Schoenick, Oscar Michel, Ranjay Krishna, Luca Weihs, Noah~A. Smith, Hannaneh Hajishirzi, Ross Girshick, Ali Farhadi, and Aniruddha Kembhavi. 2024.
\newblock \href {https://arxiv.org/abs/2409.17146} {Molmo and pixmo: Open weights and open data for state-of-the-art multimodal models}.
\newblock \emph{Preprint}, arXiv:2409.17146.

\bibitem[{Dolhansky et~al.(2019)Dolhansky, Howes, Pflaum, Baram, and Canton-Ferrer}]{Dolhansky2019TheDD}
Brian Dolhansky, Russ Howes, Ben Pflaum, Nicole Baram, and Cristian Canton-Ferrer. 2019.
\newblock \href {https://api.semanticscholar.org/CorpusID:204800939} {The deepfake detection challenge (dfdc) preview dataset}.
\newblock \emph{ArXiv}, abs/1910.08854.

\bibitem[{Driess et~al.(2023)Driess, Xia, Sajjadi, Lynch, Chowdhery, Ichter, Wahid, Tompson, Vuong, Yu et~al.}]{driess2023palm}
Danny Driess, Fei Xia, Mehdi~SM Sajjadi, Corey Lynch, Aakanksha Chowdhery, Brian Ichter, Ayzaan Wahid, Jonathan Tompson, Quan Vuong, Tianhe Yu, et~al. 2023.
\newblock Palm-e: An embodied multimodal language model.
\newblock \emph{arXiv preprint arXiv:2303.03378}.

\bibitem[{Elazar et~al.(2021)Elazar, Kassner, Ravfogel, Ravichander, Hovy, Sch{\"u}tze, and Goldberg}]{elazar2021measuring}
Yanai Elazar, Nora Kassner, Shauli Ravfogel, Abhilasha Ravichander, Eduard Hovy, Hinrich Sch{\"u}tze, and Yoav Goldberg. 2021.
\newblock Measuring and improving consistency in pretrained language models.
\newblock \emph{Transactions of the Association for Computational Linguistics}, 9:1012--1031.

\bibitem[{Fawcett(2006)}]{fawcett2006introduction}
Tom Fawcett. 2006.
\newblock An introduction to roc analysis.
\newblock \emph{Pattern recognition letters}, 27(8):861--874.

\bibitem[{Fu et~al.(2023)Fu, Chen, Shen, Qin, Zhang, Lin, Qiu, Lin, Yang, Zheng et~al.}]{fu2023mme}
Chaoyou Fu, Peixian Chen, Yunhang Shen, Yulei Qin, Mengdan Zhang, Xu~Lin, Zhenyu Qiu, Wei Lin, Jinrui Yang, Xiawu Zheng, et~al. 2023.
\newblock Mme: A comprehensive evaluation benchmark for multimodal large language models.
\newblock \emph{arXiv preprint arXiv:2306.13394}.

\bibitem[{Gong et~al.(2023)Gong, Lyu, Zhang, Wang, Zheng, Zhao, Liu, Zhang, Luo, and Chen}]{gong2023multimodal}
Tao Gong, Chengqi Lyu, Shilong Zhang, Yudong Wang, Miao Zheng, Qian Zhao, Kuikun Liu, Wenwei Zhang, Ping Luo, and Kai Chen. 2023.
\newblock Multimodal-gpt: A vision and language model for dialogue with humans.
\newblock \emph{arXiv preprint arXiv:2305.04790}.

\bibitem[{Goyal et~al.(2017)Goyal, Khot, Summers-Stay, Batra, and Parikh}]{goyal2017making}
Yash Goyal, Tejas Khot, Douglas Summers-Stay, Dhruv Batra, and Devi Parikh. 2017.
\newblock Making the v in vqa matter: Elevating the role of image understanding in visual question answering.
\newblock In \emph{Proceedings of the IEEE conference on computer vision and pattern recognition}, pages 6904--6913.

\bibitem[{Guo et~al.(2022)Guo, Schlichtkrull, and Vlachos}]{guo2022survey}
Zhijiang Guo, Michael Schlichtkrull, and Andreas Vlachos. 2022.
\newblock A survey on automated fact-checking.
\newblock \emph{Transactions of the Association for Computational Linguistics}, 10:178--206.

\bibitem[{Heinzerling and Inui(2021)}]{heinzerling2021language}
Benjamin Heinzerling and Kentaro Inui. 2021.
\newblock Language models as knowledge bases: On entity representations, storage capacity, and paraphrased queries.
\newblock In \emph{Proceedings of the 16th Conference of the European Chapter of the Association for Computational Linguistics: Main Volume}, pages 1772--1791.

\bibitem[{Hu et~al.(2024)Hu, Chen, Li, Guo, Wen, Philip, and Guo}]{hu2024large}
Xuming Hu, Junzhe Chen, Xiaochuan Li, Yufei Guo, Lijie Wen, S~Yu Philip, and Zhijiang Guo. 2024.
\newblock Do large language models know about facts?
\newblock In \emph{The Twelfth International Conference on Learning Representations}.

\bibitem[{Jaiswal et~al.(2017)Jaiswal, Sabir, AbdAlmageed, and Natarajan}]{10.1145/3123266.3123385}
Ayush Jaiswal, Ekraam Sabir, Wael AbdAlmageed, and Premkumar Natarajan. 2017.
\newblock \href {https://doi.org/10.1145/3123266.3123385} {Multimedia semantic integrity assessment using joint embedding of images and text}.
\newblock In \emph{Proceedings of the 25th ACM International Conference on Multimedia}, MM '17, page 1465–1471, New York, NY, USA. Association for Computing Machinery.

\bibitem[{Jiang et~al.(2023)Jiang, Sablayrolles, Mensch, Bamford, Chaplot, de~Las~Casas, Bressand, Lengyel, Lample, Saulnier, Lavaud, Lachaux, Stock, Scao, Lavril, Wang, Lacroix, and Sayed}]{mistral}
Albert~Q. Jiang, Alexandre Sablayrolles, Arthur Mensch, Chris Bamford, Devendra~Singh Chaplot, Diego de~Las~Casas, Florian Bressand, Gianna Lengyel, Guillaume Lample, Lucile Saulnier, L{\'{e}}lio~Renard Lavaud, Marie{-}Anne Lachaux, Pierre Stock, Teven~Le Scao, Thibaut Lavril, Thomas Wang, Timoth{\'{e}}e Lacroix, and William~El Sayed. 2023.
\newblock \href {https://doi.org/10.48550/ARXIV.2310.06825} {Mistral 7b}.
\newblock \emph{CoRR}, abs/2310.06825.

\bibitem[{Jiang et~al.(2020)Jiang, Anastasopoulos, Araki, Ding, and Neubig}]{jiang2020x}
Zhengbao Jiang, Antonios Anastasopoulos, Jun Araki, Haibo Ding, and Graham Neubig. 2020.
\newblock X-factr: Multilingual factual knowledge retrieval from pretrained language models.
\newblock In \emph{Proceedings of the 2020 Conference on Empirical Methods in Natural Language Processing (EMNLP)}, pages 5943--5959.

\bibitem[{Kadavath et~al.(2022)Kadavath, Conerly, Askell, Henighan, Drain, Perez, Schiefer, Hatfield-Dodds, DasSarma, Tran-Johnson et~al.}]{kadavath2022language}
Saurav Kadavath, Tom Conerly, Amanda Askell, Tom Henighan, Dawn Drain, Ethan Perez, Nicholas Schiefer, Zac Hatfield-Dodds, Nova DasSarma, Eli Tran-Johnson, et~al. 2022.
\newblock Language models (mostly) know what they know.
\newblock \emph{arXiv preprint arXiv:2207.05221}.

\bibitem[{Kojima et~al.(2022)Kojima, Gu, Reid, Matsuo, and Iwasawa}]{kojima2022large}
Takeshi Kojima, Shixiang~Shane Gu, Machel Reid, Yutaka Matsuo, and Yusuke Iwasawa. 2022.
\newblock Large language models are zero-shot reasoners.
\newblock In \emph{Advances in Neural Information Processing Systems}.

\bibitem[{Lample et~al.(2016)Lample, Ballesteros, Subramanian, Kawakami, and Dyer}]{lample2016neural}
Guillaume Lample, Miguel Ballesteros, Sandeep Subramanian, Kazuya Kawakami, and Chris Dyer. 2016.
\newblock Neural architectures for named entity recognition.
\newblock In \emph{Proceedings of the 2016 Conference of the North American Chapter of the Association for Computational Linguistics: Human Language Technologies}, pages 260--270.

\bibitem[{Li et~al.(2024{\natexlab{a}})Li, Zhang, Guo, Zhang, Li, Zhang, Zhang, Li, Liu, and Li}]{li2024llavaonevisioneasyvisualtask}
Bo~Li, Yuanhan Zhang, Dong Guo, Renrui Zhang, Feng Li, Hao Zhang, Kaichen Zhang, Yanwei Li, Ziwei Liu, and Chunyuan Li. 2024{\natexlab{a}}.
\newblock \href {https://arxiv.org/abs/2408.03326} {Llava-onevision: Easy visual task transfer}.
\newblock \emph{Preprint}, arXiv:2408.03326.

\bibitem[{Li et~al.(2023{\natexlab{a}})Li, Wang, Wang, Ge, Ge, and Shan}]{li2023seed}
Bohao Li, Rui Wang, Guangzhi Wang, Yuying Ge, Yixiao Ge, and Ying Shan. 2023{\natexlab{a}}.
\newblock Seed-bench: Benchmarking multimodal llms with generative comprehension.
\newblock \emph{arXiv preprint arXiv:2307.16125}.

\bibitem[{Li et~al.(2024{\natexlab{b}})Li, Tian, Hu, Luo, and Ma}]{li2024mmcode}
Kaixin Li, Yuchen Tian, Qisheng Hu, Ziyang Luo, and Jing Ma. 2024{\natexlab{b}}.
\newblock \href {https://arxiv.org/abs/2404.09486} {Mmcode: Evaluating multi-modal code large language models with visually rich programming problems}.
\newblock \emph{Preprint}, arXiv:2404.09486.

\bibitem[{Li et~al.(2023{\natexlab{b}})Li, Yu, Zhou, Schick, Zettlemoyer, Levy, Weston, and Lewis}]{li2023selfalignment}
Xian Li, Ping Yu, Chunting Zhou, Timo Schick, Luke Zettlemoyer, Omer Levy, Jason Weston, and Mike Lewis. 2023{\natexlab{b}}.
\newblock \href {https://arxiv.org/abs/2308.06259} {Self-alignment with instruction backtranslation}.
\newblock \emph{Preprint}, arXiv:2308.06259.

\bibitem[{Li and Xie(2020)}]{li2020picture}
Yiyi Li and Ying Xie. 2020.
\newblock Is a picture worth a thousand words? an empirical study of image content and social media engagement.
\newblock \emph{Journal of marketing research}, 57(1):1--19.

\bibitem[{Lin et~al.(2022{\natexlab{a}})Lin, Chen, Ma, Yang, and Chen}]{lin2022amif}
Hongzhan Lin, Liangliang Chen, Jing Ma, Zhiwei Yang, and Guang Chen. 2022{\natexlab{a}}.
\newblock Amif: A hybrid model for improving fact checking in product question answering.
\newblock In \emph{2022 International Joint Conference on Neural Networks (IJCNN)}, pages 1--8. IEEE.

\bibitem[{Lin et~al.(2024{\natexlab{a}})Lin, Luo, Gao, Ma, Wang, and Yang}]{lin2024towards}
Hongzhan Lin, Ziyang Luo, Wei Gao, Jing Ma, Bo~Wang, and Ruichao Yang. 2024{\natexlab{a}}.
\newblock Towards explainable harmful meme detection through multimodal debate between large language models.
\newblock In \emph{Proceedings of the ACM on Web Conference 2024}, pages 2359--2370.

\bibitem[{Lin et~al.(2023{\natexlab{a}})Lin, Luo, Ma, and Chen}]{lin2023beneath}
Hongzhan Lin, Ziyang Luo, Jing Ma, and Long Chen. 2023{\natexlab{a}}.
\newblock Beneath the surface: Unveiling harmful memes with multimodal reasoning distilled from large language models.
\newblock In \emph{The 2023 Conference on Empirical Methods in Natural Language Processing}.

\bibitem[{Lin et~al.(2024{\natexlab{b}})Lin, Luo, Wang, Yang, and Ma}]{lin2024goat}
Hongzhan Lin, Ziyang Luo, Bo~Wang, Ruichao Yang, and Jing Ma. 2024{\natexlab{b}}.
\newblock Goat-bench: Safety insights to large multimodal models through meme-based social abuse.
\newblock \emph{arXiv preprint arXiv:2401.01523}.

\bibitem[{Lin et~al.(2021)Lin, Ma, Cheng, Yang, Chen, and Chen}]{lin2021rumor}
Hongzhan Lin, Jing Ma, Mingfei Cheng, Zhiwei Yang, Liangliang Chen, and Guang Chen. 2021.
\newblock Rumor detection on twitter with claim-guided hierarchical graph attention networks.
\newblock In \emph{Proceedings of the 2021 Conference on Empirical Methods in Natural Language Processing}, pages 10035--10047.

\bibitem[{Lin et~al.(2024{\natexlab{c}})Lin, Yang, Luo, and Ma}]{lin2024unleashing}
Hongzhan Lin, Haiqin Yang, Ziyang Luo, and Jing Ma. 2024{\natexlab{c}}.
\newblock Unleashing trigger-free event detection: Revealing event correlations via a contrastive derangement framework.
\newblock In \emph{ICASSP 2024-2024 IEEE International Conference on Acoustics, Speech and Signal Processing (ICASSP)}, pages 10171--10175. IEEE.

\bibitem[{Lin et~al.(2023{\natexlab{b}})Lin, Yi, Ma, Jiang, Luo, Shi, and Liu}]{lin2023zero}
Hongzhan Lin, Pengyao Yi, Jing Ma, Haiyun Jiang, Ziyang Luo, Shuming Shi, and Ruifang Liu. 2023{\natexlab{b}}.
\newblock Zero-shot rumor detection with propagation structure via prompt learning.
\newblock In \emph{Proceedings of the AAAI Conference on Artificial Intelligence}, volume~37, pages 5213--5221.

\bibitem[{Lin et~al.(2022{\natexlab{b}})Lin, Hilton, and Evans}]{lin2022teaching}
Stephanie Lin, Jacob Hilton, and Owain Evans. 2022{\natexlab{b}}.
\newblock Teaching models to express their uncertainty in words.
\newblock \emph{arXiv preprint arXiv:2205.14334}.

\bibitem[{Liu et~al.(2021)Liu, Wang, Wang, and Ordonez}]{liu-etal-2021-visual}
Fuxiao Liu, Yinghan Wang, Tianlu Wang, and Vicente Ordonez. 2021.
\newblock \href {https://doi.org/10.18653/v1/2021.emnlp-main.542} {Visual news: Benchmark and challenges in news image captioning}.
\newblock In \emph{Proceedings of the 2021 Conference on Empirical Methods in Natural Language Processing}, pages 6761--6771, Online and Punta Cana, Dominican Republic. Association for Computational Linguistics.

\bibitem[{Liu et~al.(2024{\natexlab{a}})Liu, Li, Li, Li, Zhang, Shen, and Lee}]{liu2024llavanext}
Haotian Liu, Chunyuan Li, Yuheng Li, Bo~Li, Yuanhan Zhang, Sheng Shen, and Yong~Jae Lee. 2024{\natexlab{a}}.
\newblock \href {https://llava-vl.github.io/blog/2024-01-30-llava-next/} {Llava-next: Improved reasoning, ocr, and world knowledge}.

\bibitem[{Liu et~al.(2023{\natexlab{a}})Liu, Li, Wu, and Lee}]{liu2023visual}
Haotian Liu, Chunyuan Li, Qingyang Wu, and Yong~Jae Lee. 2023{\natexlab{a}}.
\newblock Visual instruction tuning.
\newblock In \emph{Thirty-seventh Conference on Neural Information Processing Systems}.

\bibitem[{Liu et~al.(2023{\natexlab{b}})Liu, Zeng, Ren, Li, Zhang, Yang, yue Li, Yang, Su, Zhu, and Zhang}]{Liu2023GroundingDM}
Shilong Liu, Zhaoyang Zeng, Tianhe Ren, Feng Li, Hao Zhang, Jie Yang, Chun yue Li, Jianwei Yang, Hang Su, Jun-Juan Zhu, and Lei Zhang. 2023{\natexlab{b}}.
\newblock \href {https://api.semanticscholar.org/CorpusID:257427307} {Grounding dino: Marrying dino with grounded pre-training for open-set object detection}.
\newblock \emph{ArXiv}, abs/2303.05499.

\bibitem[{Liu et~al.(2024{\natexlab{b}})Liu, Li, Li, Xia, Cui, Huang, Huang, Deng, and He}]{liu2024mmfakebenchmixedsourcemultimodalmisinformation}
Xuannan Liu, Zekun Li, Peipei Li, Shuhan Xia, Xing Cui, Linzhi Huang, Huaibo Huang, Weihong Deng, and Zhaofeng He. 2024{\natexlab{b}}.
\newblock \href {https://arxiv.org/abs/2406.08772} {Mmfakebench: A mixed-source multimodal misinformation detection benchmark for lvlms}.
\newblock \emph{Preprint}, arXiv:2406.08772.

\bibitem[{Liu et~al.(2023{\natexlab{c}})Liu, Duan, Zhang, Li, Zhang, Zhao, Yuan, Wang, He, Liu et~al.}]{liu2023mmbench}
Yuan Liu, Haodong Duan, Yuanhan Zhang, Bo~Li, Songyang Zhang, Wangbo Zhao, Yike Yuan, Jiaqi Wang, Conghui He, Ziwei Liu, et~al. 2023{\natexlab{c}}.
\newblock Mmbench: Is your multi-modal model an all-around player?
\newblock \emph{arXiv preprint arXiv:2307.06281}.

\bibitem[{Luo et~al.(2021)Luo, Darrell, and Rohrbach}]{luo-etal-2021-newsclippings}
Grace Luo, Trevor Darrell, and Anna Rohrbach. 2021.
\newblock \href {https://doi.org/10.18653/v1/2021.emnlp-main.545} {{N}ews{CLIP}pings: {A}utomatic {G}eneration of {O}ut-of-{C}ontext {M}ultimodal {M}edia}.
\newblock In \emph{Proceedings of the 2021 Conference on Empirical Methods in Natural Language Processing}, pages 6801--6817, Online and Punta Cana, Dominican Republic. Association for Computational Linguistics.

\bibitem[{Luo et~al.(2023{\natexlab{a}})Luo, Sun, Xu, Zhao, Lou, Tao, Geng, Lin, Chen, and Zhang}]{luo2023wizardmath}
Haipeng Luo, Qingfeng Sun, Can Xu, Pu~Zhao, Jianguang Lou, Chongyang Tao, Xiubo Geng, Qingwei Lin, Shifeng Chen, and Dongmei Zhang. 2023{\natexlab{a}}.
\newblock \href {https://arxiv.org/abs/2308.09583} {Wizardmath: Empowering mathematical reasoning for large language models via reinforced evol-instruct}.
\newblock \emph{Preprint}, arXiv:2308.09583.

\bibitem[{Luo et~al.(2023{\natexlab{b}})Luo, Xu, Zhao, Sun, Geng, Hu, Tao, Ma, Lin, and Jiang}]{luo2023wizardcoder}
Ziyang Luo, Can Xu, Pu~Zhao, Qingfeng Sun, Xiubo Geng, Wenxiang Hu, Chongyang Tao, Jing Ma, Qingwei Lin, and Daxin Jiang. 2023{\natexlab{b}}.
\newblock Wizardcoder: Empowering code large language models with evol-instruct.
\newblock \emph{arXiv preprint arXiv:2306.08568}.

\bibitem[{Maras and Alexandrou(2018)}]{Maras2018DeterminingAO}
Marie-Helen Maras and Alex Alexandrou. 2018.
\newblock \href {https://api.semanticscholar.org/CorpusID:150336977} {Determining authenticity of video evidence in the age of artificial intelligence and in the wake of deepfake videos}.
\newblock \emph{The International Journal of Evidence \& Proof}, 23:255 -- 262.

\bibitem[{Mesnard et~al.(2024)Mesnard, Hardin, Dadashi, Bhupatiraju, Pathak, Sifre, Rivi{\`{e}}re, Kale, Love, Tafti, Hussenot, Chowdhery, Roberts, Barua, Botev, Castro{-}Ros, Slone, H{\'{e}}liou, Tacchetti, Bulanova, Paterson, Tsai, Shahriari, Lan, Choquette{-}Choo, Crepy, Cer, Ippolito, Reid, Buchatskaya, Ni, Noland, Yan, Tucker, Muraru, Rozhdestvenskiy, Michalewski, Tenney, Grishchenko, Austin, Keeling, Labanowski, Lespiau, Stanway, Brennan, Chen, Ferret, Chiu, and et~al.}]{gemma}
Thomas Mesnard, Cassidy Hardin, Robert Dadashi, Surya Bhupatiraju, Shreya Pathak, Laurent Sifre, Morgane Rivi{\`{e}}re, Mihir~Sanjay Kale, Juliette Love, Pouya Tafti, L{\'{e}}onard Hussenot, Aakanksha Chowdhery, Adam Roberts, Aditya Barua, Alex Botev, Alex Castro{-}Ros, Ambrose Slone, Am{\'{e}}lie H{\'{e}}liou, Andrea Tacchetti, Anna Bulanova, Antonia Paterson, Beth Tsai, Bobak Shahriari, Charline~Le Lan, Christopher~A. Choquette{-}Choo, Cl{\'{e}}ment Crepy, Daniel Cer, Daphne Ippolito, David Reid, Elena Buchatskaya, Eric Ni, Eric Noland, Geng Yan, George Tucker, George{-}Cristian Muraru, Grigory Rozhdestvenskiy, Henryk Michalewski, Ian Tenney, Ivan Grishchenko, Jacob Austin, James Keeling, Jane Labanowski, Jean{-}Baptiste Lespiau, Jeff Stanway, Jenny Brennan, Jeremy Chen, Johan Ferret, Justin Chiu, and et~al. 2024.
\newblock \href {https://doi.org/10.48550/ARXIV.2403.08295} {Gemma: Open models based on gemini research and technology}.
\newblock \emph{CoRR}, abs/2403.08295.

\bibitem[{Meta(2024)}]{llama3}
Meta. 2024.
\newblock Introducing meta llama 3: The most capable openly available llm to date.
\newblock \url{https://ai.meta.com/blog/meta-llama-3/}.

\bibitem[{Mukherjee et~al.(2023)Mukherjee, Mitra, Jawahar, Agarwal, Palangi, and Awadallah}]{mukherjee2023orca}
Subhabrata Mukherjee, Arindam Mitra, Ganesh Jawahar, Sahaj Agarwal, Hamid Palangi, and Ahmed Awadallah. 2023.
\newblock \href {https://arxiv.org/abs/2306.02707} {Orca: Progressive learning from complex explanation traces of gpt-4}.
\newblock \emph{Preprint}, arXiv:2306.02707.

\bibitem[{Nakamura et~al.(2020)Nakamura, Levy, and Wang}]{nakamura-etal-2020-fakeddit}
Kai Nakamura, Sharon Levy, and William~Yang Wang. 2020.
\newblock \href {https://aclanthology.org/2020.lrec-1.755} {{F}akeddit: A new multimodal benchmark dataset for fine-grained fake news detection}.
\newblock In \emph{Proceedings of the Twelfth Language Resources and Evaluation Conference}, pages 6149--6157, Marseille, France. European Language Resources Association.

\bibitem[{Newman et~al.(2012)Newman, Garry, Bernstein, Kantner, and Lindsay}]{newman2012nonprobative}
Eryn~J Newman, Maryanne Garry, Daniel~M Bernstein, Justin Kantner, and D~Stephen Lindsay. 2012.
\newblock Nonprobative photographs (or words) inflate truthiness.
\newblock \emph{Psychonomic Bulletin \& Review}, 19:969--974.

\bibitem[{OpenAI(2023)}]{OpenAI2023GPT4TR}
OpenAI. 2023.
\newblock \href {https://api.semanticscholar.org/CorpusID:257532815} {Gpt-4 technical report}.
\newblock \emph{ArXiv}, abs/2303.08774.

\bibitem[{Ouyang et~al.(2022)Ouyang, Wu, Jiang, Almeida, Wainwright, Mishkin, Zhang, Agarwal, Slama, Gray et~al.}]{ouyang2022training}
Long Ouyang, Jeffrey Wu, Xu~Jiang, Diogo Almeida, Carroll Wainwright, Pamela Mishkin, Chong Zhang, Sandhini Agarwal, Katarina Slama, Alex Gray, et~al. 2022.
\newblock Training language models to follow instructions with human feedback.
\newblock In \emph{Advances in Neural Information Processing Systems}.

\bibitem[{Pan et~al.(2023)Pan, Wu, Lu, Luu, Wang, Kan, and Nakov}]{pan2023fact}
Liangming Pan, Xiaobao Wu, Xinyuan Lu, Anh~Tuan Luu, William~Yang Wang, Min-Yen Kan, and Preslav Nakov. 2023.
\newblock Fact-checking complex claims with program-guided reasoning.
\newblock In \emph{Proceedings of the 61st Annual Meeting of the Association for Computational Linguistics (Volume 1: Long Papers)}, pages 6981--7004.

\bibitem[{Petroni et~al.(2020)Petroni, Lewis, Piktus, Rockt{\"a}schel, Wu, Miller, and Riedel}]{petroni2020context}
Fabio Petroni, Patrick Lewis, Aleksandra Piktus, Tim Rockt{\"a}schel, Yuxiang Wu, Alexander~H Miller, and Sebastian Riedel. 2020.
\newblock How context affects language models' factual predictions.
\newblock In \emph{Automated Knowledge Base Construction}.

\bibitem[{Petroni et~al.(2019)Petroni, Rockt{\"a}schel, Riedel, Lewis, Bakhtin, Wu, and Miller}]{petroni2019language}
Fabio Petroni, Tim Rockt{\"a}schel, Sebastian Riedel, Patrick Lewis, Anton Bakhtin, Yuxiang Wu, and Alexander Miller. 2019.
\newblock Language models as knowledge bases?
\newblock In \emph{Proceedings of the 2019 Conference on Empirical Methods in Natural Language Processing and the 9th International Joint Conference on Natural Language Processing (EMNLP-IJCNLP)}, pages 2463--2473.

\bibitem[{Powers(2020)}]{powers2020evaluation}
David~MW Powers. 2020.
\newblock Evaluation: from precision, recall and f-measure to roc, informedness, markedness and correlation.
\newblock \emph{arXiv preprint arXiv:2010.16061}.

\bibitem[{Qi et~al.(2019)Qi, Cao, Yang, Guo, and Li}]{qi2019exploiting}
Peng Qi, Juan Cao, Tianyun Yang, Junbo Guo, and Jintao Li. 2019.
\newblock Exploiting multi-domain visual information for fake news detection.
\newblock In \emph{2019 IEEE international conference on data mining (ICDM)}, pages 518--527. IEEE.

\bibitem[{Radford et~al.(2021)Radford, Kim, Hallacy, Ramesh, Goh, Agarwal, Sastry, Askell, Mishkin, Clark, Krueger, and Sutskever}]{radford2021learning}
Alec Radford, Jong~Wook Kim, Chris Hallacy, Aditya Ramesh, Gabriel Goh, Sandhini Agarwal, Girish Sastry, Amanda Askell, Pamela Mishkin, Jack Clark, Gretchen Krueger, and Ilya Sutskever. 2021.
\newblock \href {https://arxiv.org/abs/2103.00020} {Learning transferable visual models from natural language supervision}.
\newblock \emph{Preprint}, arXiv:2103.00020.

\bibitem[{Ramesh et~al.(2022)Ramesh, Dhariwal, Nichol, Chu, and Chen}]{Ramesh2022HierarchicalTI}
Aditya Ramesh, Prafulla Dhariwal, Alex Nichol, Casey Chu, and Mark Chen. 2022.
\newblock \href {https://api.semanticscholar.org/CorpusID:248097655} {Hierarchical text-conditional image generation with clip latents}.
\newblock \emph{ArXiv}, abs/2204.06125.

\bibitem[{Roberts et~al.(2020)Roberts, Raffel, and Shazeer}]{roberts2020much}
Adam Roberts, Colin Raffel, and Noam Shazeer. 2020.
\newblock How much knowledge can you pack into the parameters of a language model?
\newblock In \emph{Proceedings of the 2020 Conference on Empirical Methods in Natural Language Processing (EMNLP)}, pages 5418--5426.

\bibitem[{Rombach et~al.(2022)Rombach, Blattmann, Lorenz, Esser, and Ommer}]{Rombach_2022_CVPR}
Robin Rombach, Andreas Blattmann, Dominik Lorenz, Patrick Esser, and Bj\"orn Ommer. 2022.
\newblock High-resolution image synthesis with latent diffusion models.
\newblock In \emph{Proceedings of the IEEE/CVF Conference on Computer Vision and Pattern Recognition (CVPR)}, pages 10684--10695.

\bibitem[{Sabir et~al.(2018)Sabir, AbdAlmageed, Wu, and Natarajan}]{10.1145/3240508.3240707}
Ekraam Sabir, Wael AbdAlmageed, Yue Wu, and Prem Natarajan. 2018.
\newblock \href {https://doi.org/10.1145/3240508.3240707} {Deep multimodal image-repurposing detection}.
\newblock In \emph{Proceedings of the 26th ACM International Conference on Multimedia}, MM '18, page 1337–1345, New York, NY, USA. Association for Computing Machinery.

\bibitem[{Shao et~al.(2023)Shao, Wu, and Liu}]{Shao_2023_CVPR}
Rui Shao, Tianxing Wu, and Ziwei Liu. 2023.
\newblock Detecting and grounding multi-modal media manipulation.
\newblock In \emph{Proceedings of the IEEE/CVF Conference on Computer Vision and Pattern Recognition (CVPR)}, pages 6904--6913.

\bibitem[{Sun et~al.(2023)Sun, Cui, Zhang, Zhang, Yu, Luo, Wang, Rao, Liu, Huang, and Wang}]{Sun2023GenerativeMM}
Quan Sun, Yufeng Cui, Xiaosong Zhang, Fan Zhang, Qiying Yu, Zhengxiong Luo, Yueze Wang, Yongming Rao, Jingjing Liu, Tiejun Huang, and Xinlong Wang. 2023.
\newblock \href {https://api.semanticscholar.org/CorpusID:266374640} {Generative multimodal models are in-context learners}.
\newblock \emph{ArXiv}, abs/2312.13286.

\bibitem[{Tay et~al.(2023)Tay, Dehghani, Tran, Garcia, Wei, Wang, Chung, Shakeri, Bahri, Schuster, Zheng, Zhou, Houlsby, and Metzler}]{tay2023ul2}
Yi~Tay, Mostafa Dehghani, Vinh~Q. Tran, Xavier Garcia, Jason Wei, Xuezhi Wang, Hyung~Won Chung, Siamak Shakeri, Dara Bahri, Tal Schuster, Huaixiu~Steven Zheng, Denny Zhou, Neil Houlsby, and Donald Metzler. 2023.
\newblock \href {https://arxiv.org/abs/2205.05131} {Ul2: Unifying language learning paradigms}.
\newblock \emph{Preprint}, arXiv:2205.05131.

\bibitem[{Team et~al.(2023)Team, Anil, Borgeaud, Wu, Alayrac, Yu, Soricut, Schalkwyk, Dai, Hauth et~al.}]{team2023gemini}
Gemini Team, Rohan Anil, Sebastian Borgeaud, Yonghui Wu, Jean-Baptiste Alayrac, Jiahui Yu, Radu Soricut, Johan Schalkwyk, Andrew~M Dai, Anja Hauth, et~al. 2023.
\newblock Gemini: A family of highly capable multimodal models.
\newblock \emph{arXiv preprint arXiv:2312.11805}.

\bibitem[{Thorne et~al.(2018)Thorne, Vlachos, Christodoulopoulos, and Mittal}]{thorne2018fever}
James Thorne, Andreas Vlachos, Christos Christodoulopoulos, and Arpit Mittal. 2018.
\newblock Fever: a large-scale dataset for fact extraction and verification.
\newblock In \emph{Proceedings of the 2018 Conference of the North American Chapter of the Association for Computational Linguistics: Human Language Technologies, Volume 1 (Long Papers)}, pages 809--819.

\bibitem[{Touvron et~al.(2023{\natexlab{a}})Touvron, Lavril, Izacard, Martinet, Lachaux, Lacroix, Rozi{\`e}re, Goyal, Hambro, Azhar et~al.}]{touvron2023llama}
Hugo Touvron, Thibaut Lavril, Gautier Izacard, Xavier Martinet, Marie-Anne Lachaux, Timoth{\'e}e Lacroix, Baptiste Rozi{\`e}re, Naman Goyal, Eric Hambro, Faisal Azhar, et~al. 2023{\natexlab{a}}.
\newblock Llama: Open and efficient foundation language models.
\newblock \emph{arXiv preprint arXiv:2302.13971}.

\bibitem[{Touvron et~al.(2023{\natexlab{b}})Touvron, Martin, Stone, Albert, Almahairi, Babaei, Bashlykov, Batra, Bhargava, Bhosale et~al.}]{touvron2023llama2}
Hugo Touvron, Louis Martin, Kevin Stone, Peter Albert, Amjad Almahairi, Yasmine Babaei, Nikolay Bashlykov, Soumya Batra, Prajjwal Bhargava, Shruti Bhosale, et~al. 2023{\natexlab{b}}.
\newblock Llama 2: Open foundation and fine-tuned chat models.
\newblock \emph{arXiv preprint arXiv:2307.09288}.

\bibitem[{Varshney et~al.(2022)Varshney, Mishra, and Baral}]{varshney2022investigating}
Neeraj Varshney, Swaroop Mishra, and Chitta Baral. 2022.
\newblock Investigating selective prediction approaches across several tasks in iid, ood, and adversarial settings.
\newblock In \emph{Findings of the Association for Computational Linguistics: ACL 2022}, pages 1995--2002.

\bibitem[{Wang and Kuo(2020)}]{Wang2020SBERTWKAS}
Bin Wang and C.-C.~Jay Kuo. 2020.
\newblock \href {https://api.semanticscholar.org/CorpusID:211133229} {Sbert-wk: A sentence embedding method by dissecting bert-based word models}.
\newblock \emph{IEEE/ACM Transactions on Audio, Speech, and Language Processing}, 28:2146--2157.

\bibitem[{Wang et~al.(2024{\natexlab{a}})Wang, Ma, Lin, Yang, Yang, Tian, and Chang}]{wang2024explainable}
Bo~Wang, Jing Ma, Hongzhan Lin, Zhiwei Yang, Ruichao Yang, Yuan Tian, and Yi~Chang. 2024{\natexlab{a}}.
\newblock Explainable fake news detection with large language model via defense among competing wisdom.
\newblock In \emph{Proceedings of the ACM on Web Conference 2024}, pages 2452--2463.

\bibitem[{Wang et~al.(2024{\natexlab{b}})Wang, Bai, Tan, Wang, Fan, Bai, Chen, Liu, Wang, Ge, Fan, Dang, Du, Ren, Men, Liu, Zhou, Zhou, and Lin}]{Qwen2VL}
Peng Wang, Shuai Bai, Sinan Tan, Shijie Wang, Zhihao Fan, Jinze Bai, Keqin Chen, Xuejing Liu, Jialin Wang, Wenbin Ge, Yang Fan, Kai Dang, Mengfei Du, Xuancheng Ren, Rui Men, Dayiheng Liu, Chang Zhou, Jingren Zhou, and Junyang Lin. 2024{\natexlab{b}}.
\newblock Qwen2-vl: Enhancing vision-language model's perception of the world at any resolution.
\newblock \emph{arXiv preprint arXiv:2409.12191}.

\bibitem[{Wang et~al.(2023{\natexlab{a}})Wang, Lv, Yu, Hong, Qi, Wang, Ji, Yang, Zhao, Song, Xu, Xu, Li, Dong, Ding, and Tang}]{Wang2023CogVLMVE}
Weihan Wang, Qingsong Lv, Wenmeng Yu, Wenyi Hong, Ji~Qi, Yan Wang, Junhui Ji, Zhuoyi Yang, Lei Zhao, Xixuan Song, Jiazheng Xu, Bin Xu, Juanzi Li, Yuxiao Dong, Ming Ding, and Jie Tang. 2023{\natexlab{a}}.
\newblock \href {https://api.semanticscholar.org/CorpusID:265034288} {Cogvlm: Visual expert for pretrained language models}.
\newblock \emph{ArXiv}, abs/2311.03079.

\bibitem[{Wang et~al.(2023{\natexlab{b}})Wang, Lv, Yu, Hong, Qi, Wang, Ji, Yang, Zhao, Song et~al.}]{wang2023cogvlm}
Weihan Wang, Qingsong Lv, Wenmeng Yu, Wenyi Hong, Ji~Qi, Yan Wang, Junhui Ji, Zhuoyi Yang, Lei Zhao, Xixuan Song, et~al. 2023{\natexlab{b}}.
\newblock Cogvlm: Visual expert for pretrained language models.
\newblock \emph{arXiv preprint arXiv:2311.03079}.

\bibitem[{Wang et~al.(2023{\natexlab{c}})Wang, Kordi, Mishra, Liu, Smith, Khashabi, and Hajishirzi}]{wang2023selfinstruct}
Yizhong Wang, Yeganeh Kordi, Swaroop Mishra, Alisa Liu, Noah~A. Smith, Daniel Khashabi, and Hannaneh Hajishirzi. 2023{\natexlab{c}}.
\newblock \href {https://arxiv.org/abs/2212.10560} {Self-instruct: Aligning language models with self-generated instructions}.
\newblock \emph{Preprint}, arXiv:2212.10560.

\bibitem[{Wei et~al.(2022)Wei, Wang, Schuurmans, Bosma, Xia, Chi, Le, Zhou et~al.}]{wei2022chain}
Jason Wei, Xuezhi Wang, Dale Schuurmans, Maarten Bosma, Fei Xia, Ed~H Chi, Quoc~V Le, Denny Zhou, et~al. 2022.
\newblock Chain-of-thought prompting elicits reasoning in large language models.
\newblock In \emph{Advances in Neural Information Processing Systems}.

\bibitem[{Xu et~al.(2023)Xu, Sun, Zheng, Geng, Zhao, Feng, Tao, and Jiang}]{xu2023wizardlm}
Can Xu, Qingfeng Sun, Kai Zheng, Xiubo Geng, Pu~Zhao, Jiazhan Feng, Chongyang Tao, and Daxin Jiang. 2023.
\newblock \href {https://arxiv.org/abs/2304.12244} {Wizardlm: Empowering large language models to follow complex instructions}.
\newblock \emph{Preprint}, arXiv:2304.12244.

\bibitem[{Xue et~al.(2024)Xue, Shu, Awadalla, Wang, Yan, Purushwalkam, Zhou, Prabhu, Dai, Ryoo, Kendre, Zhang, Qin, Zhang, Chen, Yu, Tan, Awalgaonkar, Heinecke, Wang, Choi, Schmidt, Chen, Savarese, Niebles, Xiong, and Xu}]{xue2024xgenmmblip3familyopen}
Le~Xue, Manli Shu, Anas Awadalla, Jun Wang, An~Yan, Senthil Purushwalkam, Honglu Zhou, Viraj Prabhu, Yutong Dai, Michael~S Ryoo, Shrikant Kendre, Jieyu Zhang, Can Qin, Shu Zhang, Chia-Chih Chen, Ning Yu, Juntao Tan, Tulika~Manoj Awalgaonkar, Shelby Heinecke, Huan Wang, Yejin Choi, Ludwig Schmidt, Zeyuan Chen, Silvio Savarese, Juan~Carlos Niebles, Caiming Xiong, and Ran Xu. 2024.
\newblock \href {https://arxiv.org/abs/2408.08872} {xgen-mm (blip-3): A family of open large multimodal models}.
\newblock \emph{Preprint}, arXiv:2408.08872.

\bibitem[{Yang et~al.(2023)Yang, Li, Lin, Wang, Lin, Liu, and Wang}]{yang2023dawn}
Zhengyuan Yang, Linjie Li, Kevin Lin, Jianfeng Wang, Chung-Ching Lin, Zicheng Liu, and Lijuan Wang. 2023.
\newblock The dawn of lmms: Preliminary explorations with gpt-4v (ision).
\newblock \emph{arXiv preprint arXiv:2309.17421}, 9(1).

\bibitem[{Yao et~al.(2023)Yao, Shah, Sun, Cho, and Huang}]{10.1145/3539618.3591879}
Barry~Menglong Yao, Aditya Shah, Lichao Sun, Jin-Hee Cho, and Lifu Huang. 2023.
\newblock \href {https://doi.org/10.1145/3539618.3591879} {End-to-end multimodal fact-checking and explanation generation: A challenging dataset and models}.
\newblock In \emph{Proceedings of the 46th International ACM SIGIR Conference on Research and Development in Information Retrieval}, SIGIR '23, page 2733–2743, New York, NY, USA. Association for Computing Machinery.

\bibitem[{Yao et~al.(2024)Yao, Yu, Zhang, Wang, Cui, Zhu, Cai, Li, Zhao, He et~al.}]{yao2024minicpm}
Yuan Yao, Tianyu Yu, Ao~Zhang, Chongyi Wang, Junbo Cui, Hongji Zhu, Tianchi Cai, Haoyu Li, Weilin Zhao, Zhihui He, et~al. 2024.
\newblock Minicpm-v: A gpt-4v level mllm on your phone.
\newblock \emph{arXiv preprint arXiv:2408.01800}.

\bibitem[{Ye et~al.(2023{\natexlab{a}})Ye, Xu, Xu, Ye, Yan, Zhou, Wang, Hu, Shi, Shi, Li, Xu, Chen, Tian, Qi, Zhang, and Huang}]{Ye2023mPLUGOwlME}
Qinghao Ye, Haiyang Xu, Guohai Xu, Jiabo Ye, Ming Yan, Yi~Zhou, Junyan Wang, Anwen Hu, Pengcheng Shi, Yaya Shi, Chenliang Li, Yuanhong Xu, Hehong Chen, Junfeng Tian, Qiang Qi, Ji~Zhang, and Feiyan Huang. 2023{\natexlab{a}}.
\newblock \href {https://api.semanticscholar.org/CorpusID:258352455} {mplug-owl: Modularization empowers large language models with multimodality}.
\newblock \emph{ArXiv}, abs/2304.14178.

\bibitem[{Ye et~al.(2023{\natexlab{b}})Ye, Xu, Xu, Ye, Yan, Zhou, Wang, Hu, Shi, Shi et~al.}]{ye2023mplug}
Qinghao Ye, Haiyang Xu, Guohai Xu, Jiabo Ye, Ming Yan, Yiyang Zhou, Junyang Wang, Anwen Hu, Pengcheng Shi, Yaya Shi, et~al. 2023{\natexlab{b}}.
\newblock mplug-owl: Modularization empowers large language models with multimodality.
\newblock \emph{arXiv preprint arXiv:2304.14178}.

\bibitem[{Yin et~al.(2023)Yin, Wang, Cao, Shi, Liu, Li, Sheng, Bai, Huang, Wang et~al.}]{yin2023lamm}
Zhenfei Yin, Jiong Wang, Jianjian Cao, Zhelun Shi, Dingning Liu, Mukai Li, Lu~Sheng, Lei Bai, Xiaoshui Huang, Zhiyong Wang, et~al. 2023.
\newblock Lamm: Language-assisted multi-modal instruction-tuning dataset, framework, and benchmark.
\newblock \emph{arXiv preprint arXiv:2306.06687}.

\bibitem[{Young et~al.(2024)Young, Chen, Li, Huang, Zhang, Zhang, Li, Zhu, Chen, Chang, Yu, Liu, Liu, Yue, Yang, Yang, Yu, Xie, Huang, Hu, Ren, Niu, Nie, Xu, Liu, Wang, Cai, Gu, Liu, and Dai}]{Young2024YiOF}
01.AI~Alex Young, Bei Chen, Chao Li, Chengen Huang, Ge~Zhang, Guanwei Zhang, Heng Li, Jiangcheng Zhu, Jianqun Chen, Jing Chang, Kaidong Yu, Peng Liu, Qiang Liu, Shawn Yue, Senbin Yang, Shiming Yang, Tao Yu, Wen Xie, Wenhao Huang, Xiaohui Hu, Xiaoyi Ren, Xinyao Niu, Pengcheng Nie, Yuchi Xu, Yudong Liu, Yue Wang, Yuxuan Cai, Zhenyu Gu, Zhiyuan Liu, and Zonghong Dai. 2024.
\newblock \href {https://api.semanticscholar.org/CorpusID:268264158} {Yi: Open foundation models by 01.ai}.
\newblock \emph{ArXiv}, abs/2403.04652.

\bibitem[{Yu et~al.(2023)Yu, Yang, Li, Wang, Lin, Liu, Wang, and Wang}]{yu2023mm}
Weihao Yu, Zhengyuan Yang, Linjie Li, Jianfeng Wang, Kevin Lin, Zicheng Liu, Xinchao Wang, and Lijuan Wang. 2023.
\newblock Mm-vet: Evaluating large multimodal models for integrated capabilities.
\newblock \emph{arXiv preprint arXiv:2308.02490}.

\bibitem[{Yu et~al.(2022)Yu, Iter, Wang, Xu, Ju, Sanyal, Zhu, Zeng, and Jiang}]{yu2022generate}
Wenhao Yu, Dan Iter, Shuohang Wang, Yichong Xu, Mingxuan Ju, Soumya Sanyal, Chenguang Zhu, Michael Zeng, and Meng Jiang. 2022.
\newblock Generate rather than retrieve: Large language models are strong context generators.
\newblock In \emph{The Eleventh International Conference on Learning Representations}.

\bibitem[{Zellers et~al.(2019)Zellers, Bisk, Farhadi, and Choi}]{zellers2019recognition}
Rowan Zellers, Yonatan Bisk, Ali Farhadi, and Yejin Choi. 2019.
\newblock From recognition to cognition: Visual commonsense reasoning.
\newblock In \emph{Proceedings of the IEEE/CVF conference on computer vision and pattern recognition}, pages 6720--6731.

\bibitem[{Zeng et~al.(2023)Zeng, Liu, Du, Wang, Lai, Ding, Yang, Xu, Zheng, Xia, Tam, Ma, Xue, Zhai, Chen, Zhang, Dong, and Tang}]{zeng2023glm130b}
Aohan Zeng, Xiao Liu, Zhengxiao Du, Zihan Wang, Hanyu Lai, Ming Ding, Zhuoyi Yang, Yifan Xu, Wendi Zheng, Xiao Xia, Weng~Lam Tam, Zixuan Ma, Yufei Xue, Jidong Zhai, Wenguang Chen, Peng Zhang, Yuxiao Dong, and Jie Tang. 2023.
\newblock \href {https://arxiv.org/abs/2210.02414} {Glm-130b: An open bilingual pre-trained model}.
\newblock \emph{Preprint}, arXiv:2210.02414.

\bibitem[{Zhang et~al.(2024)Zhang, Yu, Li, Dong, Su, Chu, and Yu}]{zhang2024mm}
Duzhen Zhang, Yahan Yu, Chenxing Li, Jiahua Dong, Dan Su, Chenhui Chu, and Dong Yu. 2024.
\newblock Mm-llms: Recent advances in multimodal large language models.
\newblock \emph{arXiv preprint arXiv:2401.13601}.

\bibitem[{Zhang et~al.(2022)Zhang, Roller, Goyal, Artetxe, Chen, Chen, Dewan, Diab, Li, Lin, Mihaylov, Ott, Shleifer, Shuster, Simig, Koura, Sridhar, Wang, and Zettlemoyer}]{opt}
Susan Zhang, Stephen Roller, Naman Goyal, Mikel Artetxe, Moya Chen, Shuohui Chen, Christopher Dewan, Mona Diab, Xian Li, Xi~Victoria Lin, Todor Mihaylov, Myle Ott, Sam Shleifer, Kurt Shuster, Daniel Simig, Punit~Singh Koura, Anjali Sridhar, Tianlu Wang, and Luke Zettlemoyer. 2022.
\newblock \href {https://arxiv.org/abs/2205.01068} {Opt: Open pre-trained transformer language models}.
\newblock \emph{Preprint}, arXiv:2205.01068.

\bibitem[{Zhou et~al.(2023)Zhou, Liu, Xu, Iyer, Sun, Mao, Ma, Efrat, Yu, Yu, Zhang, Ghosh, Lewis, Zettlemoyer, and Levy}]{zhou2023lima}
Chunting Zhou, Pengfei Liu, Puxin Xu, Srini Iyer, Jiao Sun, Yuning Mao, Xuezhe Ma, Avia Efrat, Ping Yu, Lili Yu, Susan Zhang, Gargi Ghosh, Mike Lewis, Luke Zettlemoyer, and Omer Levy. 2023.
\newblock \href {https://arxiv.org/abs/2305.11206} {Lima: Less is more for alignment}.
\newblock \emph{Preprint}, arXiv:2305.11206.

\bibitem[{Zhu et~al.(2023)Zhu, Chen, Shen, Li, and Elhoseiny}]{zhu2023minigpt}
Deyao Zhu, Jun Chen, Xiaoqian Shen, Xiang Li, and Mohamed Elhoseiny. 2023.
\newblock Minigpt-4: Enhancing vision-language understanding with advanced large language models.
\newblock \emph{arXiv preprint arXiv:2304.10592}.

\end{thebibliography}

\appendix

\appendix

\section{Distribution}

The dataset is publicly available on the Hugging Face anonymous page: \href{https://huggingface.co/datasets/Anonymous-2024/manipulation-mfc-bench}{Manipulation Classification},  \href{https://huggingface.co/datasets/Anonymous-2024/ooc-mfc-bench}{OOC  Classification} and \href{https://huggingface.co/datasets/Anonymous-2024/veracity-mfc-bench}{Veracity Classification}.

The dataset is accompanied by Croissant metadata and licensing information all available on Hugging Face Hub.

\section{Descriptions of LVLM Baselines}
\label{sec:baselines}
We conduct extensive experiments on the \benchname{} to evaluate the following representative LVLMs:
\begin{itemize}
    \item \textsf{GPT-4o}, the latest flagship model developed by OpenAI, designed for real-time reasoning across audio, visual, and textual inputs. It excels in understanding both vision and audio, offering significant improvements over previous models in these areas.  We specifically utilize the ``gpt-4o-2024-05-13'' version. 
    \item \textsf{GPT-4V}~\cite{OpenAI2023GPT4TR}, developed by OpenAI, is a version of the GPT-4 architecture that includes capabilities for processing and generating images in addition to text.  We specifically utilize the ``gpt-4-vision-preview'' version.
    \item \textsf{Claude3.5-Sonnet} developed by Anthropic with significant improvements most evident in visual reasoning tasks like interpreting charts and graphs, and it can accurately transcribe text from imperfect images We specifically utilize the ``claude-3-5-sonnet-20240620'' version.
    \item \textsf{Claude3-Haiku}\footnote{https://claude.ai/}, developed by Anthropic, possesses sophisticated vision capabilities comparable to other leading models. It can process a wide range of visual formats, including photos, charts, graphs, and technical diagrams. We specifically utilize the ``claude-3-haiku-20240307'' version.
    \item \textsf{Gemini-1.5-Pro} developed by google, can perform highly-sophisticated understanding and reasoning tasks for different modalities, including vision. We specifically utilize the ``gemini-1.5-pro'' version
    \item \textsf{Emu2}~\cite{Sun2023GenerativeMM} is a generative multimodal model with 37 billion parameters, designed to enhance task-agnostic in-context learning capabilities through effective scaling. We specifically utilize the ``Emu2'' version.
    \item \textsf{InternVL}~\cite{Chen2023InternVLSU} is a large-scale vision-language foundation model, scaling up the vision foundation model to 6 billion parameters and progressively aligning it with the LLM, using web-scale image-text data from various sources. We specifically utilize the ``InternVL-Chat-V1-5'' version.
    \item \textsf{CogVLM}~\cite{Wang2023CogVLMVE} is a powerful open-source visual language foundation model that achieves state-of-the-art performance on multiple cross-modal benchmarks by using a trainable visual expert module for deep fusion of vision and language features. We specifically utilize the ``cogvlm-chat'' version.
    \item \textsf{LLaVA-NeXT}~\cite{liu2024llavanext} is the new version of LLaVA~\cite{liu2023visual}, with improved reasoning, OCR, and world knowledge capabilities.  We specifically utilize the ``llava-v1.6-vicuna-7b, llava-v1.6-vicuna-13b, llava-v1.6-34b'' version.
    \item \textsf{InstructBLIP}~\cite{Dai2023InstructBLIPTG} introduces a novel vision-language instruction-tuning framework utilizing BLIP-2 models to enhance zero-shot generalization performance across diverse vision-language tasks. We specifically utilize the ``instructblip-vicuna-7b, instructblip-vicuna-13b'' version.
    \item \textsf{Pixtral}\footnote{https://mistral.ai/news/pixtral-12b/} developed by Mistral Ai, is trained to understand both natural images and documents, demonstrates strong abilities in tasks such as chart and figure understanding, document question answering, multimodal reasoning, and instruction following. We specifically utilize the ``Pixtral-12B-2409'' version.
    \item \textsf{MiniCPM-V-2.6}~\citep{yao2024minicpm} is the latest and most capable model in the MiniCPM-V series developed by OpenBMB, achieves an average score of 65.2 on the latest version of OpenCompass, a comprehensive evaluation over 8 popular benchmarks.  We specifically utilize the ``openbmb/MiniCPM-V-2\_6'' version.
    \item \textsf{LLaVA-OneVision}~\citep{li2024llavaonevisioneasyvisualtask} is the first single model that can simultaneously push the performance boundaries of open LMMs in three important computer vision scenarios: single-image, multi-image, and video scenarios.  We specifically utilize the ``lmms-lab/llava-onevision-qwen2-7b-ov'' version.
    \item \textsf{Molmo}~\citep{deitke2024molmopixmoopenweights} developed by Allen Ai, is powerful model closes the gap between open and proprietary systems across a wide range of academic benchmarks as well as human evaluation.  We specifically utilize the ``allenai/Molmo-7B-D-0924'' version.
    \item \textsf{Qwen-VL}~\cite{bai2023qwen} is Alibaba Cloud's multimodal large vision-language model that excels in multilingual text recognition, fine-grained understanding, and multi-image interleaved conversations, significantly outperforming other large vision-language models in various benchmarks. We specifically utilize the ``Qwen/Qwen-VL-Chat'' version.
    \item \textsf{Qwen2-VL}~\citep{Qwen2VL} is the latest addition to the vision-language models in the Qwen series, building upon the capabilities of Qwen-VL.  We specifically utilize the ``Qwen/Qwen2-VL-7B-Instruct'' version.
    \item \textsf{mPLUG-Owl}~\cite{Ye2023mPLUGOwlME}, developed by DAMO Academy, is a training approach that enhances LLMs with multimodal capabilities by integrating a foundational LLM with a visual knowledge module and a visual abstractor module, using a two-stage method to align image and text. We specifically utilize the ``mplug-owl-llama-7b'' version.
    \item \textsf{MiniGPT-v2}~\cite{Chen2023MiniGPTv2LL} is a unified vision-language model designed for diverse tasks such as image description and visual question answering, utilizing unique task identifiers for improved performance and efficiency. We specifically built the model based on the ``llama-2-7b-chat'' LLaMA version with the checkpoint of the online developing demo.
    \item \textsf{Yi-VL}~\cite{Young2024YiOF} is an open-source multimodal vision-language model from the Yi LLM series, excelling in content comprehension and multi-round image conversations, and leading in recent English and Chinese benchmarks. We specifically utilize the ``Yi-VL-6B'' version.
    \item \textsf{xGen-MM}~\citep{xue2024xgenmmblip3familyopen} is a series of the latest foundational Large Multimodal Models (LMMs) developed by Salesforce AI Research. This series advances upon the successful designs of the BLIP series, incorporating fundamental enhancements that ensure a more robust and superior foundation. We specifically utilize the ``Salesforce/xgen-mm-phi3-mini-instruct-r-v1'' version.
    \item \textsf{MiniCPM-V-2}\footnote{https://huggingface.co/openbmb/MiniCPM-V-2} is a robust multimodal large language model designed for efficient end-side deployment. It is built on the foundation of SigLip-400M and MiniCPM-2.4B, connected by a perceiver resampler. We specifically utilize the ``MiniCPM-V 2.0'' version.
    

\end{itemize}

\begin{figure*}[htbp]
    \centering
    \includegraphics[width=\linewidth]{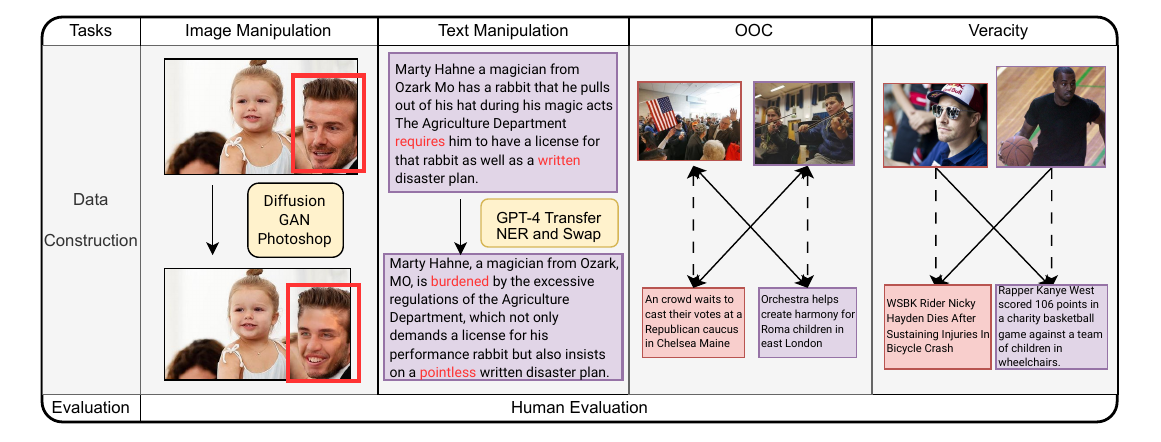}
    \caption{The pipeline of dataset construction.}
    \label{fig:data_construction}
\end{figure*}

\section{Implementation Details}
The data processing for our datasets is centered around Figure \ref{fig:data_construction}, which handles both image and text data to construct the benchmark.
\subsection{Data Construction}
\subsubsection{Manipulation Classification}
To explore the potential capacity of LVLMs on Manipulation Classification in a multimodal context, we designed seven types of manipulation, selecting data from the DGM4 dataset~\cite{Shao_2023_CVPR} and constructing additional datasets ourselves. The initial data was sourced from the VisualNews~\cite{liu-etal-2021-visual} datasets. The DGM4 dataset complies with the Apache-2.0 license. The VisualNews dataset is available upon request.

\begin{itemize}
  \item \textbf{Method 1: \underline{F}ace \underline{S}wap (FS).} Face Swap involves the process of cutting a face from one image and replacing it with a different face in another image. It can be used to create realistic but fake images of public figures, such as politicians, celebrities, or journalists, appearing to do things they never did. It is important for LVLMs not only to verify the authenticity of news text content but also to accurately identify whether the individuals in the accompanying photos correspond to the reported events. We have sampled and chosen a Face Swap subset of the DGM4 dataset~\cite{Shao_2023_CVPR} as part of our benchmark to \textit{detect Whether LVLM can recognize public figures and retrieve information related to individuals from its internal parametric knowledge} through multimodal data. 
  
  \textbf{Data processing}: A Face Swap subset of the DGM4 dataset~\cite{Shao_2023_CVPR} was sampled and selected.
  \item \textbf{Method 2: Face \underline{A}ttribute \underline{E}dit (AE).} Unlike Face Swap, Face Attribute Edit achieves deception by altering the facial expressions of humans like newsmakers. This can be potentially harmful to the public, as it can particularly portray a public figure laughing inappropriately in a serious context, which is highly misleading and infuriating. To identify such discrepancies, LVLMs must precisely recognize the type of event and the expected demeanor of the individuals involved.
  Our benchmark randomly selected visual and textual samples related to face attribute editing from the previously established DGM4 dataset \cite{Shao_2023_CVPR}. This inclusion allows us to \textit{evaluate the multimodal fact-checking capabilities of LVLMs in recognizing the scene, identifying personal information and detecting the correctness of face's status} in visual content in the multimodal context. 
  
  \textbf{Data processing}: Visual and textual samples related to face attribute editing were randomly selected from the previously established DGM4 dataset~\cite{Shao_2023_CVPR}.
  \item \textbf{Method 3: \underline{B}ackground \underline{C}hange (BC).} The same individuals, involving the same events, can take place in different locations. Before the emergence of diffusion models, manipulating a suitable scene was extremely challenging. However, with the advent of diffusion models \cite{Rombach_2022_CVPR}, we can now effortlessly alter the background of images, thereby creating scenes that did not originally exist in fact. Specifically, we are interested in whether LVLMs can exactly determine if the time and location of an event align with the actual scene. We utilized Grounding DINO~\cite{Liu2023GroundingDM} and \texttt{stable-diffusion-inpainting}\footnote{https://huggingface.co/runwayml/stable-diffusion-inpainting} models to generate a background for a manipulated and unrealistic scene. Our objective was to \textit{examine the capability of LVLMs in faithfully identifying these artificially constructed counterfactual scenarios}. 
  
  \textbf{Data processing}: Backgrounds for outdoor scenes were generated using Grounding DINO~\cite{Liu2023GroundingDM} and stable-diffusion-inpainting techniques. First, we used Grounding DINO to detect the people in the photos and create inverse masks. Then, we provided these masks along with the original images for stable-diffusion-inpainting. The pipeline was implemented using ComfyUI.

  \item \textbf{Method 4: \underline{C}LIP-based Stable Diffusion  \underline{G}enerate (CG).} Stable diffusion (SD) traditionally employs the text-to-image generation. However, by incorporating CLIP~\cite{radford2021learning}, we can transform the process into an image-to-image generation \cite{Ramesh2022HierarchicalTI}, enabling the manipulated image to retain the linguistic information from the original image. It is crucial for LVLMs to accurately discern between authentic and fabricated images by incorporating their internal knowledge,
  Using StabilityAI's \texttt{Stable-Diffusion-2-1-Unclip}\footnote{https://huggingface.co/stabilityai/stable-diffusion-2-1-unclip}, we generated stable diffusion versions of the original images for replacement. This design allows us to \textit{test the fact-checking capacity of LVLMs for awareness of whether the multimodal contents have been manipulated with the original image information}. 
  
  \textbf{Data processing}: Stable diffusion versions of the original images were generated using StabilityAI's Stable-Diffusion-2-1-Unclip. By utilizing Stable-Diffusion-2-1-Unclip, we input the original claim and image into the model to generate the manipulated images.
  
  \item \textbf{Method 5: \underline{P}hoto\underline{s}hop (PS).}  Photoshop has long been a leading tool for manual image editing, enabling users to alter human figures and merge different images to create potentially misleading visuals. This capability can have serious consequences, as it may lead to the spread of misinformation, manipulate public perception, and distort reality. LVLMs must leverage their inherent knowledge, which encompasses a vast understanding of context, patterns, and nuances in visual data, to effectively identify and analyze such issues of manipulation and misinformation. This facilitates our assessment of \textit{whether LVLMs can discern the traces of human manipulation, thereby fulfilling the requirements of the fact-checking task.}
  
 \textbf{Data processing}: To evaluate the effectiveness of LVLMs in detecting human manipulation, we utilize the photoshop subset of Fakeddit \cite{nakamura-etal-2020-fakeddit}.

  \item \textbf{Method 6: Textual \underline{E}ntity \underline{R}eplace (ER).} Textual Entity Replace is a traditional method of text manipulation. Using Named Entity Recognition (NER)~\cite{lample2016neural} from \texttt{bert-base-NER}\footnote{https://huggingface.co/dslim/bert-base-NER}, we identified named entities corresponding to persons within a given claim where newsmakers are mentioned. Subsequently, we randomly selected the location or time from an NER candidate set consisting of thousands of entities, to replace the target location or time entities in the claim. This creates counterfactual scenarios where the photos and claims contain the same individuals, but the events depicted are different. \textit{This scenario challenges the ability of LVLMs to keenly associate individuals with events, relying on their internal factual knowledge.}
  
  \textbf{Data processing}: Named entities corresponding to persons within a given claim were identified using Named Entity Recognition (NER)~\cite{lample2016neural} from bert-base-NER, and the surrounding contextual texts of the person would be replaced with other contexts of contradicted and different locations and time. To ensure that the claims contain people, we first screened the data and selected only the claims that included individuals.
  
  \item \textbf{Method 7: Text \underline{S}tyle \underline{T}ransfer (ST).} Similar to Face Attribute Edit, Text Style Transfer can alter the perception of the same person and event, giving a different factual impression. For instance, an originally sad event can be described in a way that makes it seem humorous. This poses a substantial challenge for fact-checking efforts as it requires LVLMs not only to detect the factual content but also to understand the tone and style nuances that might misrepresent the underlying truth of the situation. Hence, we first utilized GPT-4~\cite{OpenAI2023GPT4TR} to determine whether the sentiment of the text is positive or negative. Then, leveraging the advanced text style transfer capabilities of GPT-4, we rewrote the text to express the opposite sentiment. \textit{The process examines LVLMs' ability to rigorously comprehend the events and associated sentiments depicted in images and claims, and to correctly correlate them.}
  
  \textbf{Data processing}:The sentiment of the text was first determined using GPT-4~\cite{OpenAI2023GPT4TR}, and then the text was rewritten to express the opposite sentiment using GPT-4's advanced text style transfer capabilities.
\end{itemize}


\subsubsection{Out-of-Context Classification}
Out-of-Context (OOC) Classification~\citep{luo-etal-2021-newsclippings} aims to evaluate the coherence and correspondence of context across various modalities. Unlike the aforementioned manipulation techniques that require modifying images and texts, OOC Classification combines real but misused images and texts. If the image and claim are contextually aligned, we define the relationship as true. Conversely, if the image and claim are not contextually aligned, we define the relationship as false.
We collected multimodal samples from the NewsCLIPpings dataset~\cite{luo-etal-2021-newsclippings}, using embedding methods such as CLIP and SBERT-WK~\cite{Wang2020SBERTWKAS} to extract the most similar misused images, for \textit{the evaluation of LVLMs' ability in discerning subtle semantic inconsistencies between images and texts} in OOC Classification.

\textbf{Data processing}: The Out-of-Context Classification data is sourced from the NewsCLIPpings\citep{luo-etal-2021-newsclippings} dataset. The NewsCLIPpings dataset is available upon request.

\subsubsection{Veracity Classification}
Veracity Classification~\citep{10.1145/3539618.3591879} involves classifying the veracity of textual claims~\cite{lin2021rumor} given retrieved visual evidence. 
Based on the image evidence, the LVLMs need to predict the truthfulness (Supported, Refuted) of the claim. We curated a subset of the Mocheg dataset~\cite{10.1145/3539618.3591879} for this task. If the image supports the truthfulness of the claim, we label the relationship between the image and the claim as ``Supported'' indicating a true label. Otherwise, it is labeled as ``Refuted'' indicating a false label.
This is a cross-modal semantic transformation task designed to 
\textit{test whether LVLMs can accurately interpret and analyze visual information to support or refute textual claims}. 

\textbf{Data processing}: the Veracity Classification data is obtained and sampled randomly from the Mocheg dataset \cite{10.1145/3539618.3591879}. Mocheg dataset complies with the Apache-2.0 license.
\label{sec:license}

\begin{table*}[htbp] 
    \centering
    \begin{tabular}{lcccccccccc}
    \toprule
    \multirow{2}{*}{\textbf{Datasets}} & \multicolumn{7}{c}{\textbf{Manipulation}} & \textbf{OOC} & \textbf{Veracity} \\
    \cmidrule(lr){2-8}
    & \textbf{FS} &\textbf{AE} & \textbf{BC} & \textbf{CG} & \textbf{PS} & \textbf{ER} &\textbf{ST} \\
    \midrule
    Fakeddit~\citep{nakamura-etal-2020-fakeddit}  &\XSolidBrush  &\XSolidBrush & \XSolidBrush & \XSolidBrush & \textcolor{red}\CheckmarkBold & \XSolidBrush & \XSolidBrush & \textcolor{red}\CheckmarkBold & \XSolidBrush \\
    DGM4~\citep{Shao_2023_CVPR} & \textcolor{red}\CheckmarkBold  &\textcolor{red}\CheckmarkBold & \XSolidBrush & \XSolidBrush & \XSolidBrush & \XSolidBrush & \XSolidBrush & \XSolidBrush & \XSolidBrush \\
    MEIR~\citep{10.1145/3240508.3240707} &\XSolidBrush  &\XSolidBrush & \XSolidBrush & \XSolidBrush & \textcolor{red}\CheckmarkBold & \textcolor{red}\CheckmarkBold & \XSolidBrush & \XSolidBrush & \XSolidBrush \\ 
    EMU~\citep{da-etal-2021-edited} &\XSolidBrush  &\XSolidBrush & \XSolidBrush & \XSolidBrush & \textcolor{red}\CheckmarkBold & \XSolidBrush & \XSolidBrush & \XSolidBrush & \XSolidBrush \\ 
    \midrule
    
    Mocheg~\citep{10.1145/3539618.3591879}  &\XSolidBrush  &\XSolidBrush & \XSolidBrush & \XSolidBrush & \XSolidBrush & \XSolidBrush & \XSolidBrush & \XSolidBrush & \textcolor{red}\CheckmarkBold \\
    NewsCLIPpings~\citep{luo-etal-2021-newsclippings} &\XSolidBrush  &\XSolidBrush & \XSolidBrush & \XSolidBrush & \XSolidBrush & \XSolidBrush & \XSolidBrush & \textcolor{red}\CheckmarkBold & \XSolidBrush \\
    MAIM~\citep{10.1145/3123266.3123385}&\XSolidBrush  &\XSolidBrush & \XSolidBrush & \XSolidBrush & \XSolidBrush & \XSolidBrush & \XSolidBrush & \textcolor{red}\CheckmarkBold & \XSolidBrush \\ 
    COSMOS~\citep{10.1609/aaai.v37i12.26648}&\XSolidBrush  &\XSolidBrush & \XSolidBrush & \XSolidBrush & \XSolidBrush & \XSolidBrush & \XSolidBrush & \textcolor{red}\CheckmarkBold & \XSolidBrush \\
    \midrule
    
    MMFakeBench~\citep{liu2024mmfakebenchmixedsourcemultimodalmisinformation} &\XSolidBrush  &\XSolidBrush & \XSolidBrush & \textcolor{red}\CheckmarkBold & \textcolor{red}\CheckmarkBold & \textcolor{red}\CheckmarkBold & \XSolidBrush & \textcolor{red}\CheckmarkBold & \textcolor{red}\CheckmarkBold\\

    \midrule
    \benchname{} &  \textcolor{red}\CheckmarkBold &  \textcolor{red}\CheckmarkBold &  \textcolor{red}\CheckmarkBold &  \textcolor{red}\CheckmarkBold &  \textcolor{red}\CheckmarkBold &  \textcolor{red}\CheckmarkBold &  \textcolor{red}\CheckmarkBold & \textcolor{red}\CheckmarkBold &  \textcolor{red}\CheckmarkBold \\
    \bottomrule
    \end{tabular}
     \caption{Comparison of datasets related to multimodal fact-checking.}
    \label{tab:comparison}
\end{table*}

\subsection{Quality Assurance}
This research involved a human subjects study to evaluate the quality of multimodal data manipulated by our adopted techniques. To assure the quality of the self-constructed data, we employed three human evaluators, who are senior undergraduate or graduate students majoring in computer science. Each student was presented with the manipulated data and the original data to judge whether the data has been successfully manipulated with the manipulation techniques for the reliability and credibility of the multimodal data. Each evaluator completed the quality assurance process independently.

The manipulation accuracy for each task is presented in Table \ref{tab:accurary}, which highlights the effectiveness of our techniques. Additionally, the intra-class agreement score is 0.705. The average Spearman’s correlation coefficient between any two annotators is 0.714. These figures reflect the reliability of our data manipulation methods and the consistency of the evaluators' assessments.

\begin{table}[ht]\small
\centering
\begin{tabular}{lcc}
\toprule
\textbf{Types}                  & \textbf{Accuracy}  \\
\midrule
Background Change                & 0.97               \\
CLIP-based SD Generate           & 1.00               \\
Textual Entity Replace           & 0.99               \\
Text Style Transfer              & 0.98               \\
\bottomrule
\end{tabular}
\caption{Manipulation Accurary for Different Types.}
\label{tab:accurary}
\end{table}

The following considerations were adhered to ensure the protection and ethical treatment of participants: 1) Voluntary Participation: All participants were informed about the nature of the research and their role in it. Participation was entirely voluntary, with participants having the right to withdraw at any time without any consequences. 2) Informed Consent: Written informed consent was obtained from all participants. This consent form detailed the purpose of the research, the procedures involved, potential risks, and measures taken to safeguard participant data. 3) Data Anonymity and Confidentiality: All data collected during the study were anonymized. Personal identifiers were removed to maintain confidentiality and data were stored securely to prevent unauthorized access. 4) Minimal Risk: The study involved minimal risk to participants. The tasks performed were similar to everyday activities, and no sensitive personal information was requested or recorded.

\subsection{Comparison}
As shown in Table \ref{tab:comparison}, our benchmark includes more comprehensive data and covers a wider range of sub-tasks in multimodal fact-checking. Our dataset consists of three types of tasks and nine specific data categories.

\subsection{GPUs Usage}
\label{sec:gpus}
We utilized the high-performance computing platform and employed Slurm to request 2-4 A800 GPUs for benchmarking multimodal fact-checking with LVLMs.

\section{Related Work}

\subsection{LLMs and LVLMs}
Recent advancements have seen LLMs excel across various domains, with major tech companies developing high-performing proprietary models such as OpenAI's GPT-3~\citep{brown2020language}  and GPT-4~\citep{OpenAI2023GPT4TR}, Google's PaLM~\citep{chowdhery2022palm} and Gemini~\citep{team2023gemini}, and Anthropic's Claude. These models, however, are often only accessible via specific APIs or not at all. In contrast, the AI community has embraced the emergence of open-source LLMs, making significant contributions like MistralAI's Mistral-series~\citep{mistral}, Google's UL2-20B~\citep{tay2023ul2} and Gemma~\cite{gemma}, Tsinghua University's GLM-130B~\citep{zeng2023glm130b}, and Meta's OPT~\citep{opt} and the LLaMA series~\citep{touvron2023llama,touvron2023llama2,llama3}, enhanced by extensive alignment efforts~\citep{wang2023selfinstruct,xu2023wizardlm,luo2023wizardcoder,luo2023wizardmath,mukherjee2023orca,zhou2023lima,li2023selfalignment}.

LVLMs have significantly advanced the understanding of both textual and visual data within a unified framework~\cite{chen2023vlp, lin2023beneath, zhang2024mm}. Innovative models such as Flamingo~\citep{alayrac2022flamingo} and PaLM-E~\citep{driess2023palm} have demonstrated the ability to integrate visual and textual information effectively, without the need for task-specific training. Concurrently, the development of diverse multimodal datasets~\citep{yang2023dawn} stemming from GPT-4 and GPT-4V~\citep{OpenAI2023GPT4TR} has spurred the fine-tuning of models like LLaVA~\citep{liu2023visual}, MiniGPT-4~\citep{zhu2023minigpt}, mPLUG-Owl~\citep{ye2023mplug}, InstructBLIP~\citep{Dai2023InstructBLIPTG}, and others~\citep{bai2023qwen, wang2023cogvlm, gong2023multimodal, team2023gemini, fuyu-8b}, highlighting a trend towards more versatile and real-world applicable multimodal systems.

\subsection{Factual Knowledge in LMs}
Previous studies have established that language models (LMs) can function as repositories of factual knowledge, serving effectively as knowledge bases~\citep{petroni2019language, petroni2020context, heinzerling2021language}. This reservoir of factual information acquired during pretraining proves beneficial for knowledge-intensive tasks, such as question-answering and fact-checking~\citep{roberts2020much, lin2022amif, yu2022generate, pan2023fact}. \citet{petroni2019language} used cloze tests involving triples and tailored prompts to evaluate the factual knowledge embedded in language models, while \citet{jiang2020x} focused on optimizing prompt design to enhance factual retrieval from these models.

Despite these advancements, the reliability of these methods has been questioned. \citet{elazar2021measuring} highlighted the inconsistency in rank-based probing methods when using paraphrased contexts. Similarly, \citet{cao2021knowledgeable} argued that biased prompting and the leakage of correct answers can often lead to an overestimation of LMs' knowledge retention. On the other hand, \citet{varshney2022investigating} employed question-answering formats to gauge models’ uncertainty about specific facts, suggesting a different approach to measure factual accuracy. Our methodology aligns more closely with the approaches of \citet{kadavath2022language, lin2022teaching, hu2024large}, which involve querying models directly to self-evaluate their accuracy in delivering factual responses, offering a more direct assessment of their knowledge capabilities.  But differently, this work focuses on the multimodal nature of fact checking to explore the complex reasoning capability of LVLMs.


\subsection{Multimodal Fact-Checking}
Multimodal Fact-Checking refers to the systematic process of identifying counterfactuals or inconsistencies between facts across different modalities within multimodal data \citep{akhtar2023multimodal}. Common manifestations of multimodal misinformation include claims about digitally manipulated context \citep{Agarwal2019ProtectingWL,Shao_2023_CVPR} and the amalgamation of context from disparate modalities and contexts \citep{luo-etal-2021-newsclippings, aneja2021cosmos}. The former is predominantly associated with deepfake technologies \cite{Maras2018DeterminingAO,Dolhansky2019TheDD}, while the latter is linked with cheapfake methodologies \cite{aneja2021cosmos}. An essential Multimodal Fact-Checking pipeline consists of evidence retrieval and the adjudication process. Evidence retrieval furnishes the foundational basis for subsequent multimodal judgments. Within the adjudication phase, tasks are delineated into distinct categories, such as Manipulation Classification, Out-of-Context Classification, and Veracity Classification.

Manipulation Classification \citep{Shao_2023_CVPR} is a task meticulously designed to ascertain whether multimodal data encompasses fabricated elements. Out-of-context Classification \citep{luo-etal-2021-newsclippings} aims to evaluate the coherence and correspondence of context across various modalities. Veracity Classification \citep{10.1145/3539618.3591879} involves assessing whether the context from one modality aligns with or accurately reflects the context from another modality. Collectively, these tasks constitute the comprehensive process of multimodal fact-checking. In this work, we employed six different manipulation techniques to assess whether LVLMs can detect manipulations in multimodal news. Data from the NewsCLIPpings dataset is used to challenge LVLMs' ability to discern semantic differences between real images and real text, specifically for OOC classification. Similar to text, the cross-modal Veracity task is used to evaluate LVLMs' ability to perform factual inference across different modalities.

\subsection{Benchmarks for LVLMs}
Traditional multimodal benchmarks have been centered around specific skills such as visual recognition~\citep{goyal2017making}, image description~\citep{agrawal2019nocaps}, and visual commonsense reasoning~\citep{zellers2019recognition}. However, the advent of advanced LVLMs has necessitated the development of new benchmarks to keep pace with their robust zero-shot capabilities, which often exceed those measured by conventional metrics. This has exposed shortcomings in their ability to match answers accurately, highlighting issues with robustness. To address these limitations, the research community has introduced several innovative benchmarks, such as MME~\citep{fu2023mme}, MMBench~\citep{liu2023mmbench}, MM-Vet~\citep{yu2023mm}, SEED-Bench~\citep{li2023seed}, GOAT-Bench~\citep{lin2024goat}, LAMM~\citep{yin2023lamm} and MMCode~\cite{li2024mmcode}. These benchmarks are designed to facilitate structured evaluations of complex multimodal tasks and reveal the flaws of traditional methods. Distinct from these, our proposed benchmark is tailored to systematically assess multimodal factual knowledge, especially concerning disinformation detection in the realm of deepfakes and cheapfakes. This testbed would allow for a more thorough exploration of LVLMs' trustworthy awareness concerning a wider range of task types associated with multimodal factuality.

\begin{table*}[t] 
    \centering
   \small
    \begin{tabular}{lccccccc}
        \toprule
        \multirow{2}{*}{\textbf{Models}} & \multirow{2}{*}{\textbf{Size}} & \multicolumn{2}{c}{\textbf{Manipulation}} & \multicolumn{2}{c}{\textbf{OOC}}& \multicolumn{2}{c}{\textbf{Veracity}} \\
        \cmidrule(lr){3-4} \cmidrule(lr){5-6}  \cmidrule(lr){7-8} 
        {} & {} & \textbf{Accuracy} & \textbf{F1} & \textbf{Accuracy} & \textbf{F1} &  \textbf{Accuracy} & \textbf{F1} \\
        \midrule
        
        \rowcolor{pink!50}
        \multicolumn{8}{c}{\textit{Proprietary Models}}\\
        \Openaiemoji{}~\textbf{GPT-4o} & - & \underline{65.7} & \underline{60.4} & \textbf{84.8 }& \textbf{84.8} & 80.1 & 63.0\\
        \Openaiemoji{}~\textbf{GPT-4V} & - & 58.4 & 50.2 & 75.8 & 75.2 & 77.4& 60.0 \\
         \Claudeemoji{}~\textbf{Claude3.5-Sonnet} & - &59.9 & 41.7 & 49.9 & 37.6&72.7&47.4\\
        \Claudeemoji{}~\textbf{Claude3-Haiku} & - & 51.4 & 37.8 & 59.8 & 59.5 & 80.3 & 57.4 \\
        \Googleemoji{}~\textbf{Gemini-1.5-Pro}& - & 57.7 & 36.6 & \underline{80.2} & \underline{80.1} & 79.6 & 56.6\\
        \midrule
        \rowcolor{green!30}
        \multicolumn{8}{c}{\textit{Open-Source Models}}\\
        \BAAIemoji{}~\textbf{Emu2} & 37B & 38.7 & 33.0  & 51.9 & 51.1 & 70.0 & 52.6 \\
        \InternVLemoji{}~\textbf{InternVL} & 25.5B & 60.1 & 44.6 & 73.4 & 73.0 & 80.0 & 57.4\\
        \CogVLMemoji{}~\textbf{CogVLM} & 17B & 56.3 & 52.3& 61.4&56.2 &76.4 &63.4 \\
        \llavanextmoji{}~\textbf{LLaVA-NeXT} & 13B & 62.5 & 56.5 & 61.8 & 57.2 & 78.4 & 51.3\\
        \Instructblipemoji{}~\textbf{InstructBLIP} & 13B & 41.7 & 30.5& 59.5& 52.3 & 49.6 & 49.3\\
        \pixtralemoji{}~\textbf{Pixtral} & 12B & 58.5 & 43.9 & 64.8 &63.5 & 80.9 & 65.0 \\
        \MiniCPMVemoji{}~\textbf{MiniCPM-V-2.6} & 8B& 58.9 & 39.7 & 71.2 & 71.0 & 80.4 & 65.1 \\
        \llavanextmoji{}~\textbf{LLaVA-OneVision} &7B&61.5&55.5&75.7&75.4&80.9&60.3 \\
        \molmoemoji{}~\textbf{Molmo} & 7B & 59.3&59.3&58.9&52.3&79.9&57.6 \\
        \Qwenemoji{}~\textbf{Qwen-VL} & 7B  & 45.7 & 45.4 & 69.7& 69.4 & 82.7 & 69.3 \\
        \Qwenemoji{}~\textbf{Qwen2-VL} & 7B &59.9&46.6&80.1&\underline{80.1}&\underline{85.7}&\underline{75.5}\\
        \mPLUGOwlemoji{}~\textbf{mPLUG-Owl} & 7B & 45.7 & 45.4  & 48.3 & 46.1 & 60.8 & 49.7 \\
        \YiVLemoji{}~\textbf{Yi-VL} & 6B & 56.4& 43.8 & 70.4 &70.4 & 78.4 & 60.0\\
        \Instructblipemoji{}~\textbf{xGen-MM} & 5B&42.7&33.8&50.0&44.8&64.7&48.7 \\
        \MiniCPMVemoji{}~\textbf{MiniCPM-V-2} & 2.8B & 64.0 & 56.6  & 67.2 & 66.3 & 81.8 & 65.5 \\
        \midrule
        \rowcolor{blue!30}
        \multicolumn{8}{c}{\textit{Human}}\\
        \humanemoji{}~\textbf{Human} & - & \textbf{75.7} & \textbf{75.6} & 74.0 & 73.5 & \textbf{96.0} & \textbf{91.7} \\
        \bottomrule
    \end{tabular}
     \caption{Results of different LVLMs on the \benchname{}, in the zero-shot setting. The accuracy and macro-averaged F1 score (\%) are reported as the metrics.}
    \label{tab:result-zero-shot-full}
\end{table*}

\section{More Results and Analysis}
\label{analysis}
\subsection{Zero-shot Evaluation Results}
Table~\ref{tab:result-zero-shot-full} shows the zero-shot evaluation results of a total of 20 LVLMs on the \benchname{} in the zero-shot setting.

\subsection{Zero-shot CoT Evaluation Results} 
Table~\ref{tab:result-zero-shot-cot} shows the zero-shot CoT evaluation results of a total of 7 LVLMs on the \benchname{} in the zero-shot  CoT setting.

\subsection{Potential Test Set Leakage}
For the open-source LVLMs, test set leakage is not a concern, as the literature explicitly delineates the datasets and instruction-tuning procedures employed in their training, none of which encompass the multimodal data utilized in our \benchname{}. However, we cannot fully guarantee the exclusion of potential data leakage with the proprietary models, as its internal workings remain opaque. Nevertheless, as evidenced by the results in the experiments, where all LVLMs were evaluated directly on the \benchname{}, the absence of significant test set leakage is implied. This is inferred from the fact that direct application of the LVLMs did not yield disproportionately high performance, which would be expected if the models were benefiting from test set leakage.

\begin{figure*}
  \centering
    \includegraphics[page=1,width=\textwidth]{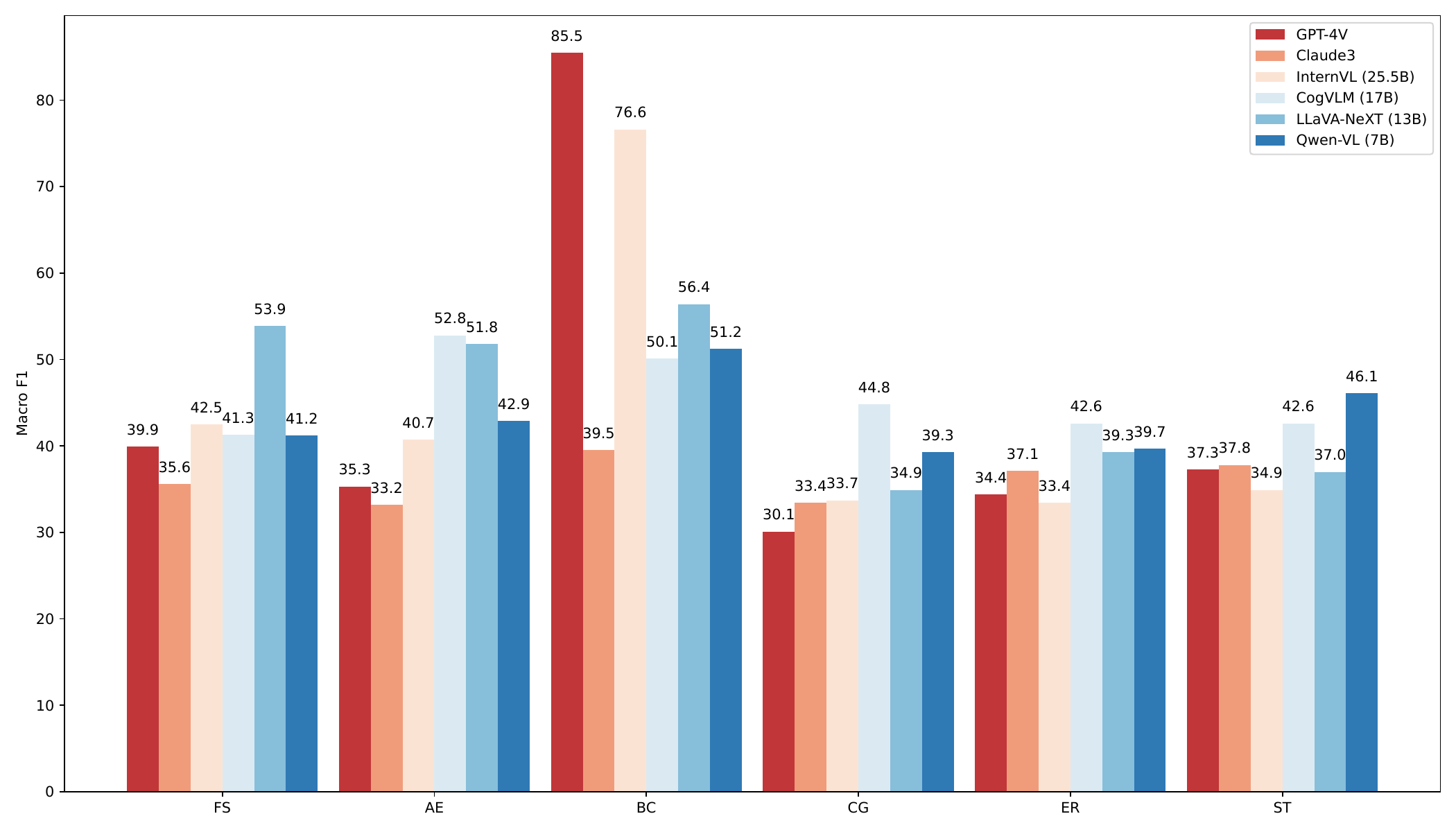}
  \caption{Effect of prompts specifically designed for different types of manipulation techniques.}
  \label{fig:fine}
  \vspace{-0.4cm}
\end{figure*}

\subsection{Results on Different Manipulation Techniques}
We further provide the detailed results of the representative LVLMs on the Manipulation Classification with respect to the seven manipulation methods, as depicted in Table~\ref{tab:result-manipulation-zero-shot}.

\subsection{Effect of Prompts on Manipulation Classification}
To verify the model's understanding of manipulation data, we designed prompts for six different manipulation methods and tested them on twelve models (see \S\ref{sec:prompts}). As shown in Figure \ref{fig:fine}, the model's performance on each sub-task was consistent with that of a single prompt. This suggests that the model struggles with manipulation fact-checking. For the Background Change task, the scenarios we set might have been too simple, making it easy for the model to detect the manipulations.

\begin{table*}[htbp] \LARGE
    \scriptsize
    \centering
    \begin{tabular}{lccccccccccccccc}
        \toprule
        \multirow{2}{*}{Models}& \multirow{2}{*}{Size} & \multicolumn{2}{c}{FS} & \multicolumn{2}{c}{AE}& \multicolumn{2}{c}{BC}& \multicolumn{2}{c}{CG}& \multicolumn{2}{c}{PS}& \multicolumn{2}{c}{ER}& \multicolumn{2}{c}{ST} \\
        \cmidrule(lr){3-4} \cmidrule(lr){5-6}  \cmidrule(lr){7-8}  \cmidrule(lr){9-10} \cmidrule(lr){11-12} \cmidrule(lr){13-14}\cmidrule(lr){15-16}\\
        {} & {}& Acc. & F1 & Acc. & F1 &  Acc. & F1 
        & Acc. & F1 & Acc. & F1 &  Acc. & F1&  Acc. & F1 \\
        \midrule
        \rowcolor{pink!50}
        \multicolumn{16}{c}{\textit{Proprietary Models}}\\
        \Openaiemoji{}~\textbf{GPT-4o} & - & 61.4 & 45.7 & 60.8 & 42.9 & 78.6 & 73.2& 63.6&60.8&80.4&80.4&58.1&53.7&56.8&49.5\\
        \Openaiemoji{}~\textbf{GPT-4V} & - &52.5 & 40.7  & 49.5 & 37.1 & 82.2 & 81.3 & 52.3 & 44.6&77.3&77.2 &47.5 & 36.3 & 47.3 & 34.2 \\
         \Claudeemoji{}~\textbf{Claude3.5-Sonnet} & - &62.7&41.0&64.4&39.7&69.0&47.6&53.7&36.9&59.3&49.2&58.7&38.8&51.0&35.5\\
        \Claudeemoji{}~\textbf{Claude3-Haiku} & - & 50.2&35.8&50.2&36.1&50.0&35.5&50.2&35.7&51.4&42.3&57.4&42.3&50.7&37.2 \\
        \Googleemoji{}~\textbf{Gemini-1.5-Pro}& - & 63.2&49.1&62.8&47.3&77.8&71.1&54.4&45.5&84.3&84.3&61.2&51.0&56.8&48.3\\
        \midrule
        \rowcolor{green!30}
        \multicolumn{16}{c}{\textit{Open-Source Models}}\\
        \BAAIemoji{}~\textbf{Emu2} & 37B & 35.5 & 30.7 & 35.3 & 30.0 & 32.7 & 25.9 & 33.6 & 28.8 & 57.3 & 52.6 & 57.7 & 42.6 & 49.8 & 38.1 \\
        \InternVLemoji{}~\textbf{InternVL} & 25.5B & 64.4 & 44.4 & 65.1 & 43.9 & 78.9 & 71.3 & 53.0 & 41.5 & 52.1 & 39.4 & 57.8 & 37.0 & 50.5 & 36.2\\
        \CogVLMemoji{}~\textbf{CogVLM} & 17B &54.0 & 51.6 & 53.1 & 50.4 & 71.7 & 70.5 & 60.7 & 58.9 & 50.0 & 33.4 & 41.9 & 29.5 & 48.2 & 41.1\\
        \llavanextmoji{}~\textbf{LLaVA-NeXT} & 13B &  60.7 & 51.2 & 60.5 & 50.7 & 81.8 & 79.9 & 61.9 & 59.6 & 63.5 & 59.9 & 54.2 & 41.5 & 55.5 & 51.2\\
        \Instructblipemoji{}~\textbf{InstructBLIP} & 13B &33.6 & 25.7 & 33.6 & 25.8 & 33.6 & 25.7 & 50.5 & 35.8 & 49.1 & 33.4 & 42.2 & 30.9 & 50.7 & 36.7\\
        \pixtralemoji{}~\textbf{Pixtral} & 12B &64.4&44.9&64.5&44.9&66.9&50.5&50.5&38.7&57.3&52.7&57.2&42.5&52.0&41.1\\
        \MiniCPMVemoji{}~\textbf{MiniCPM-V-2.6} & 8B& 66.2&41.6&66.3&42.0&68.1&45.8&50.4&35.5&54.0&43.3&57.6&37.5&49.9&34.2\\
        \llavanextmoji{}~\textbf{LLaVA-OneVision} &7B&59.9&51.3&58.7&49.9&78.5&73.0&60.9&56.2&71.6&71.0&55.2&37.9&48.2&35.1 \\
        \molmoemoji{}~\textbf{Molmo} & 7B & 51.4&50.2&52.3&51.0&64.6&64.3&70.4&69.8&61.4&56.0&47.1&45.9&51.2&51.1 \\
        \Qwenemoji{}~\textbf{Qwen-VL} & 7B  & 45.4 & 45.2 & 46.3 & 46.1 & 46.9 & 46.8 & 46.9 & 46.2 & 41.6 & 41.6 & 47.2 & 46.4 & 40.2 & 40.0 \\
        \Qwenemoji{}~\textbf{Qwen2-VL} & 7B &64.8&45.5&64.7&44.7&74.5&64.5&51.0&37.9&65.8&65.7&55.5&37.6&51.7&39.0\\
        \mPLUGOwlemoji{}~\textbf{mPLUG-Owl} & 7B &45.5 & 45.5 & 45.1 & 45.1 & 47.7 & 47.7 & 50.5 & 49.4 & 47.1 & 46.2 & 50.3 & 44.7 & 49.2 & 48.2 \\
        \YiVLemoji{}~\textbf{Yi-VL} & 6B & 65.3 & 44.2 & 64.7 & 43.7 & 68.9 & 50.5 & 51.2 & 40.2 & 64.7 & 63.5 & 56.4 & 37.4 & 49.6 & 36.8\\
        \Instructblipemoji{}~\textbf{xGen-MM} & 5B&35.3&29.6&35.4&29.7&35.1&29.5&49.9&36.5&50.0&33.6&48.4&43.0&49.5&36.3 \\
        \MiniCPMVemoji{}~\textbf{MiniCPM-V-2} & 2.8B &  62.2 & 50.4 & 62.5 & 50.1 & 83.7 & 85.8 & 63.1 & 59.9 & 70.7 & 70.2 & 56.8 & 39.2 & 49.6 & 38.9  \\
        \midrule
        \rowcolor{blue!30}
        \multicolumn{16}{c}{\textit{Human}}\\
        \humanemoji{}~\textbf{Human} & - & 63.0 & 62.9 & 71.0 & 70.9 & 92.0 & 92.0 &91.0&91.0& 75.9 & 75.4&59.0&58.8&78.0&77.9 \\
        \bottomrule

    \end{tabular}
    \caption{Detailed results of LVLMs on the Manipulation Classification in the zero-shot setting.}

    \label{tab:result-manipulation-zero-shot}
\end{table*}

\subsection{Human Evaluation} 
\label{he}
To assess the effectiveness of the \benchname{} and better evaluate the performance of LVLMs, we conducted human evaluation experiments. For each sub-task, as illustrated in Figure~\ref{fig:architecture}, we randomly selected 100 samples, resulting in a total of 900 examples for human evaluation. 3 professional fact-checking annotators (between the ages of 26 and 29) were asked to judge the truthfulness of each sample (i.e., ``Fact.'' or ``Non-Fact.'') in the zero-shot evaluation setting. Then the voting results were regarded as the answers.

As demonstrated in Table~\ref{tab:result-human} and Table~\ref{tab:result-human-fine}:
1) The accuracy of human predictions significantly surpasses LVLMs in Manipulation Classification. Humans achieved an accuracy of 75.67\% and an F1 score of 75.58\%. In Background Change and CLIP-based Stable Diffusion Generation methods, human accuracy exceeded 90\%. Human fact-checking ability in Manipulation Classification surpasses that of LVLMs, suggesting that there is considerable room for improvement in LVLM performance.
2) Human performance in OOC classification is on par with the best-performing LVLMs, such as GPT-4V. Without manipulating the text and image, LVLMs can effectively identify the false connections between them. 
3) For Veracity Classification, humans achieved an accuracy of over 95\%. This high accuracy can be attributed to two factors: the strong fact-checking abilities of humans and the high degree of correlation within the dataset, which allowed humans to draw on their experience.

Human performance exceeds that of most LVLMs, especially in Manipulation Classification. This indicates that there is still significant potential for improvement in the fact-checking capabilities of LVLMs.

\begin{table}[t] 
    
    \begin{tabular}{ccc}
        \toprule
        Tasks &    Accuracy & F1  \\
        \midrule
        Manipulation Classification  & 75.67 & 75.58 \\
        OOC Classification  & 74.00 & 73.50 \\
        Veracity Classification  & 96.00 & 91.70 \\
        \bottomrule
    \end{tabular}
    \caption{Results of human evaluation on the \benchname{} across different multimodal fact-checking tasks in a zero-shot setting.}
    \label{tab:result-human}
\end{table}

\begin{table}[t] 
    \centering
    \begin{tabular}{ccc}
        \toprule
        Tasks &    Accuracy & F1  \\
        \midrule
        FS & 63.0 & 62.9  \\
        AE  & 71.0 & 70.9 \\
        BC & 92.0 & 92.0 \\
        CG & 91.0 & 91.0 \\
        PS & 75.9 & 75.4 \\
        ER & 59.0 & 58.8 \\
        ST & 78.0 & 77.9 \\
        \bottomrule
    \end{tabular}
    \caption{Detailed results of human evaluation on the Manipulation Classification in the zero-shot setting.}

    \label{tab:result-human-fine}
\end{table}

\subsection{Model Interpretability}
\label{interpretability}

\begin{table*}[t]
    \centering
    \scalebox{0.8}{
    \begin{tabular}{lccccccccccccc}
    \toprule
    \multirow{2}{*}{\textbf{Models}} & \multirow{2}{*}{\textbf{Size}} & \multicolumn{4}{c}{\textbf{Manipulation}} & \multicolumn{4}{c}{\textbf{OOC}}& \multicolumn{4}{c}{\textbf{Veracity}} \\
    \cmidrule(lr){3-6} \cmidrule(lr){7-10}  \cmidrule(lr){11-14} 
    {} & {} & \textbf{M} & \textbf{I} & \textbf{S} & \textbf{R}  & \textbf{M} & \textbf{I} & \textbf{S} & \textbf{R} & \textbf{M} & \textbf{I} & \textbf{S} & \textbf{R} \\
    \midrule
    \rowcolor{pink!50}
    \multicolumn{14}{c}{\textit{Evaluated by GPT-4}}\\
    \llavanextmoji{}~\textbf{LLaVA-NeXT(7B)} & 7B& 3.95&3.09&3.24&4.39&3.82&3.09&3.54&	4.56 &3.68&2.69&3.12&4.22 \\
    \llavanextmoji{}~\textbf{LLaVA-NeXT(13B)} & 13B &3.83& 3.16  & 3.36 & 4.46  &3.57 & 3.17 & 3.70 & 4.61  &3.44& 2.89  & 3.39 & 4.41\\
    \Instructblipemoji{}~\textbf{InstructBLIP(7B)} & 7B & 3.86 & 1.06 & 1.47 & 2.24 & 3.04 & 1.11 & 1.87 &2.60  & 3.32 & 1.00 & 1.54 &2.21 \\
    \Instructblipemoji{}~\textbf{InstructBLIP(13B)} & 13B & 3.67 & 1.42 &1.92& 2.71  & 2.88 & 1.06 &1.69& 2.44  & 3.42 & 1.00 &1.53& 2.23\\
    \Qwenemoji{}~\textbf{Qwen-VL} & 7B & 4.02 & 1.83 & 2.61 & 3.73 & 3.82  & 1.64 & 2.45 & 3.47 & 3.43  & 1.85 & 2.83 & 3.85  \\
    \YiVLemoji{}~\textbf{Yi-VL} & 6B & 3.44 & 2.18 & 3.20 & 4.20  & 3.02 & 2.12 & 3.35 & 4.23 & 2.65 & 1.82 & 3.39 & 4.16 \\
    \rowcolor{blue!50}
    \multicolumn{14}{c}{\textit{Evaluated by Human}}\\
    \llavanextmoji{}~\textbf{LLaVA-NeXT(7B)} & 7B& 3.43 & 3.15  & 3.83 & 4.47 &  3.82 & 2.09  & 3.54 & 4.56  & 3.42 & 3.82  & 3.76 & 4.34   \\
    \llavanextmoji{}~\textbf{LLaVA-NeXT(13B)} & 13B&3.63& 3.43  & 3.96 & 4.87 &3.57 & 3.17 & 3.70 & 4.61 &3.83& 3.89  & 3.64 & 4.42\\
    \Instructblipemoji{}~\textbf{InstructBLIP(7B)}& 7B & 3.80 & 2.13 & 2.41 & 2.63 & 3.04 & 2.11 & 2.87 &3.45 & 3.25 & 2.41 & 2.06 &3.57 \\
    \Instructblipemoji{}~\textbf{InstructBLIP(13B)}& 13B & 3.78 & 2.17 &2.83& 2.76  & 2.88 & 2.06 &2.69& 3.95  & 3.30 & 2.40 &2.11& 3.83\\
    \Qwenemoji{}~\textbf{Qwen-VL} & 7B& 3.46  & 2.74 & 3.52 & 3.13 & 3.45  & 2.20 & 2.45 & 3.47& 3.91  & 2.96 & 3.35 & 4.31  \\
    \YiVLemoji{}~\textbf{Yi-VL} & 6B& 3.54 & 2.53 & 3.81 & 4.56  & 3.23 & 2.20 & 3.35 & 4.23 & 3.14 & 2.28 & 3.52 & 4.72 \\
    \bottomrule
    \end{tabular}
    }
    \caption{Model Interpretability Evaluated by GPT-4 and Human.}
    \label{tab:model_interpretability}
\end{table*}

To gain deeper insights into the model interpretability of LVLMs, we expand our research on the evaluation on the justfication production of LVLMs. The output format \( F\): ``Answer yes or no.'' was removed to allow the model to produce more intermediate reasoning steps. 

For the evaluation of justification production, traditional automated evaluation metrics are inadequate to assess the output results of LVLMs~\cite{chang2024survey}. Fortunately, GPT-4 has been demonstrated to excel in assessing text quality from multiple angles, even in the absence of reference texts~\cite{lin2024towards, wang2024explainable}. Thus the model's justification was evaluated by GPT-4 and Human subjects across four dimensions: Misleadingness (M), Informativeness (I), Soundness (S), and Readability (R). A 5-point Likert scale was used, where 1 indicates the lowest quality and 5 the highest for Informativeness, Soundness, and Readability, but the scale is reversed for Misleadingness.
\begin{itemize}
    \item \textbf{\underline{M}isleadingness (M)} assesses whether the model’s explanation is consistent with the real veracity label of a claim, with a rating scale ranging from 1 (not misleading) to 5 (very misleading).
    \item \textbf{\underline{I}nformativeness (I)}  measures how much the explanation provides new information, such as explaining the background and additional context, with a rating scale ranging from 1 (not informative) to 5 (very informative).
    \item \textbf{\underline{S}oundness (S)} describes whether the explanation seems valid and logical, with a rating scale ranging from 1 (not sound) to 5 (very sound).
    \item \textbf{\underline{R}eadability (R)} evaluates whether the explanation follows proper grammar and structural rules, and whether the sentences in the explanation fit together and are easy to follow with a rating scale ranging from 1 (not fluent) to 5 (very fluent).
\end{itemize}

To use GPT-4 to evaluate the model interpretability of LVLMs, we carefully designed the following prompt.
First, we give the GPT-4 system prompt ``\textit{You are now the judge of the model output.}''; Next, we provide GPT-4 with both the label $L$ and model output $Y$ using the format ``Label:\{L\}, Model output \{Y\}''. Finally, GPT-4 evaluates the output in four dimensions and  return with json format. Below is the complete prompt we use for GPT-4:

\textit{Label:\{L\}} 

\textit{Model output: \{Y\}}

\textit{Please rate in four dimensions}: 

\textit{1. Misleadingness  -assesses whether the model’s explanation is consistent with the real veracity label of a claim, with a rating scale ranging from 1 (not misleading) to 5 (very misleading) }

\textit{2. Informativeness - assesses whether the explanation provides new information, such as explaining the background and additional context, with a rating scale ranging from 1 (not informative) to 5 (very informative)}

\textit{3. Soundness - describes whether the explanation seems valid and logical, with a rating scale ranging from 1 (not sound) to 5 (very sound)}

\textit{4. Readability - evaluates whether the explanation follows proper grammar and structural rules, and whether the sentences in the explanation fit together and are easy to follow with a rating scale ranging from 1 (poor) to 5 (excellent).}

\textit{Scores 1-5, returned in json format.}

We conducted model interpretability analysis across six models: LLaVA-NeXT (7B), LLaVA-NeXT (13B), InstructBLIP (7B), InstructBLIP (13B), Qwen-VL, and Yi-VL. This investigation explored the differences within the same model family with varying parameter sizes, as well as the differences between distinct models.

\subsection{Effect of Model Size}
\begin{figure*}[ht]
    \centering
    \includegraphics[width=.98\linewidth]{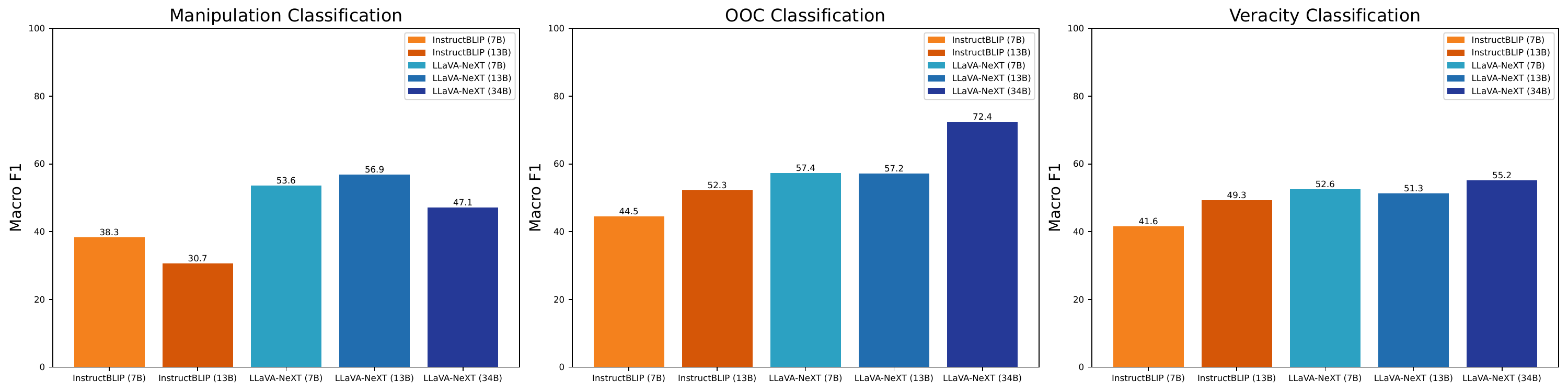}
    \caption{Model size effects of LVLMs.}
    \label{fig:model_size}
\end{figure*}

To explore the impact of model size on factual capabilities, we analyzed two families of LVLMs: InstructBLIP and LLaVA-NeXT, which both utilize the same language backbone, i.e., Vicuna~\cite{chiang2023vicuna}, and employ similar CLIP models, with InstructBLIP using EVA CLIP-g and LLava-NeXT using CLIP ViT-L/14. Specifically, we examined InstructBLIP (7B), InstructBLIP (13B), LLava-NeXT (7B), LLava-NeXT (13B), and LLava-NeXT (34B). As shown in Figure~\ref{fig:model_size}, the following observations were made: 1) In Manipulation Classification, there is a minimal correlation between the model size of the specific LVLMs family and the performance.
2) Regarding OOC Classification and Veracity Classification, the model performance generally improves with the increased model size.



\subsection{Yes/No Bias}
\label{sec:yes-no-bias}



\begin{figure}[ht]
    \centering
    \scalebox{1.}{\includegraphics[width=.95\linewidth]{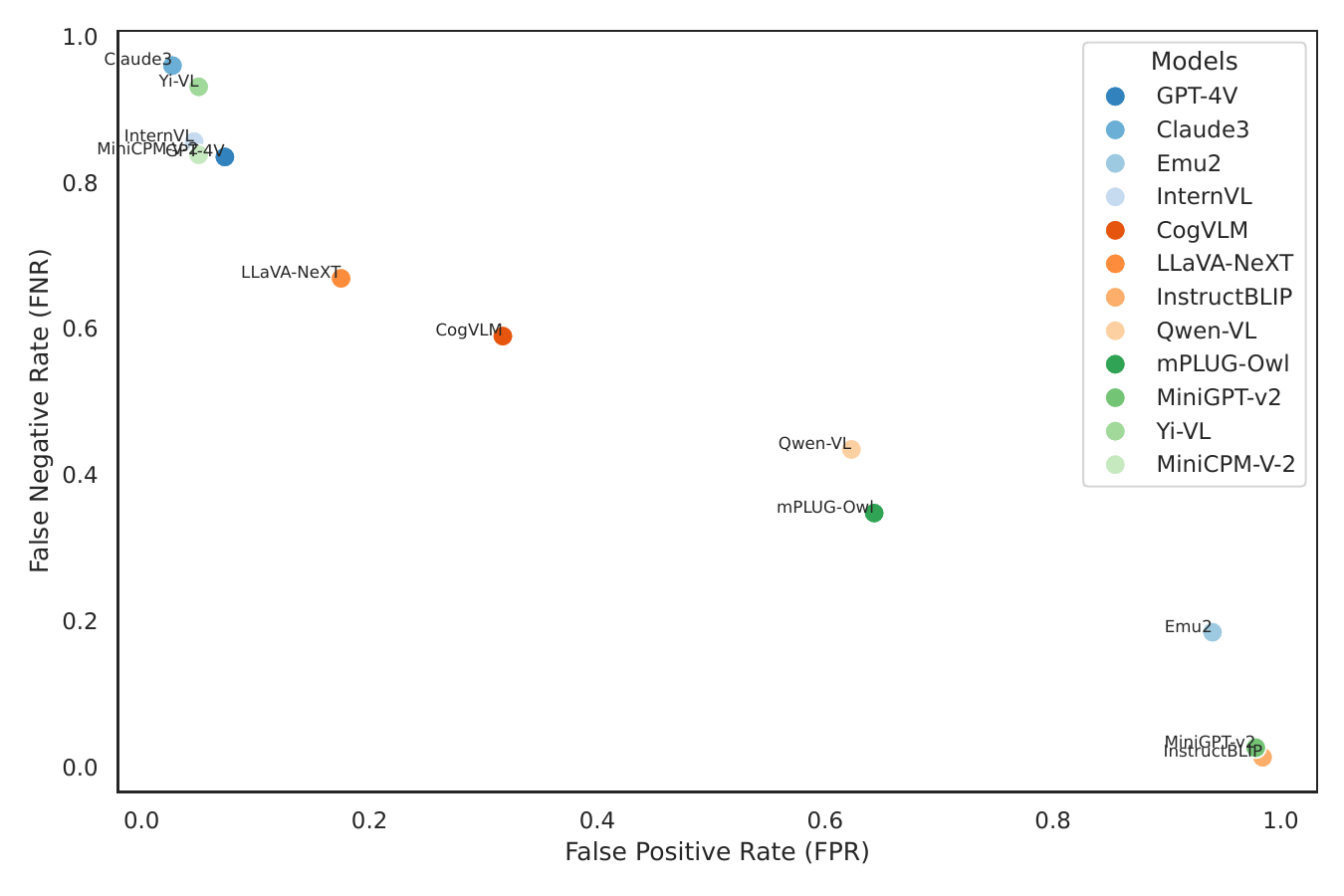}}
    \caption{Yes/No Bias in tested LVLMs.}
    \label{fig:yes-no-bias}
\end{figure}

During benchmarking, we identified a Yes/No Bias issue with the tested LVLMs, where it tends to consistently respond with either ``yes'' or ``no''.
We have chosen two key metrics to evaluate the Yes/No bias of the model for the Manipulation Classification task: 1) False Positive Rate (FPR)~\cite{fawcett2006introduction} and 2) False Negative Rate (FNR)~\cite{powers2020evaluation}.
In Figure~\ref{fig:yes-no-bias}, models such as GPT-4V, Claude3-Haiku, Yi-VL, and InternVL tend to answer ``no'' more frequently. Conversely, models like Emu2, MiniGPT-v2, and InstructBLIP are more inclined to answer ``yes''. Meanwhile, LLaVA-NeXT, CogVLM, Qwen-VL, and mPLUG-Owl exhibit a balanced performance without a strong bias towards either affirmative or negative classifications. Given that these models were not specifically trained for this task, the presence of such biases is not unexpected. This underscores the necessity of \benchname{}, aiming to guide the enhancement of fact-checking capabilities in LVLMs for future developments.

\subsection{Case Study}
\label{sec:case-study}
\begin{figure*}[t]
  \centering
    \includegraphics[width=\textwidth]{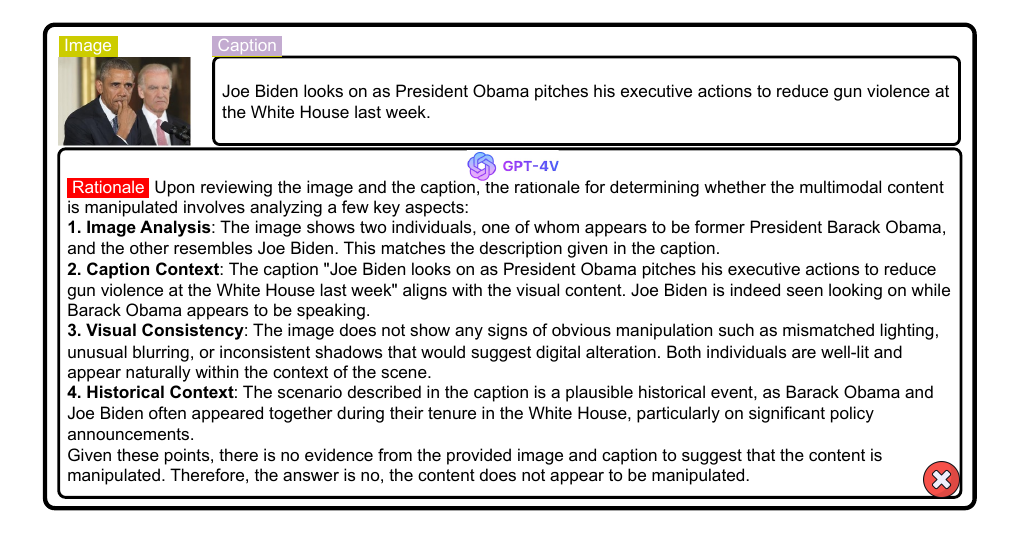}
  \caption{Example of wrongly predicted manipulation by GPT-4V with rationale.}
  \label{fig:Rationales-wrong}
\end{figure*}

\begin{figure*}[t]
  \centering
    \includegraphics[width=\textwidth]{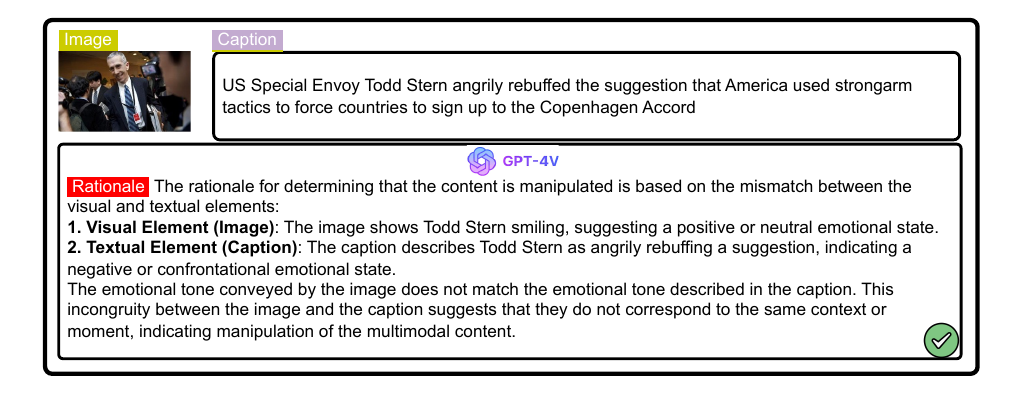}
  \caption{Example of correctly predicted manipulation by GPT-4V with rationale.}
  \label{fig:Rationales-correct}
\end{figure*}

To better understand the reasoning process of the model in fact-checking, we are conducting a study on the correct and incorrect reasoning processes of the GPT-4V model. 
Figure~\ref{fig:Rationales-wrong} illustrates an instance where GPT-4V fails to identify manipulated content, specifically a face swap involving Joe Biden and another individual. This oversight underscores a significant limitation of GPT-4V in accurately recognizing individuals within images. The model's rationale primarily emphasizes overall scene consistency and plausible historical context, but it fails to detect the specific manipulation of Joe Biden's identity. In contrast, Figure~\ref{fig:Rationales-correct} showcases GPT-4V's successful identification of manipulated content by accurately discerning the discrepancy between the emotional state depicted in the image and the corresponding caption. Todd Stern's smiling expression contrasts with the caption's description of him angrily rebuffing a suggestion. GPT-4V effectively recognizes this emotional mismatch, demonstrating its capability to evaluate the coherence between visual and textual elements.

\subsection{Error Analysis}
In zero-shot settings, the model's performance relies solely on its understanding of the instructions, its comprehension of the images and claims, and ultimately making a judgment based on this understanding(see also~\S\ref{sec:case-study}). The main results indicate that the model's fact-checking ability is weak. As discussed in \S\ref{sec:yes-no-bias}, the Yes/No Bias also highlights this issue.

In few-shot settings, the model does not gain insights from the examples. As Figure~\ref{fig:few-shot-cot} shows,  LLaVA-NeXT's usable response rate decreases, and it starts outputting gibberish instead of ``yes'' or ``no''. Specifically, in few-shot with CoT conditions, LLaVA-NeXT does not generate its own reasoning process but merely copies the rationale from previous examples. For example, one output from LLaVA-NeXT is, ``Answer yes or no. Rationale: The image shows what seems to be an unnatural or edited blend of faces, particularly noticeable in the features of the man and the child. This indicates that the image may have been digitally altered.'', which is already included in the demonstrations of the prompt.

\section{Prompts Designed for Manipulation Techniques}
\label{sec:prompts}

1.
Face Swap is a manipulation technique of cutting a face from one image and replacing it with a different face in another image. Your task is to determine if the claim and its image have used Face Swap. Answer yes or no.

2.
Face Attribute Edit is a manipulation technique for altering facial expressions. Your task is to determine if the claim and its image have used Face Attribute Edit. Answer yes or no.

3.
Background Change is a manipulation technique that involves altering the background of images. Your task is to determine if the claim and its image have used Background Change. Answer yes or no.

4.
CLIP-based Stable Diffusion Generation is a manipulation technique that utilizes an image-to-image generation pipeline to produce manipulated images. Your task is to determine if the claim and its image have used CLIP-based Stable Diffusion Generate. Answer yes or no.

5.
Textual Entity Replace is a manipulation technique that involves identifying named entities corresponding to persons in one text, locating these entities in another text, and swapping the surrounding contextual texts between the two. Your task is to determine if the claim and its image have used Textual Entity Replace. Answer yes or no.

6.
Text Style Transfer is a manipulation technique that rewrites text to express the opposite sentiment. Your task is to determine if the claim and its image have used Text Style Transfer. Answer yes or no.

\section{Discussion of Label Setting}
We considered the following points in adopting this design philosophy for label setting:
\begin{itemize}
    \item Simplicity and Clarity: As the first study to benchmark MFC with LVLMs, our design allows us to quantitatively assess the performance of LVLMs in a straightforward and intuitive manner. This simplicity facilitates preliminary in-depth analyses that more complex settings might not easily provide. We find it exciting to cleverly and flexibly unify three significant data types under the MFC umbrella without adding unnecessary complexity.
    \item Poor Performance of LVLMs:
    Despite high F1 score of 84.8\% on OOC Classification, the tasks are not too simple, as evidenced by lower F1 scores of 61.6\% and 75.5\% on Manipulation Classification and Veracity Classification. Besides, the best Accuracy and F1 on Manipulation Classification only achieve 64.0\% and 56.6\% by a lightweight LVLM, MiniCPM-V-2 (2.8B), leaving significant room to improve larger LVLMs that perform worse on this task.
    \item Appropriate Difficulty Levels: Our benchmark is designed to balance difficulty levels (i.e., OOC Classification: relatively easy; Veracity Classification: moderate; Manipulation Classification: relatively difficult), reflecting varying complexities to assess LVLM capabilities comprehensively. This integration allows for a broader evaluation of LVLMs' adaptability and generalization across diverse MFC data types.
    \item Foundation for Future Research: Our work lays the groundwork for future studies, which could incorporate more systematic human subject studies to explore interpretability and additional analytical dimensions. This potential for expansion underscores the value of our initial simplification and sets the stage for more complex investigations.

\end{itemize}

\section{Discussion of Real-World Scenarios}

The main contribution of our benchmark is to provide insights into the trustworthy issue for current researchers studying existing emerging LVLMs. For a real-world fact-checking process, there are stages like claim detection, evidence retrieval, claim verification, justification production, etc. Our work just directly provides the check-worthy data so that the claim detection stage could be omitted. Then, the LVLM is evaluated by retrieving the inherent evidence embedded in its internal parameters, which can be regarded as the evidence retrieval stage in this benchmark work. Finally, for fact verification, the LVLM is used to verify the factuality in the verdict prediction stage with produced justification. All the scenes are suitable for real-world events \cite{lin2024unleashing}. Our human subjects evaluations have verified the soundness and alignment of the multimodal data for real-world needs.




\end{document}